\documentclass[10pt,journal,compsoc]{IEEEtran}

\ifCLASSOPTIONcompsoc
  \usepackage[nocompress]{cite}
\else
  \usepackage{cite}
\fi

\ifCLASSINFOpdf
  \usepackage[pdftex]{graphicx}
  \DeclareGraphicsExtensions{.pdf,.jpeg,.png}
\else

\fi

\usepackage{amsmath}
\usepackage{acronym}
\usepackage{mdwmath}
\usepackage{mdwtab}
\usepackage{amssymb}
\usepackage{subcaption}
\usepackage{array}
\usepackage{kotex}

\begin{document}

\title{Exploring Adversarial Robustness of Vision Transformers in the Spectral Perspective}

\author{Gihyun Kim \qquad Juyeop Kim \qquad Jong-Seok Lee \\ 
Yonsei University, Republic of Korea \\
{\tt\small \{kkh9314,juyeopkim,jong-seok.lee\}@yonsei.ac.kr}\\}


\IEEEtitleabstractindextext{%
\begin{abstract}
The Vision Transformer has emerged as a powerful tool for image classification tasks, surpassing the performance of convolutional neural networks (CNNs). Recently, many researchers have attempted to understand the robustness of Transformers against adversarial attacks. However, previous researches have focused solely on perturbations in the spatial domain. This paper proposes an additional perspective that explores the adversarial robustness of Transformers against frequency-selective perturbations in the spectral domain. To facilitate comparison between these two domains, an attack framework is formulated as a flexible tool for implementing attacks on images in the spatial and spectral domains. The experiments reveal that Transformers rely more on phase and low frequency information, which can render them more vulnerable to frequency-selective attacks than CNNs. This work offers new insights into the properties and adversarial robustness of Transformers.
\end{abstract}

}

\maketitle

\IEEEdisplaynontitleabstractindextext

\IEEEpeerreviewmaketitle

\ifCLASSOPTIONcompsoc
\IEEEraisesectionheading{\section{Introduction}\label{sec:introduction}}
\else
\section{Introduction}
\label{sec:introduction}
\fi

\IEEEPARstart{C}{onvolution} neural networks (CNNs) have served as the dominant architecture for computer vision for a long time. However, Transformer-based structures have recently emerged as another promising architecture \cite{i:5}, achieving even better performance than CNNs especially in image classification.

CNNs are known to be vulnerable to adversarial attacks, i.e., an imperceptible perturbation added to an image can fool a trained CNN so that it misclassifies the attacked image \cite{a:7}. Investigating the robustness of a model against adversarial attacks is important because not only the vulnerability issue is critical in security-sensitive applications but also such investigation can lead to a better understanding of the operating mechanism of the model. Then, a naturally arising question is: how vulnerable are Transformers compared to CNNs?

The researches comparing adversarial robustness of CNNs and Transformers do not reach consistent conclusions. One group of studies claims that Transformers are more robust to adversarial attacks than CNNs \cite{a:1, a:6, m:2, m:3}. However, another group of studies claims that the two architectures have similar levels of robustness \cite{a:2, i:1, i:2}.

This paper aims to explore the adversarial robustness of Transformers from a previously unexplored perspective. It has been noted in previous studies that Transformers rely more on low frequency features \cite{i:13, a:6} while CNNs focus more on high frequency features \cite{i:15, m:6}. From this point of view, popular gradient-based attack methods, which is mostly used in the existing studies comparing adversarial robustness of CNNs and Transformers, tend to perturb high frequency features in images through spatial domain perturbations and this might cause CNNs to be fooled more easily than Transformers. To alleviate such bias, we formulate an attack framework that allows flexible perturbations in both spatial and spectral domains, with the hope to find certain types of adversarial perturbations for which Transformers become more vulnerable than CNNs. Note that we do not intend to develop a new stronger attack in the frequency domain, but aim to formulate a unified attack framework that can directly perturb the pixel values, magnitude spectrum, and phase spectrum of an image. 

\begin{figure}[t]
\centering
\includegraphics[width=0.35\textwidth]{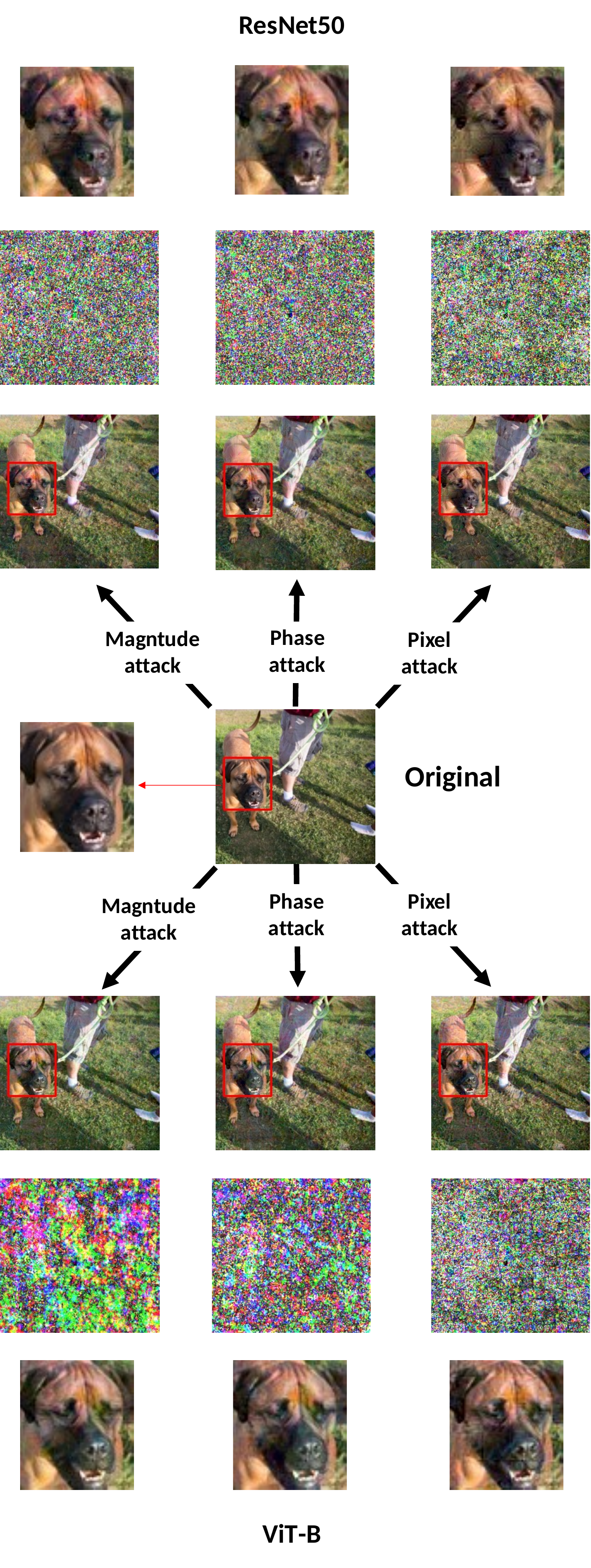}
\caption{Example case of our attack perturbing the magnitude, phase, and pixel values, respectively, for ResNet50 and ViT-B. The attacked image, the difference between the original and attacked images (after amplification for visualization), and an enlarged area of the attacked image are shown in each case.}
\vspace{-1em}
\label{figure1}
\end{figure}

Figure \ref{figure1} shows an example, where each of the magnitude, phase, and pixel components is perturbed using our attack framework for ResNet50 and ViT-B. It can be observed that attacking different components induces different distortion patterns in the image. The distortion pattern also varies depending on the target model. Thus, a standardized scale is required to make proper comparison among different target models and attack domains. We choose to utilize an image quality metric to measure the attack strength across various models and attack methods. In addition, we consider a wide range of attack strength because the superiority in terms of robustness between models may change depending on the amount of perturbation.

We conduct extensive experiments to compare the adversarial robustness of off-the-shelf pre-trained CNN and Transformer models. The results demonstrate that Transformers are not necessarily more robust than CNNs, and in particular, Transformers tend to be more vulnerable to perturbations inserted to the magnitude and phase components in the frequency domain. Our contributions can be summarized as follows.

\begin{itemize}
\item To explore adversarial robustness in both the spatial domain and spectral domain, we formulate an attack framework that can perturb the magnitude and phase spectra in the spectral domain and the pixel values in the spatial domain. In particular, examining frequency-selective perturbations by the attacks is one of the key factors to a deeper understanding of the adversarial robustness of Transformers.
\item Using the attack framework, we evaluate various models of CNN and Transformer over a wide range of image quality of attacked images. Relative vulnerability among the models is analyzed in various viewpoints such as attack type, attack strength, model size, and training data. As a main result, it is found that Transformers are particularly vulnerable to phase perturbations concentrated in the low frequency region.
\item We conduct in-depth analyses to investigate the frequency-dependent behaviors and importance of spectral information in Transformers.
Additionally, we explain the vulnerability of Transformers to the phase attack from the viewpoint of linearity of models and attacks.
\end{itemize}

The remainder of the paper is organized as follows. Section~\ref{related_works} briefly reviews the related works. In Section~\ref{method}, we present our method to explore the robustness of Transformers. The results are presented in Section~\ref{analysis} with rich analysis. Finally, conclusion is made in Section~\ref{conclusion}.

\begin{figure*}
    \centering
    \begin{subfigure}[t]{0.25\textwidth}
    \includegraphics[width=\textwidth]{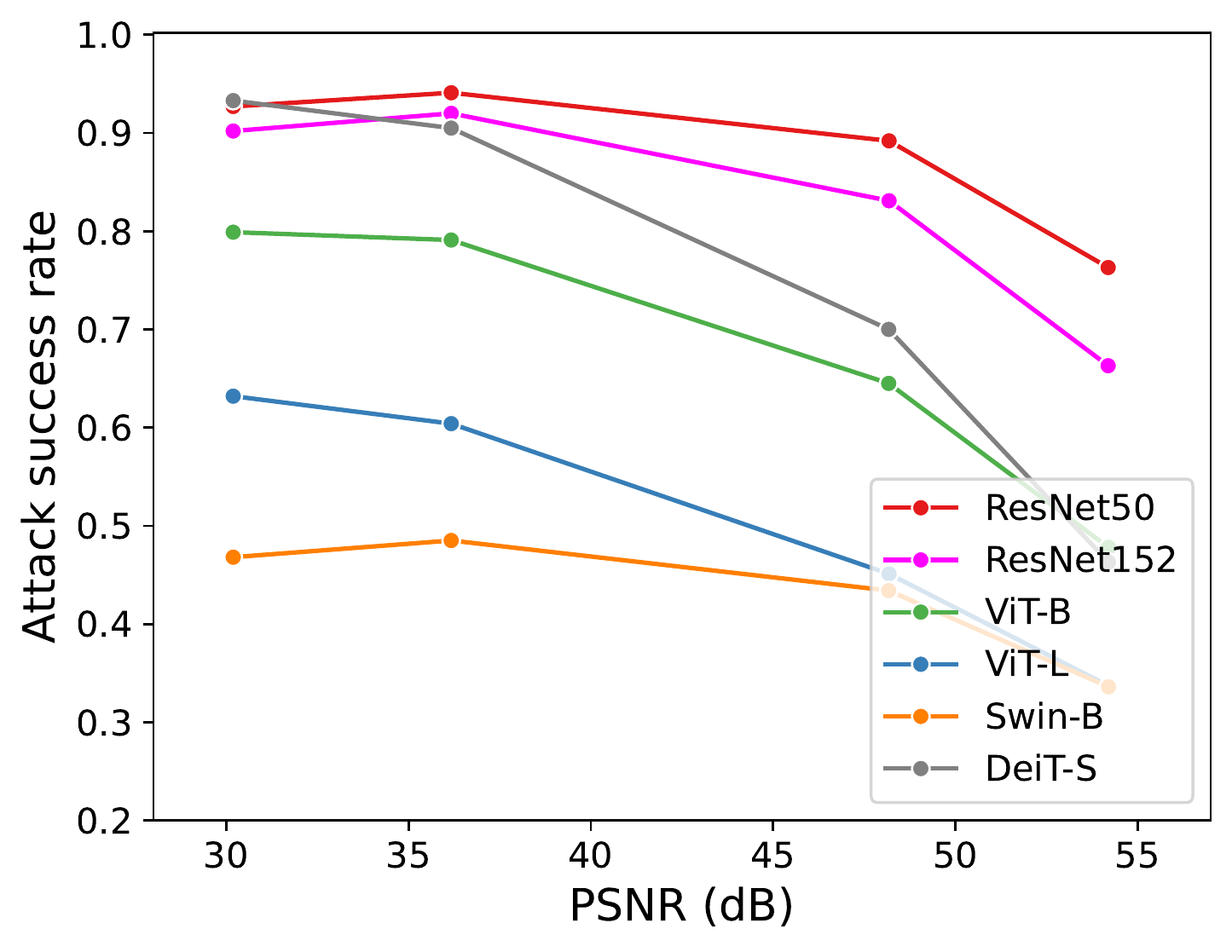}
    \caption{FGSM}
    \label{figure2:a} 
    \end{subfigure}
    \begin{subfigure}[t]{0.25\textwidth}
    \includegraphics[width=\textwidth]{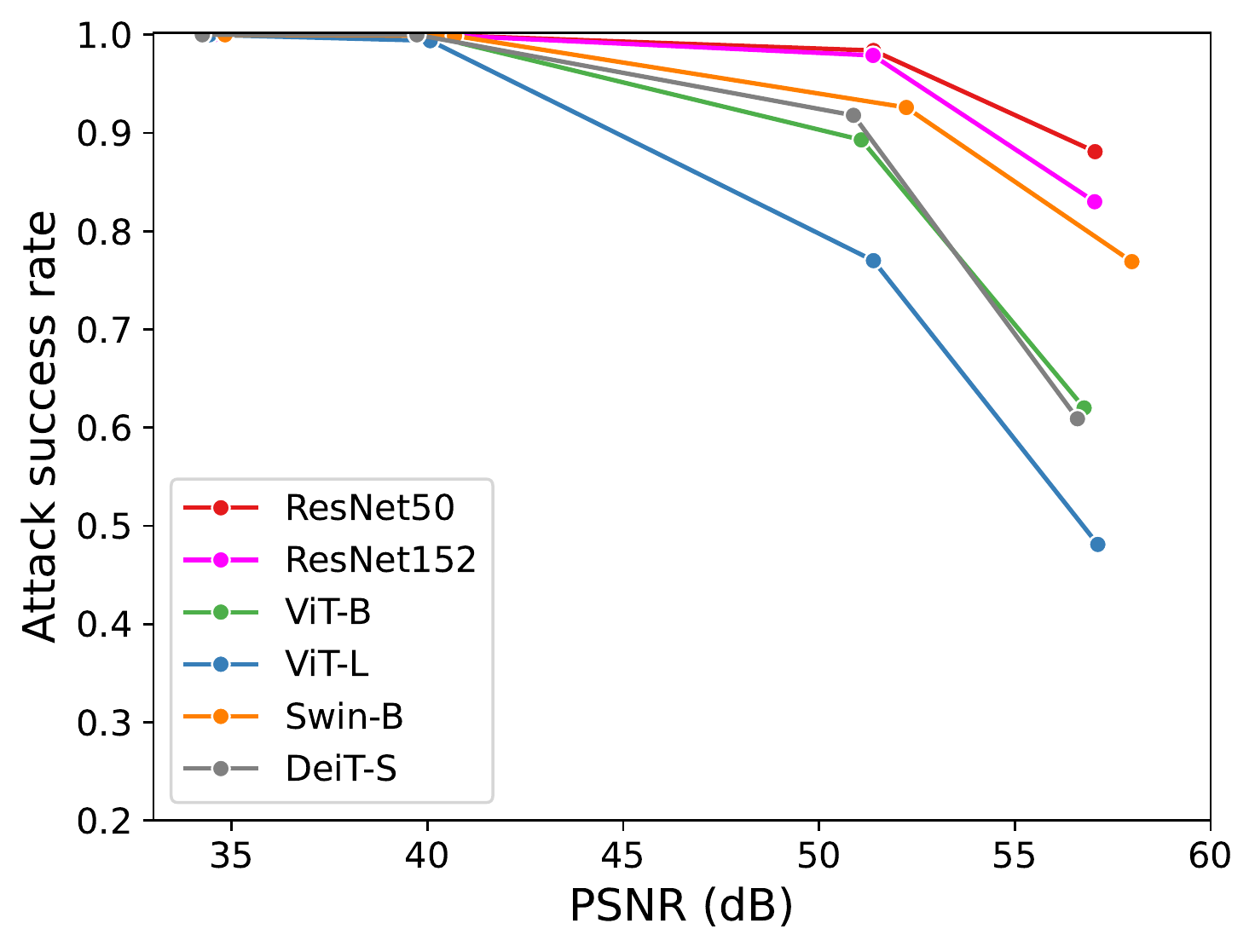}
    \caption{PGD}
    \label{figure2:b} 
    \end{subfigure}
    \begin{subfigure}[t]{0.26\textwidth}
    \includegraphics[width=\textwidth]{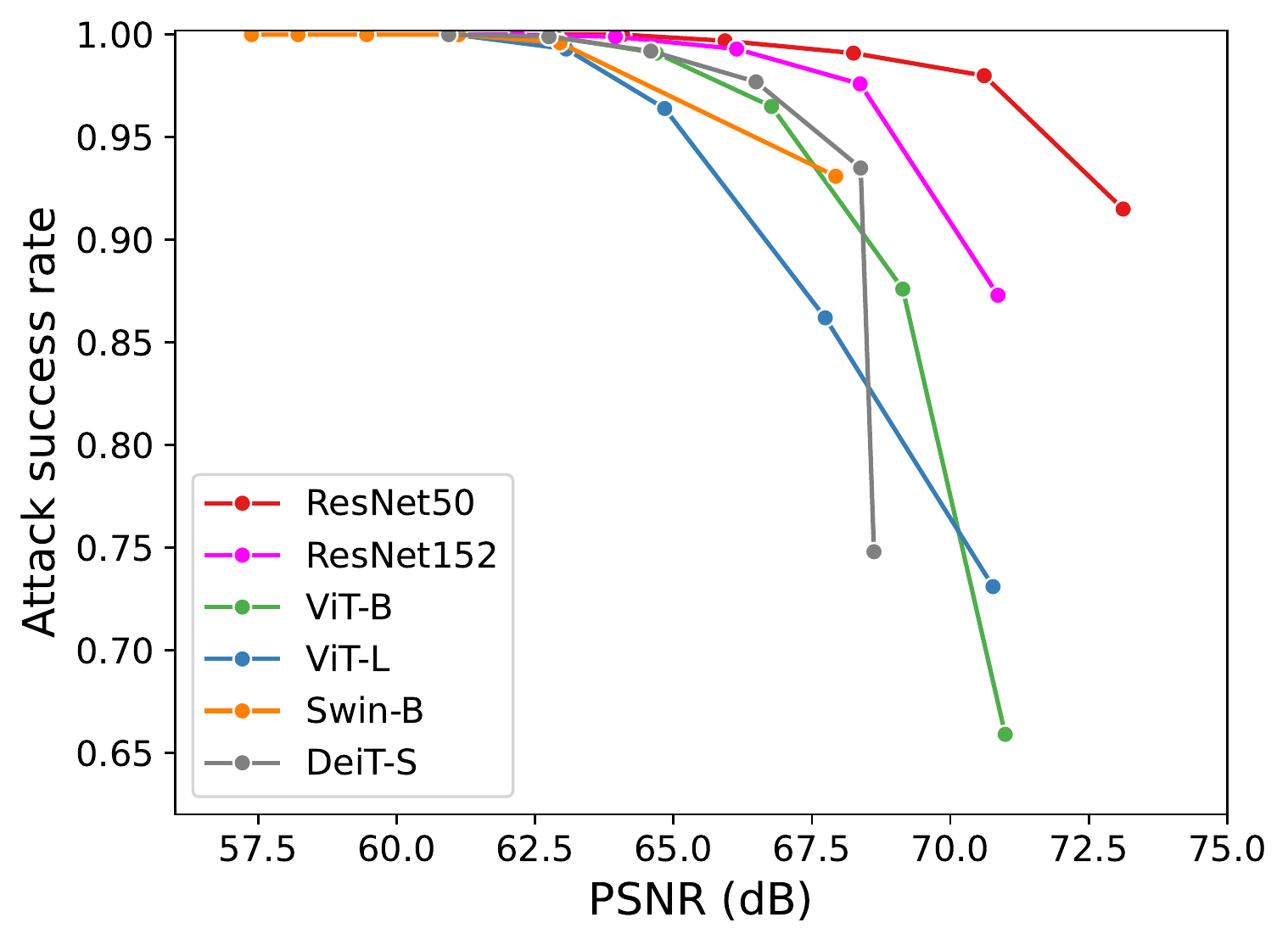}
    \caption{C\&W}
    \label{figure2:c} 
    \end{subfigure}
    \caption{Example of evaluations showing that the adversarial vulnerability is dependent on the attack strength (represented as PSNR)}
    \label{figure2}
\end{figure*}

\section{Related Works}
\label{related_works}
\subsection{Vision Transformers}
\noindent The vision Transformer (ViT) has appeared as a powerful neural architecture using the self-attention mechanism \cite{i:5}. Several variants of ViT have been also proposed. The Swin Transformer \cite{i:8} improves the efficiency over ViT using a shifted window scheme for self-attention. To resolve the issue that Transformers require a large dataset for training, the data-efficient image Transformer (DeiT) \cite{i:6} is trained through distillation from a CNN teacher. Other variants include token-to-token ViT \cite{i:23}, pyramid ViT \cite{i:24}, Transformer in Transformer \cite{a:8}, cross-covariance image Transformer \cite{a:9}, etc.

\subsection{Adversarial Attack Methods}
\noindent The goal of a typical adversarial attack is to change the classification result of a model by injecting a noise-like perturbation to the image while the perturbation is kept imperceptible in order not to be detected easily. The perturbation is usually found via gradient-based optimization. The fast gradient sign method (FGSM) \cite{i:9} uses the sign of gradient of the classification loss. The projected gradient descent (PGD) method \cite{i:12} implements a stronger attack by iteratively optimizing the perturbation. These attacks limit the $L_p$ norm of the perturbation to control the attack strength. The C\&W attack \cite{i:11} optimizes the weighted sum of the amount of perturbation and the classification loss, which is known to be one of the strongest attacks. Note that our attack formulation (Section \ref{our_attack}) is inspired by the C\&W.

Frequency-domain filtering can be used to constrain perturbations only in certain frequency regions \cite{i:3, i:21, i:22}, which are still based on image-domain attacks. The work in \cite{i:33} suggests an attack in the DCT domain for CNNs, which only drops certain frequency components via quantization. In \cite{a:14}, an attacked image is generated by swapping a frequency component (among four single-level wavelet components) of the original image with that of another image, which considers only a limited type of spectral perturbation and, furthermore, is agnostic to the target model and thus is not suitable for direct comparison of the robustness of CNNs and ViTs. Recently, direct perturbation in the frequency domain is also tried on CNNs \cite{i:27}, which is different from our method that can perturb the magnitude and phase components separately. 

\subsection{Adversarial Robustness of Transformers}
\noindent As mentioned in the introduction, two conflicting conclusions exist in the literature regarding the relative adversarial robustness of CNNs and Transformers. In one group of studies, it is claimed that Transformers are more robust than CNNs. The studies in \cite{a:6, m:2, m:3, i:25} commonly make a conclusion that Transformers are more robust against gradient-based attacks including FGSM, PGD, and C\&W because CNNs rely on high frequency information while Transformers do not. In \cite{a:15}, it is suggested that the severe nonlinearity of the input-output relationship of Transformers causes their higher robustness than CNNs. When adversarial training is considered, it is recognized that ViTs exhibit more robust generalization than CNNs \cite{i:35}. Another group of studies argues that Transformers are as vulnerable to attacks as CNNs. The work in \cite{i:1} finds that ViTs are not advantageous over ResNet in terms of robustness against various attack methods such as FGSM, PGD, and C\&W. In \cite{i:2}, it is shown that CNNs and Transformers are similarly vulnerable against various natural and adversarial perturbations. In \cite{a:2}, it is attempted to compare CNNs and Transformers on a common training setup, from which it is concluded that they have similar adversarial robustness. An attack perturbing single patches is designed in \cite{i:29} to induce vulnerability of ViTs, and similarly the patch attack \cite{i:31} is applied to Transformers in \cite{i:30}.
\raggedbottom
\subsection{Our Distinguished Contributions}
Our work is distinguished from the previous works as follows. (1) Compared to some previous works where only a limited number of models are compared \cite{a:6, m:3, i:25, i:29} or a limited range of attack strength is considered \cite{i:1, i:2, a:2, i:30}, we consider the trade-off characteristics between vulnerability and attack strength in an extensive manner for diverse CNNs and Transformers over a wide range of image quality degradation. (2) Some previous works explain different levels of vulnerability of CNNs and Transformers in terms of their reliance on different frequency components \cite{a:6, m:2, m:3, i:25}. While this conclusion is based on the results of attacks in the spatial domain, we directly impose perturbations in the spectral domain to implement frequency-selective attacks for Transformers. (3) In \cite{i:29} and \cite{i:30}, it is shown that localized perturbations on image patches effectively attack Transformers due to the patch-wise self-attention mechanism of Transformers. However, successful patch perturbations are usually visible, which is undesirable as adversarial attacks, whereas we show that Transformers can become vulnerable by spectral-domain perturbations inducing imperceptible global distortion in the images.

\section{Method}
\label{method}
\subsection{Consideration of Attack Strength}
\noindent In many popular attack methods, the attack strength can be controlled by certain parameters (e.g., $L_\infty$ norm of perturbation in FGSM, the balancing parameter between the amount of distortion and the change of the classification loss in C\&W). As an attack becomes strong, the target model becomes more vulnerable, but the image distortion becomes more perceptible. We notice that depending on the considered attack strength, the superiority of one model to another in terms of robustness may vary. 

Figure \ref{figure2} shows an example of this issue. For each trained model, we apply the FGSM, PGD, or C\&W attack to the images from the NeurIPS 2017 Adversarial Challenge \cite{m:11}. For FGSM and PGD, the $L_\infty$ norm constraint of the perturbation varies among \{0.1/255, 0.5/255, 1/255, 4/255, 8/255\}. For C\&W, the balance parameter between the amount of perturbation and the cross-entropy varies among \{1, 0.4, 0.2, 0.1, 0.05, 0.01\}. The figure shows the attack success rate (ASR) of the attacked images with respect to the attack strength. Here, the image quality of the attacked images (peak signal-to-noise ratio (PSNR) in this figure) is used to represent the attack strength. Overall, ResNet models tend to be more vulnerable than Transformers in all attacks by showing higher ASR, which is consistent with the results in \cite{a:6, m:2, m:3}. When we have a look at the details, observations demonstrating the vulnerability dependent on the attack strength can be also made. For instance, in Figure \ref{figure2:a}, DeiT-S is more robust than both ResNet models over 40 dB but they show similar vulnerability at 30-35 dB; 
in Figure \ref{figure2:b}, all models are similarly vulnerable showing almost 100\% of ASR for low PSNR values, but their vulnerability becomes different after about 40 dB; Figure \ref{figure2:c} shows a similar trend to Figure \ref{figure2:b} but in a higher PSNR range. These results demonstrate that it is important to examine a wide range of attack strength with various models in order to better understand the vulnerability of different models.

\subsection{Attack Method}
\label{our_attack}
\noindent Different from the existing works, we formulate a unified attack framework that is capable of perturbing images both spatially and spectrally. There are two key motivations behind our approach. First, the previous studies point out that CNNs tend to rely on high frequency information in images, which may be why they appear more vulnerable than Transformers \cite{a:6, m:2}. In other words, the popular attack methods injecting high frequency noise may be unfavorable to CNNs. The unified framework tries to alleviate such an inherent bias by enabling to flexibly perturb images in both spatial location-selective and frequency-selective manners. Second, we aim to analyze the mechanisms of Transformers in various viewpoints through the results of attacks applied in different domains.

The Fourier transform of an image $X$ can be written by
\begin{equation}
\mathcal{F}\{X\} = M\cdot e^{j \phi},
\label{eq:1}
\end{equation}
where $M$ and $\phi$ are the magnitude and phase spectra, respectively. The attacked image $X'$ is obtained by the combination of multiplicative magnitude perturbation\footnote{We also tried an additive magnitude perturbation but it was not optimized well because the magnitude spectrum has values over a wide range.} $\delta_\mathrm{mag}$, additive phase perturbation $\delta_\mathrm{phase}$, and additive pixel perturbation $\delta_\mathrm{pixel}$ as follows:
\begin{equation}
\tilde{X}' = \mathcal{F}^{-1}\Big\{\mathrm{clip}_{0,\infty}(M\otimes \delta_\mathrm{mag}) \cdot e^{j(\phi+\delta_\mathrm{phase})}\Big\}+\delta_\mathrm{pixel},
\label{eq:2}
\end{equation}
\begin{equation}
X'=\mathrm{clip}_{0,1}(\tilde{X}'),
\label{eq:3}
\end{equation}
where $\mathcal{F}^{-1}$ is the inverse Fourier transform, $\otimes$ is the element-wise multiplication, 
and $\mathrm{clip}_{a,b}(x)$ limits the value of each element of $x$ within $a$ and $b$. Here, we assume that the pixel values are normalized within 0 and 1.
Note that $\delta_\mathrm{mag}$ and $\delta_\mathrm{phase}$ are kept to be symmetric in order to ensure the resulting image after the inverse Fourier transform to have real-valued pixel values.

We consider attacks employing one among $\delta_\mathrm{mag}$, $\delta_\mathrm{phase}$, and $\delta_\mathrm{pixel}$, denoted as ``magnitude attack,'' ``phase attack,'' and ``pixel attack,'' respectively. It is also possible to employ two or all types of perturbation at the same time, the results of which are in Appendix.

The process to optimize the perturbations is inspired by the C\&W attack \cite{i:11}. In other words, we minimize the $L_2$ difference between the original and attacked images to keep the amount of perturbation as small as possible, while the cross-entropy (CE) loss is maximized to fool the classifier. Thus, the loss function of the unified attack framework is given by
\begin{equation}
\mathrm{Loss} = \lambda \cdot {L_2}(X', X) - \mathrm{CE}(f(X'), y),
\label{eq:4}
\end{equation}
where $\lambda$ is a parameter balancing the $L_2$ difference and CE, which controls the attack strength, $f(\cdot)$ is the classifier, and $y$ is the ground truth. This loss can be minimized by a gradient-descent approach to obtain $\delta_\mathrm{mag}$, $\delta_\mathrm{phase}$, and/or $\delta_\mathrm{pixel}$, and consequently the attacked image $X'$.

\begin{figure*}
\centering
    \begin{subfigure}[t]{0.25\textwidth}
    \includegraphics[width=\textwidth]{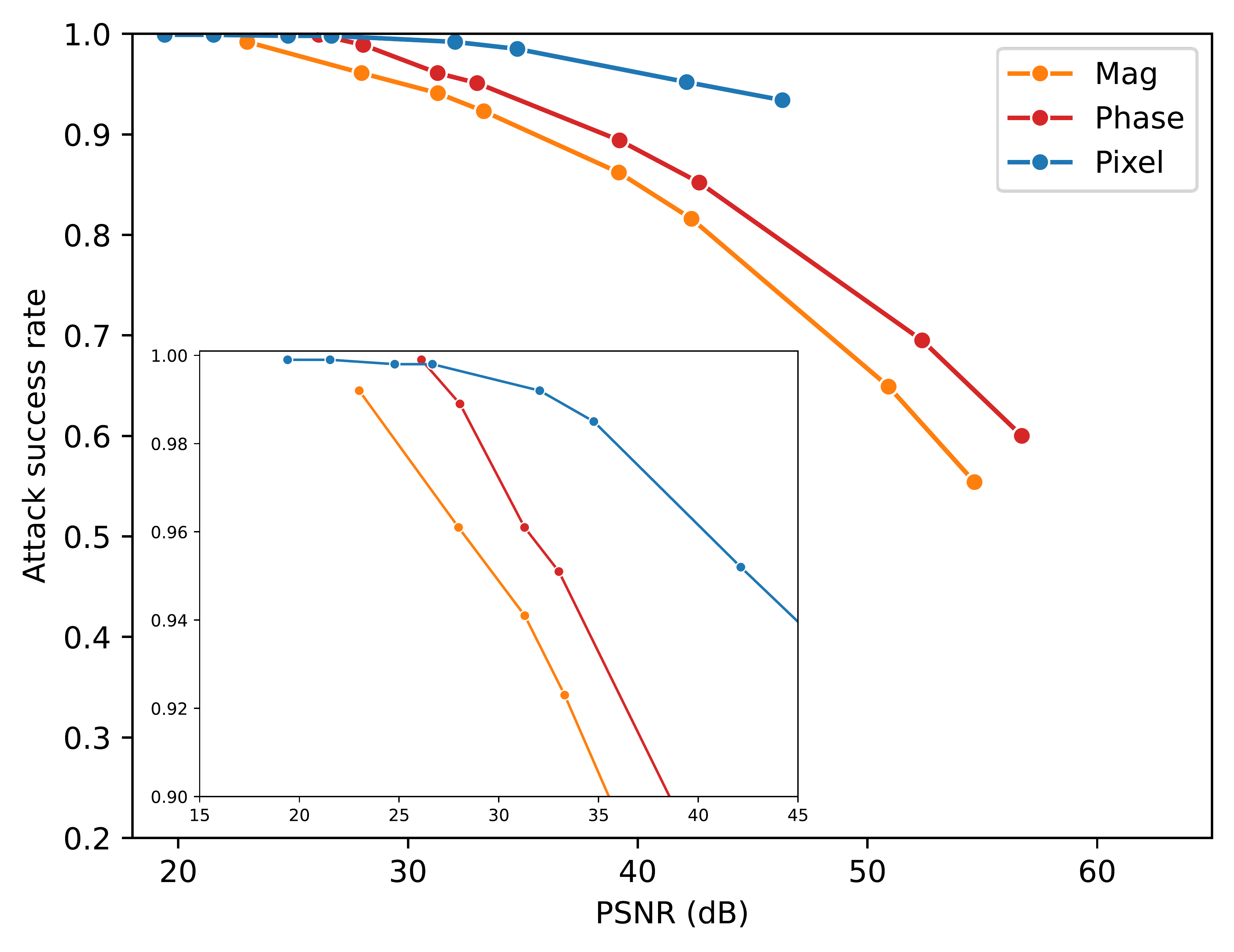}
    \caption{ResNet50}
    \label{figure3:a}
    \end{subfigure}
    \begin{subfigure}[t]{0.25\textwidth}
    \includegraphics[width=\textwidth]{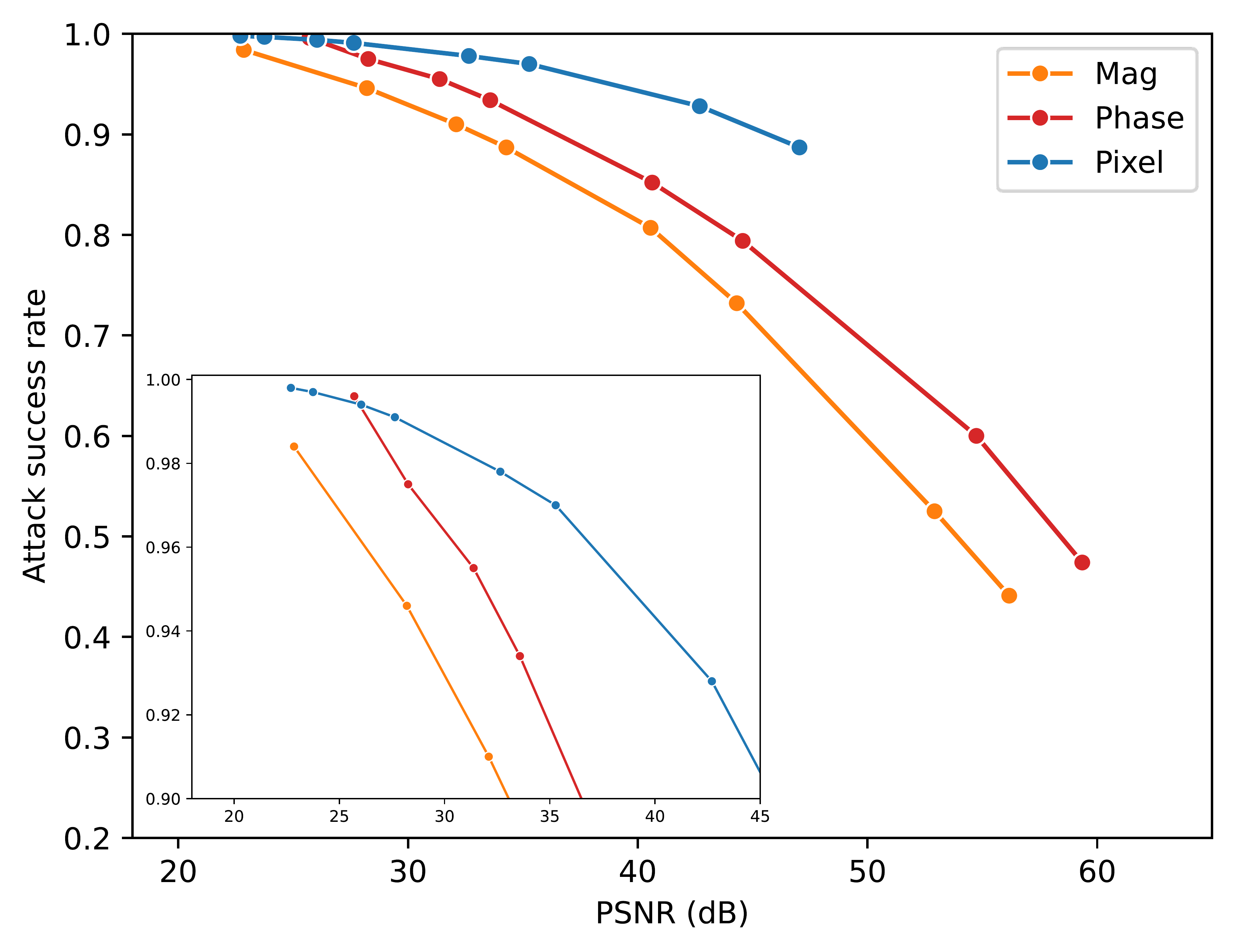}
    \caption{ResNet152}
    \label{figure3:b}
    \end{subfigure}
    \begin{subfigure}[t]{0.25\textwidth}
    \includegraphics[width=\textwidth]{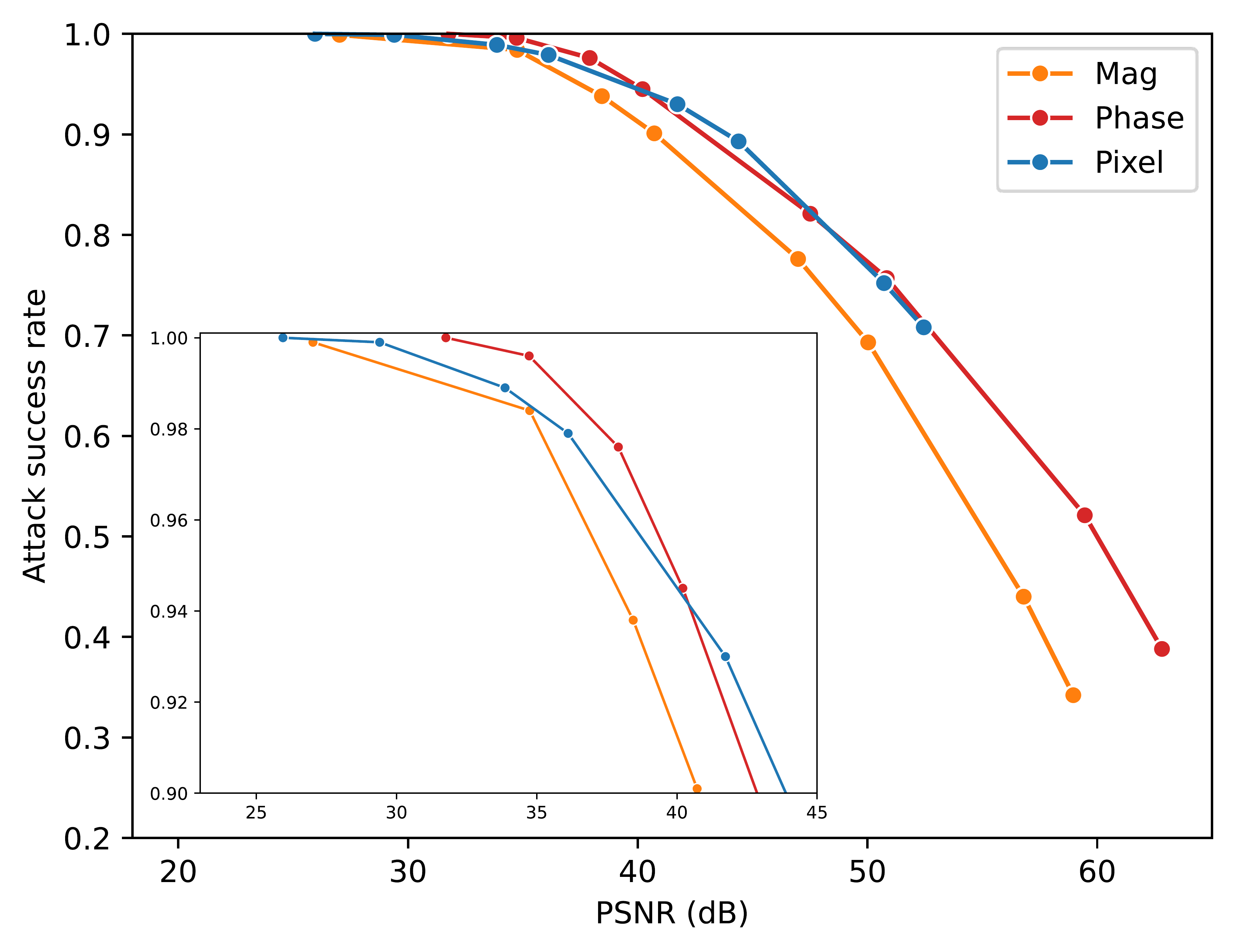}
    \caption{ViT-B}
    \label{figure3:c}
    \end{subfigure}
    \begin{subfigure}[t]{0.25\textwidth}
    \includegraphics[width=\textwidth]{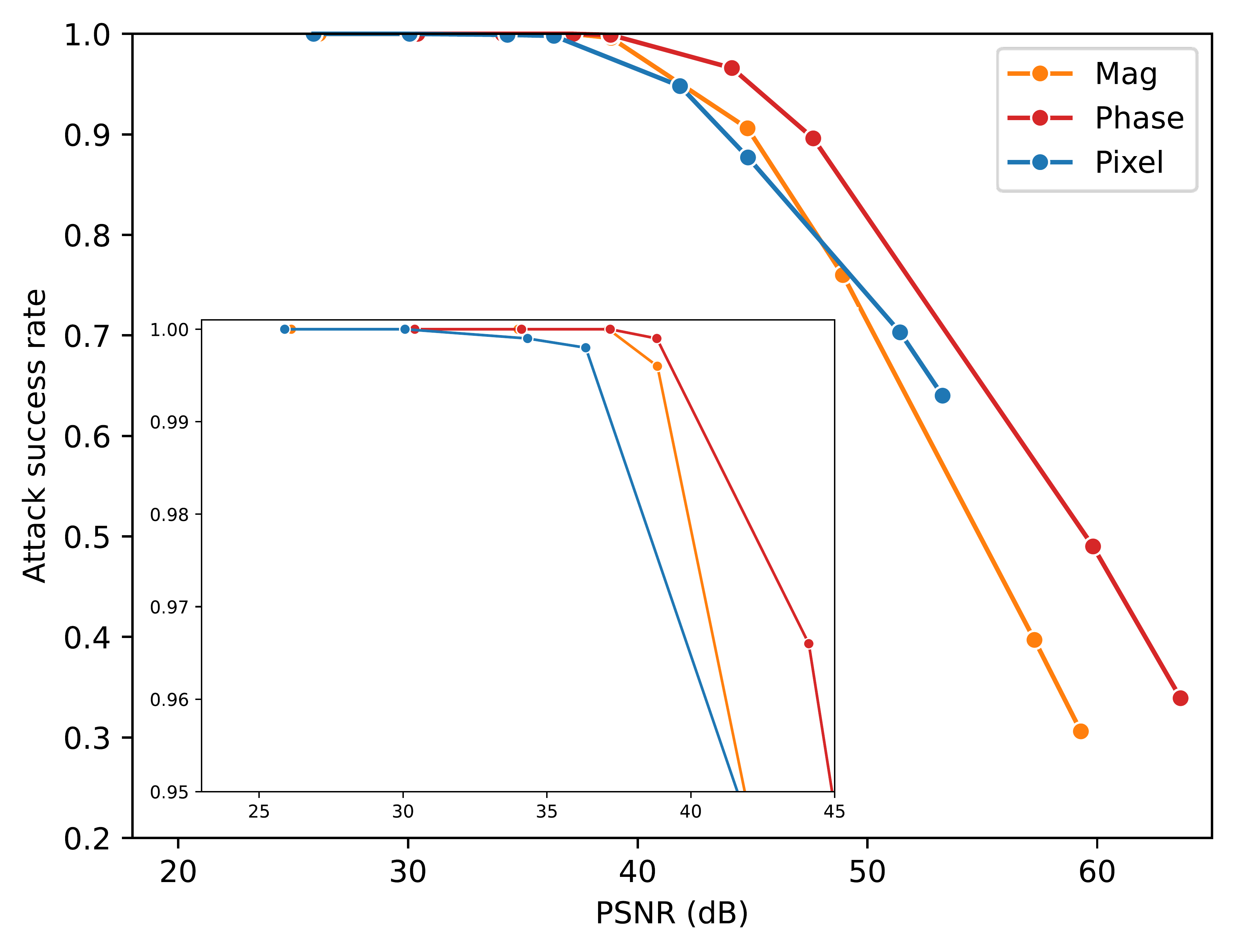}
    \caption{ViT-B-1k}
    \label{figure3:d}
    \end{subfigure}
    \begin{subfigure}[t]{0.25\textwidth}
    \includegraphics[width=\textwidth]{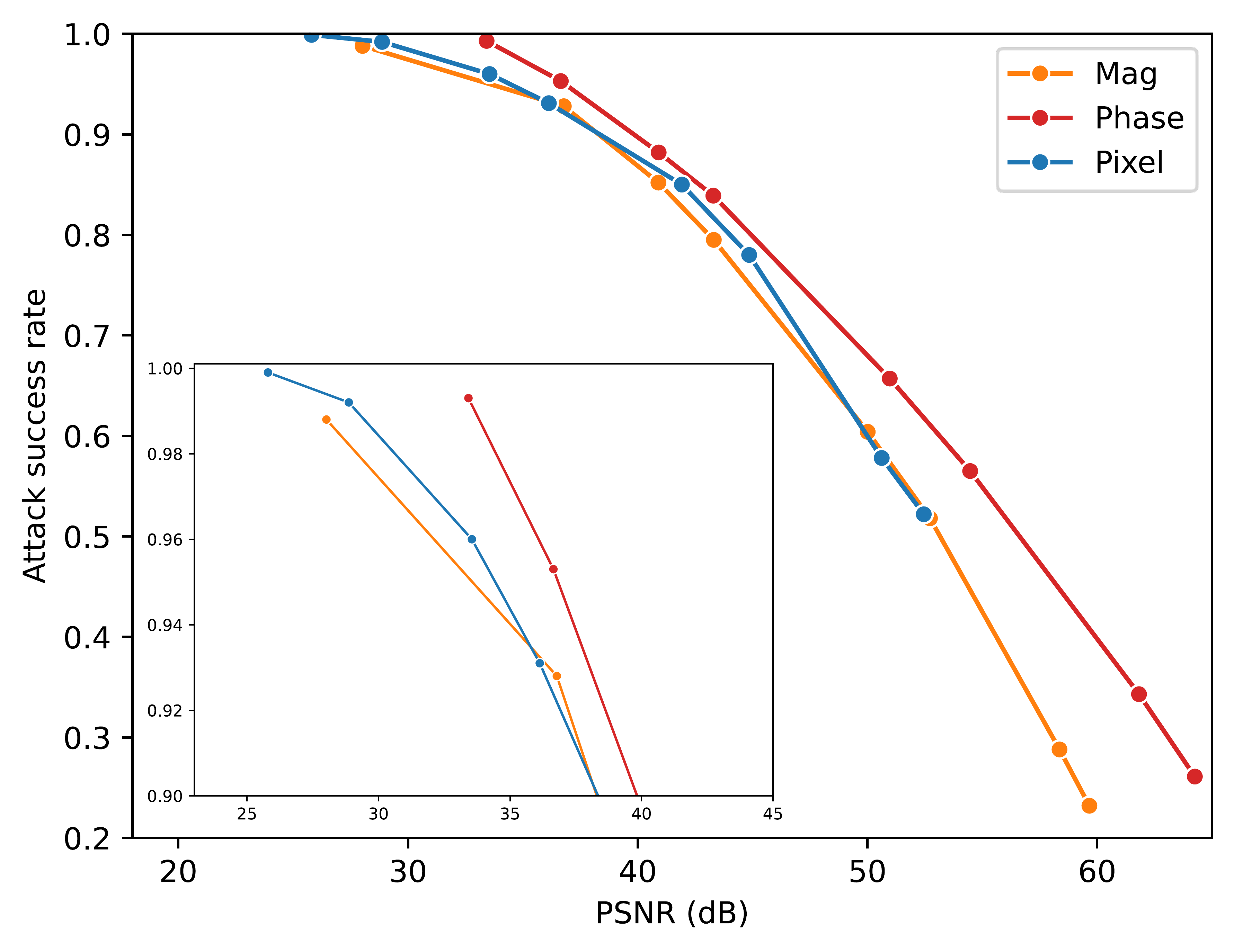}
    \caption{ViT-L}
    \label{figure3:e}
    \end{subfigure}
    \begin{subfigure}[t]{0.25\textwidth}
    \includegraphics[width=\textwidth]{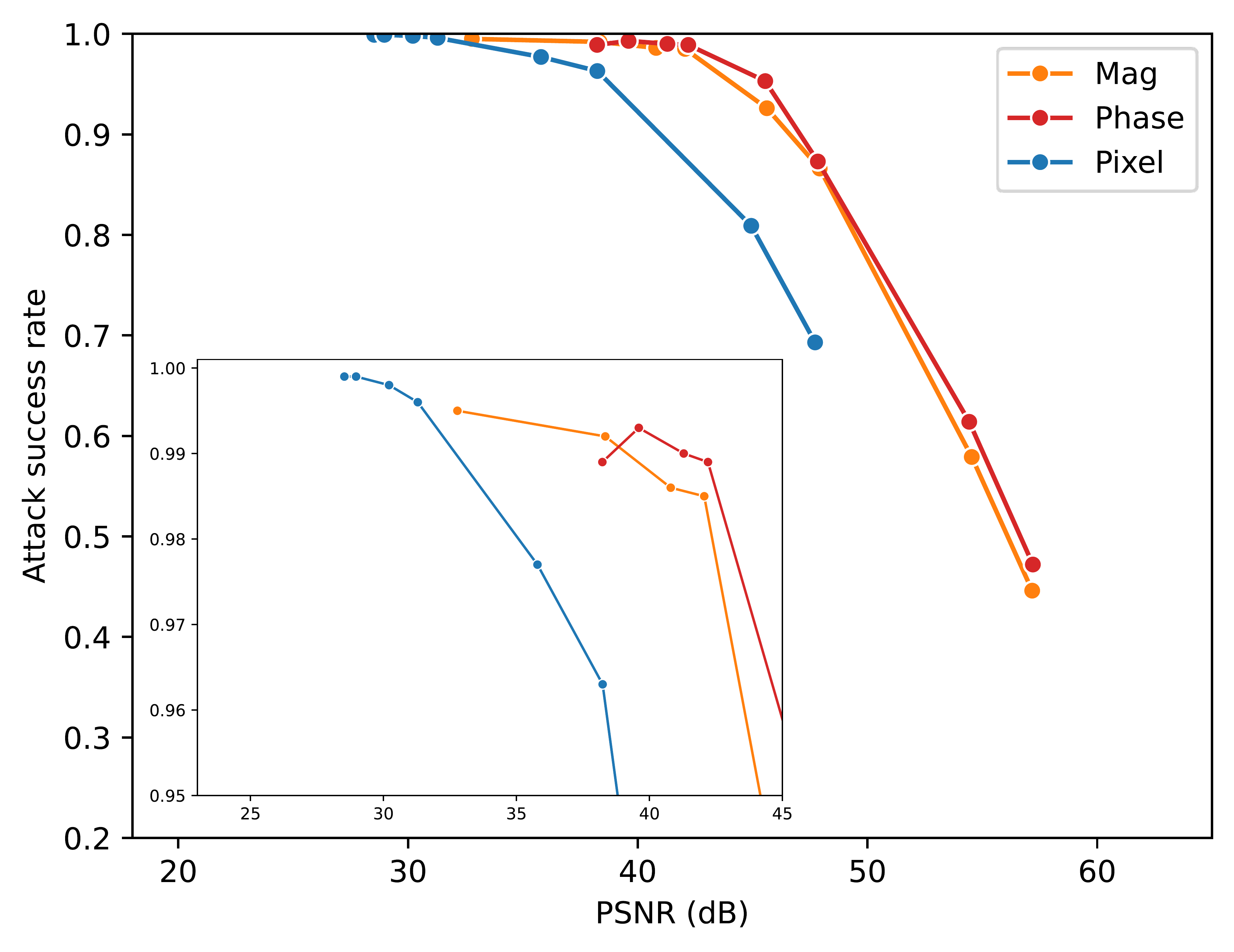}
    \caption{Swin-B}
    \label{figure3:f}
    \end{subfigure}
    \begin{subfigure}[t]{0.25\textwidth}
    \includegraphics[width=\textwidth]{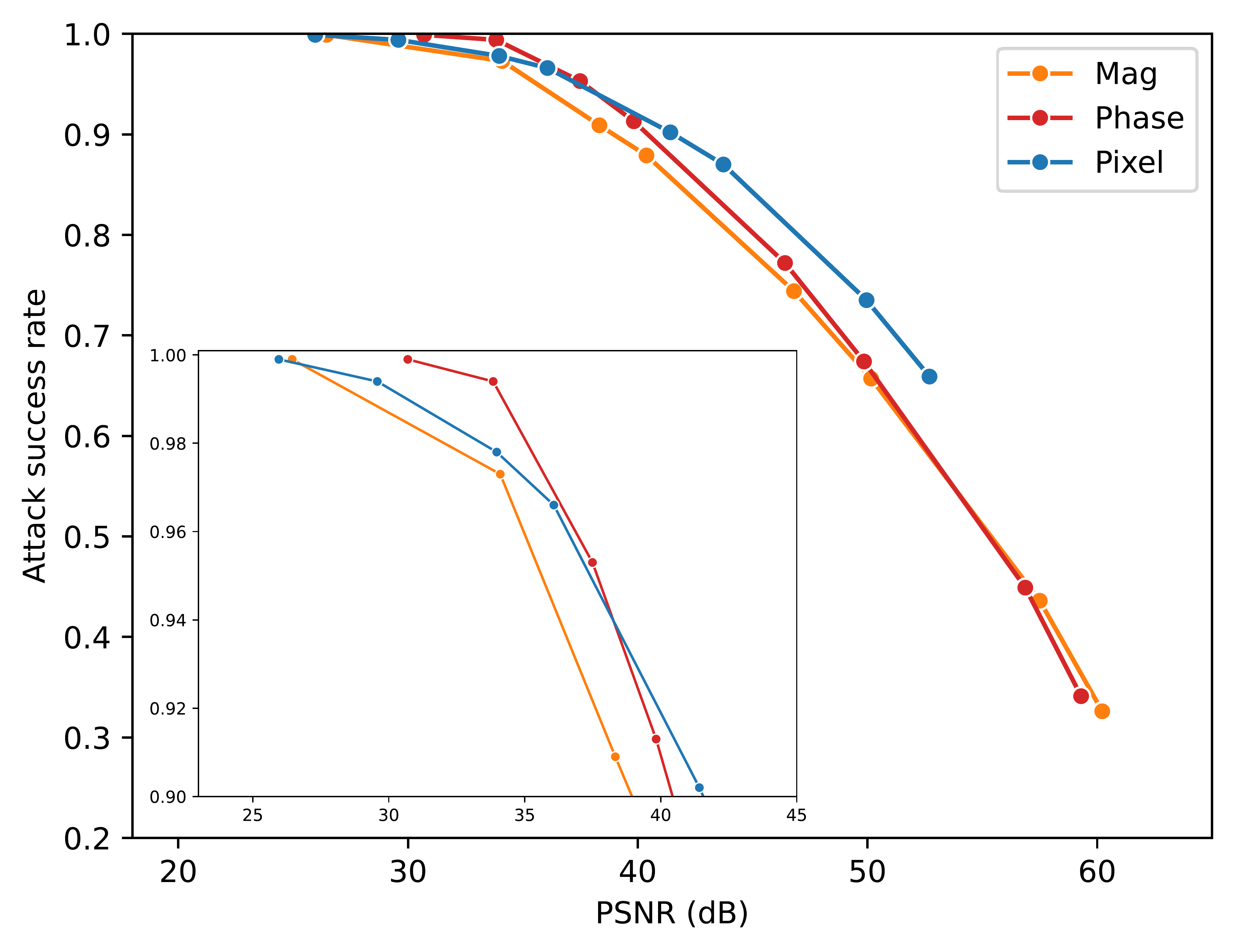}
    \caption{DeiT-S}
    \label{figure3:g}
    \end{subfigure}
    \begin{subfigure}[t]{0.25\textwidth}
    \includegraphics[width=\textwidth]{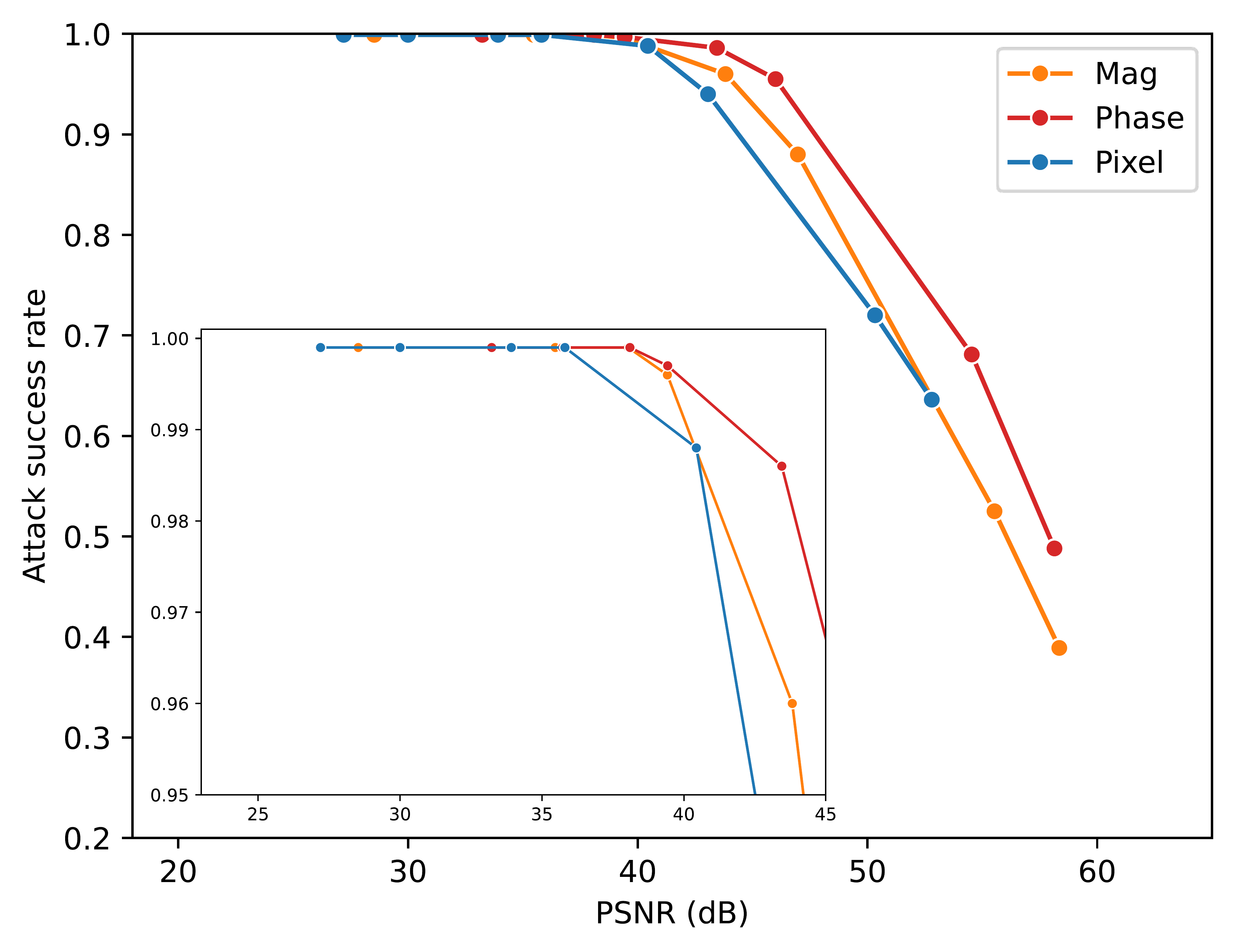}
    \caption{DeiT-S with no distillation}
    \label{figure3:h}
    \end{subfigure}

    \caption{Comparison of different attacks for each model. The range for high attack success rates is enlarged for better visualization.}
    \label{figure3}
\end{figure*}

\section{Experiments}
\label{analysis}
\subsection{Setup}
\noindent We aim to benchmark the adversarial robustness of the pre-trained models that serve as off-the-shelf solutions in general image classification applications. We consider ResNet50, ResNet152 as CNNs, and ViT-B/16, ViT-L/16, DeiT-S, and Swin-B as Transformers. Here, S, B and L mean small, base, and large, and /16 means the patch size. ResNet50 and ResNet152 are from the torchvision models \cite{a:4} trained on ImageNet-1k \cite{a:11}. ViT-B, ViT-L, DeiT-S, and Swin-B are from the timm module \cite{m:10}, which are pre-trained on ImageNet-21k \cite{i:26} and finetuned on ImageNet-1k. ViT trained on ImageNet-1k and DeiT-S without distillation are also considered.

To evaluate the adversarial robustness of the models, the image dataset from the NeurIPS 2017 Adversarial Challenge \cite{m:11} is used.

To obtain perturbations by minimizing the loss in Eq. (\ref{eq:4}), we use the Adam optimizer with a fixed learning rate of $5\times 10^{-3}$ and a weight decay parameter of $5 \times 10^{-6}$. The maximum number of iterations is set to 1000. We also set a termination condition that the optimization stops when the loss does not improve for five consecutive iterations. We vary the value of $\lambda$ as $\{1, 10^3, 5 \times 10^3, 10^4, 5 \times 10^4, 10^5, 5 \times 10^5, 10^6 \}$ to control the attack strength.

We employ image quality metrics to enable representation of the attack strength commonly across different attacks introduced in the previous section, including PSNR, multi-scale structural similarity index measure (MS-SSIM) \cite{a:12}, mean deviation similarity index (MDSI) \cite{a:5}, and learned perceptual image patch similarity (LPIPS) \cite{i:18}. Here, the results using PSNR are shown; those using the other metrics are provided in Appendix, which show similar trends to those using PSNR.

\begin{figure*}
\centering
  \begin{subfigure}[t]{0.3\textwidth}
  \includegraphics[width=\textwidth]{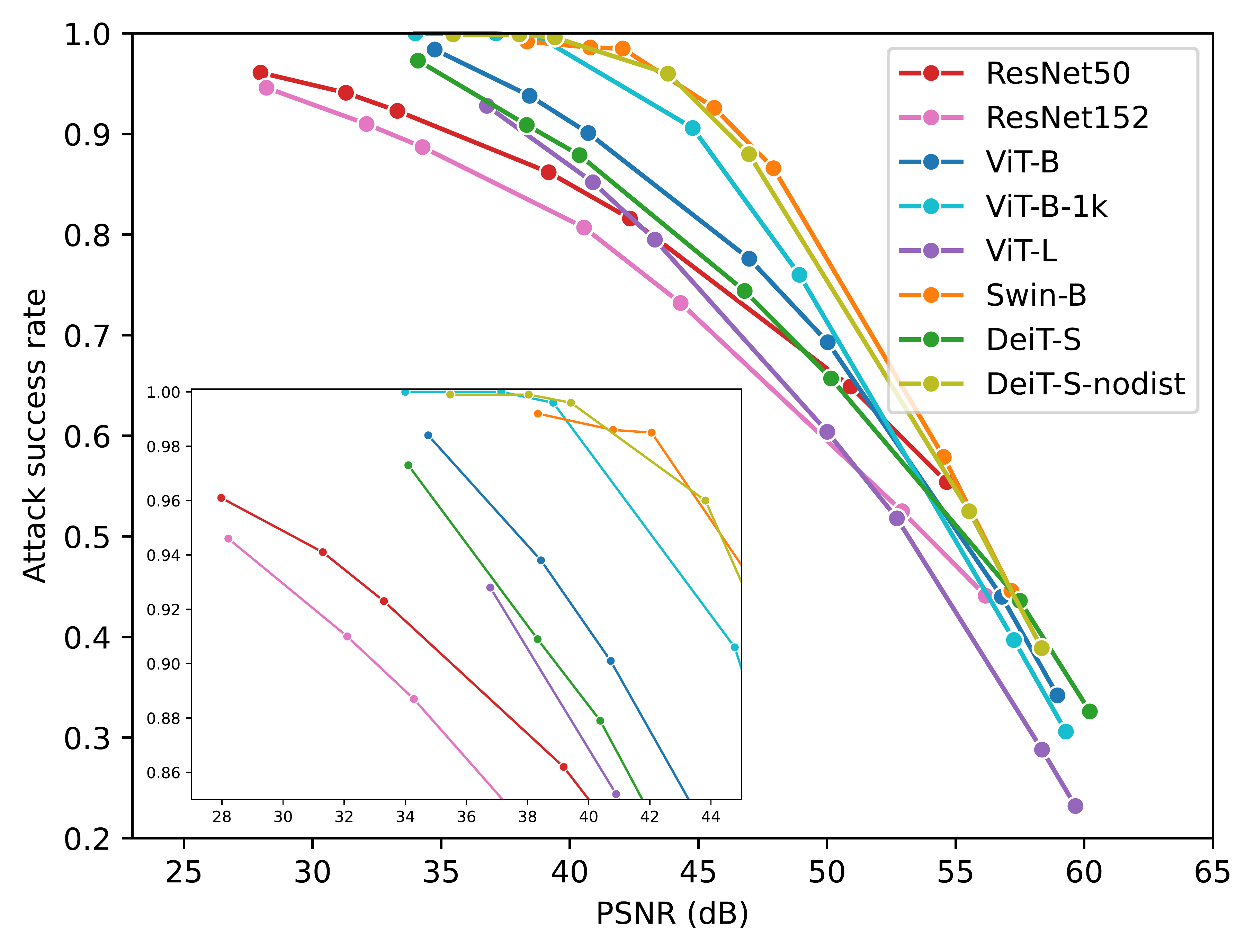}
  \caption{Magnitude attack}
  \label{figure4:a}
  \end{subfigure}
  \hfill
  \begin{subfigure}[t]{0.3\textwidth}
    \includegraphics[width=\textwidth]{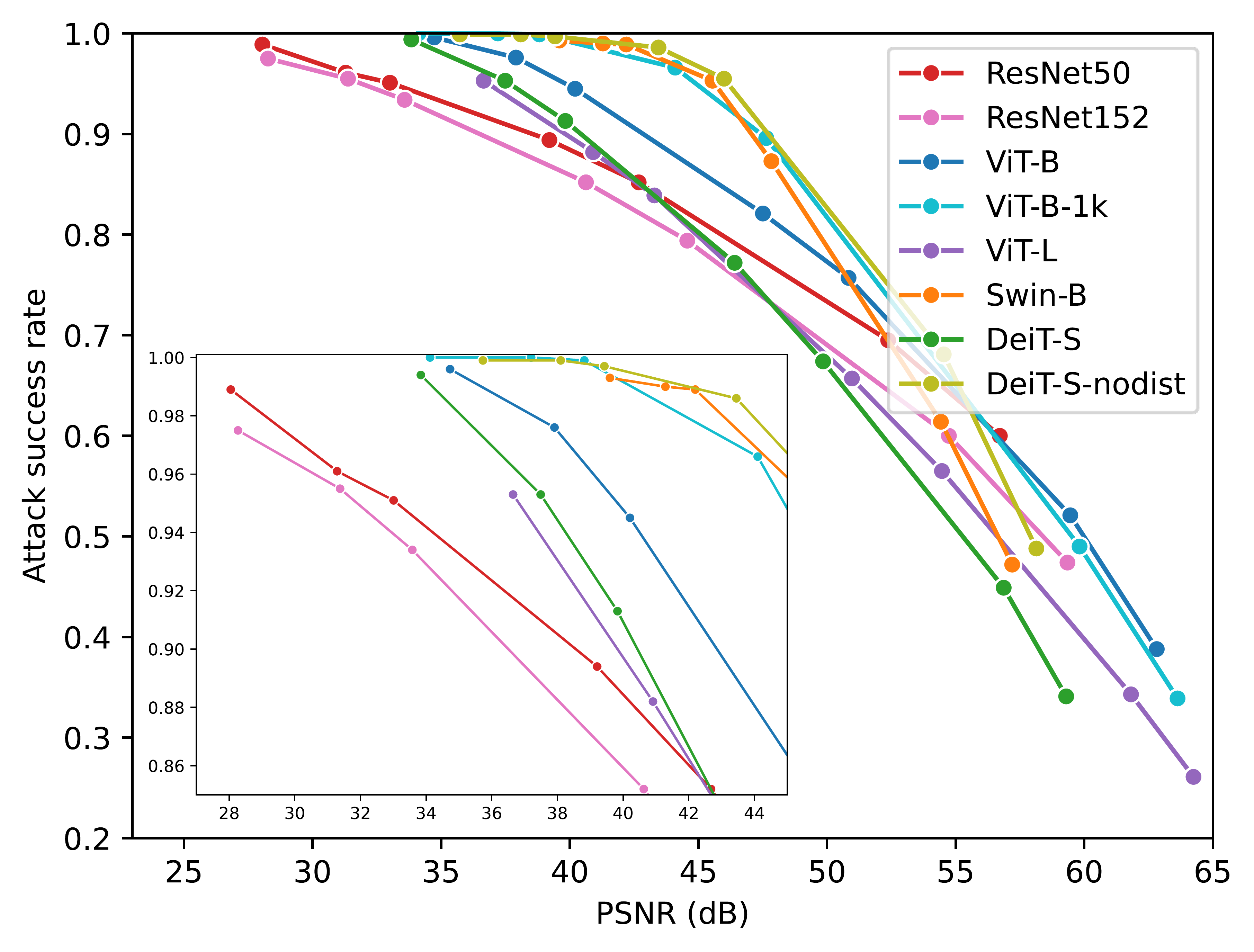}
  \caption{Phase attack}
  \label{figure4:b}
  \end{subfigure}
  \hfill
  \begin{subfigure}[t]{0.3\textwidth}
    \includegraphics[width=\textwidth]{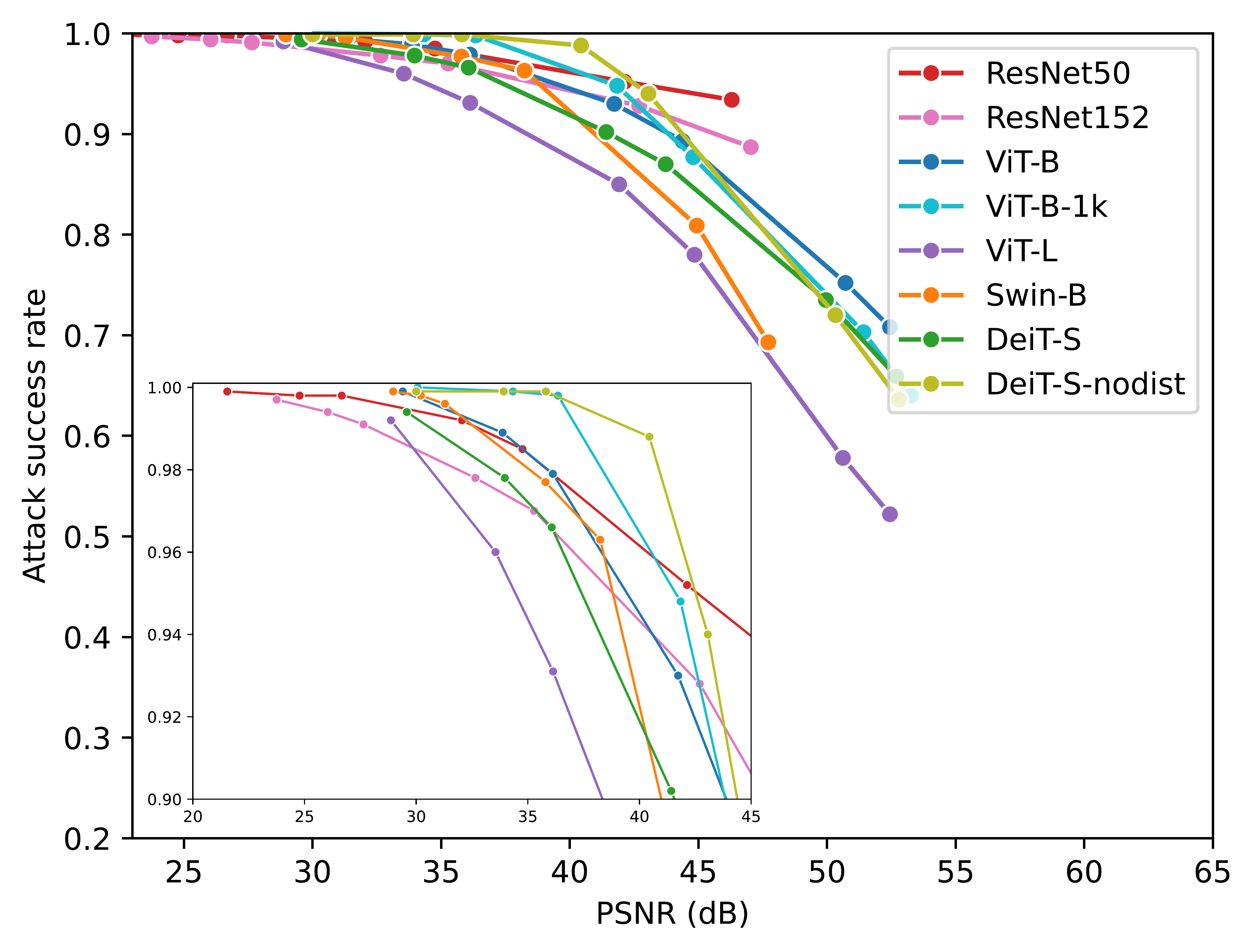}
  \caption{Pixel attack}
  \label{figure4:c}
  \end{subfigure}
\caption{Comparison of different models for each attack type. The range for high attack success rates is enlarged for better visualization.}
\label{figure4}
\end{figure*}

\subsection{Results}
\label{results}
\noindent We present the results in two perspectives: (1) comparison of different attacks for each model, and (2) comparison of different models for each attack type.
\subsubsection{Comparison of Attacks}
We first compare various attacks implemented using our unified attack framework. Figure \ref{figure3} shows the results in terms of ASR with respect to PSNR. All the attacks can achieve (almost) 100\% of ASR at the lower extremes of PSNR for all models. However, their relative effectiveness varies depending on the model. For ResNet50 and ResNet152, the pixel attack appears to be the strongest, whereas the phase attack is the strongest for ViTs and Swin-B. This shows that the flexible frequency domain attack is able to overcome the limitation of the pixel attack primarily perturbing high frequency information, and becomes a potent tool for attacking Transformers. We assume that the vulnerability of DeiT-S to the pixel attack is attributed to its training method, which involves distillation from a CNN teacher. This assumption is supported by the observation that DeiT-S without distillation is also more vulnerable to the phase attack, which is consistent with the vulnerability of ViTs and Swin-B. 

It is observed that the phase attack is mostly stronger than the magnitude attack. Further analysis on this is presented in Section \ref{phase_mag}. Attention maps before and after attacks are also compared in Appendix.

\subsubsection{Comparison of Vulnerability of Models}
In Figure \ref{figure4}, ASR of the models is compared with respect to PSNR for each attack type. Notably, when ASR is relatively high, Transformers are either equally or more robust than ResNet models under the pixel attack but are more vulnerable to the magnitude and phase attacks. Again, this is attributed to the flexible perturbations in the frequency domain, which will be further analyzed later. At PSNR over 45-50 dB, where ASR is low, the observed trends do not hold anymore, i.e., some Transformers (ViTs and DeiT-S) become similarly robust to ResNet models under the magnitude and phase attacks, and ResNet models become more vulnerable to all Transformers under the pixel attack. When Swin-B is compared to ViT-B and ViT-L, the former shows higher vulnerability than the latter for the magnitude and phase attacks. When the model size is concerned, larger models (ViT-L and ResNet152) are more robust than smaller models (ViT-B and ResNet50) under all attacks.
While \cite{i:2} also observed a similar trend using pixel-domain attacks (FGSM and PGD), we find that the same also holds for the attacks in the spectral domain.

We also compare ViTs trained in different environments, i.e. ViT-B and ViT-B-1k. It is observed that ViT-B is more robust than ViT-B-1k below 45 dB, but becomes more vulnerable when PSNR increases, particularly for the pixel attack. It is worth noting that \cite{i:2} also pointed out the advantage of training with a larger dataset to enhance robustness; we additionally find that the benefit of a larger dataset is even more prominent for the magnitude and phase attacks than for the pixel attack, but the benefit disappears when the amount of perturbation is small.

DeiT-S behaves more like ResNet than the other Transformers due to the distillation using CNN. In Appendix, the models pre-trained on the same dataset (i.e., ImageNet-1k) are compared, where similar trends to the above results are observed.

\subsection{Analysis}
\noindent We further investigate the particular vulnerability of Transformers to the phase attack.

\subsubsection{Frequency Analysis of Perturbations}
In order to investigate the effect of the phase attack on the spectral characteristics of images, we apply the Fourier transform to the difference between the original and attacked images, and analyze the magnitudes in different frequency regions, which are defined as illustrated in Figure \ref{figure5}(b). Figures \ref{figure5}(c) to \ref{figure5}(f) show the attacked image, distortion in the pixel domain, and magnitude distribution. The averaged results over images are shown in Appendix. In the case of ResNet50, the distortion is concentrated on the high frequency regions, whereas low frequency regions are mainly distorted in the other models. Since CNNs and Transformers rely more on high and low frequency information, respectively \cite{i:13, a:6, i:15, m:6}, the attack effectively injects perturbations in such vulnerable frequency regions. Consequently, the distortion pattern significantly differs according to those properties.

\subsubsection{Frequency-Restricted Attacks}
\begin{figure}[t]
    \centering
    \includegraphics[width=0.46\textwidth]{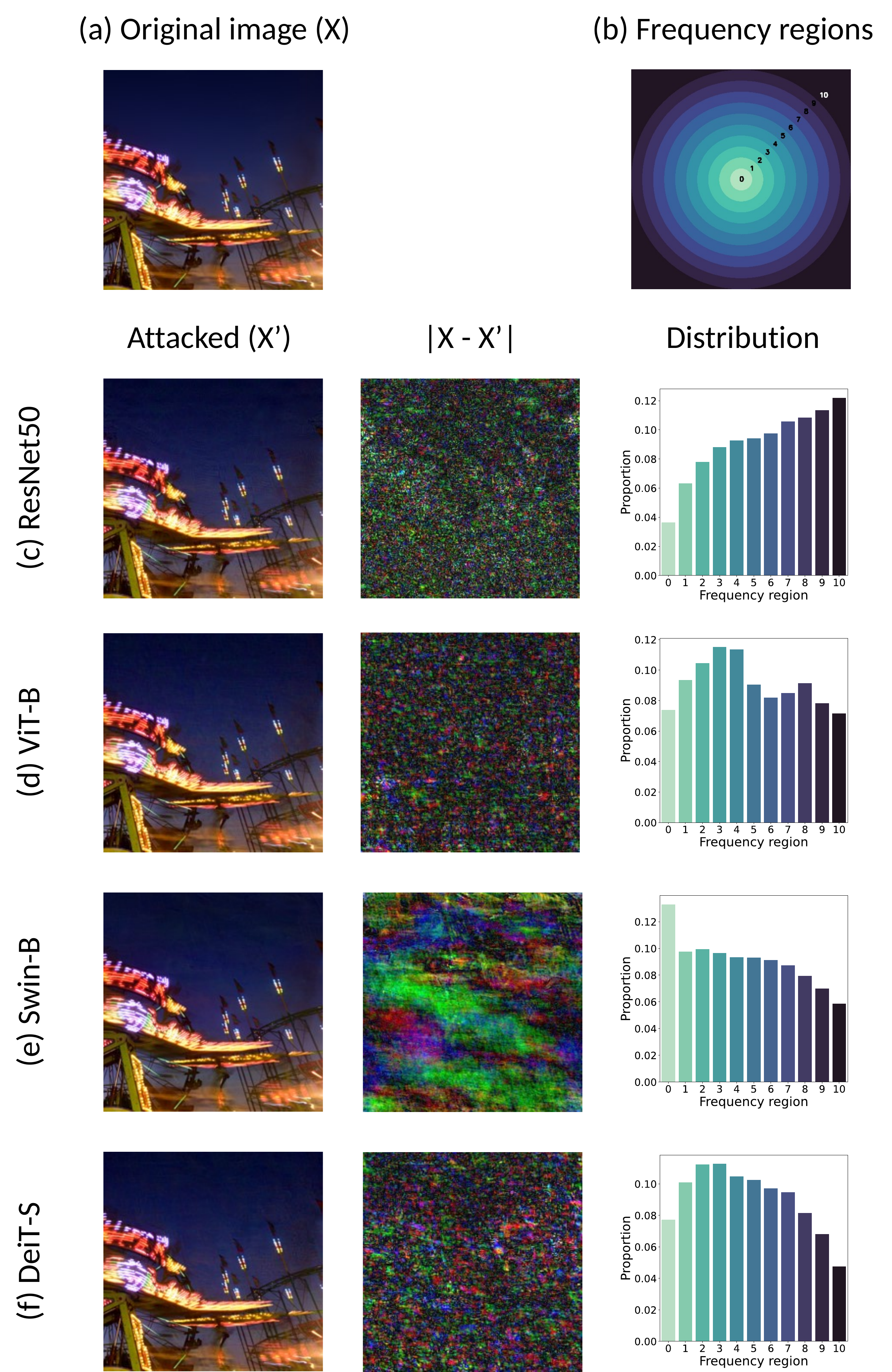}
    \caption{Example of the attacked image under the phase attack, distortion in the pixel domain (magnified by $\times$20), and distribution of the distortion over different frequency regions.} 
    \label{figure5}
\end{figure}

\begin{figure}[t]
\begin{subfigure}[t]{0.2\textwidth}
\includegraphics[width=\textwidth]{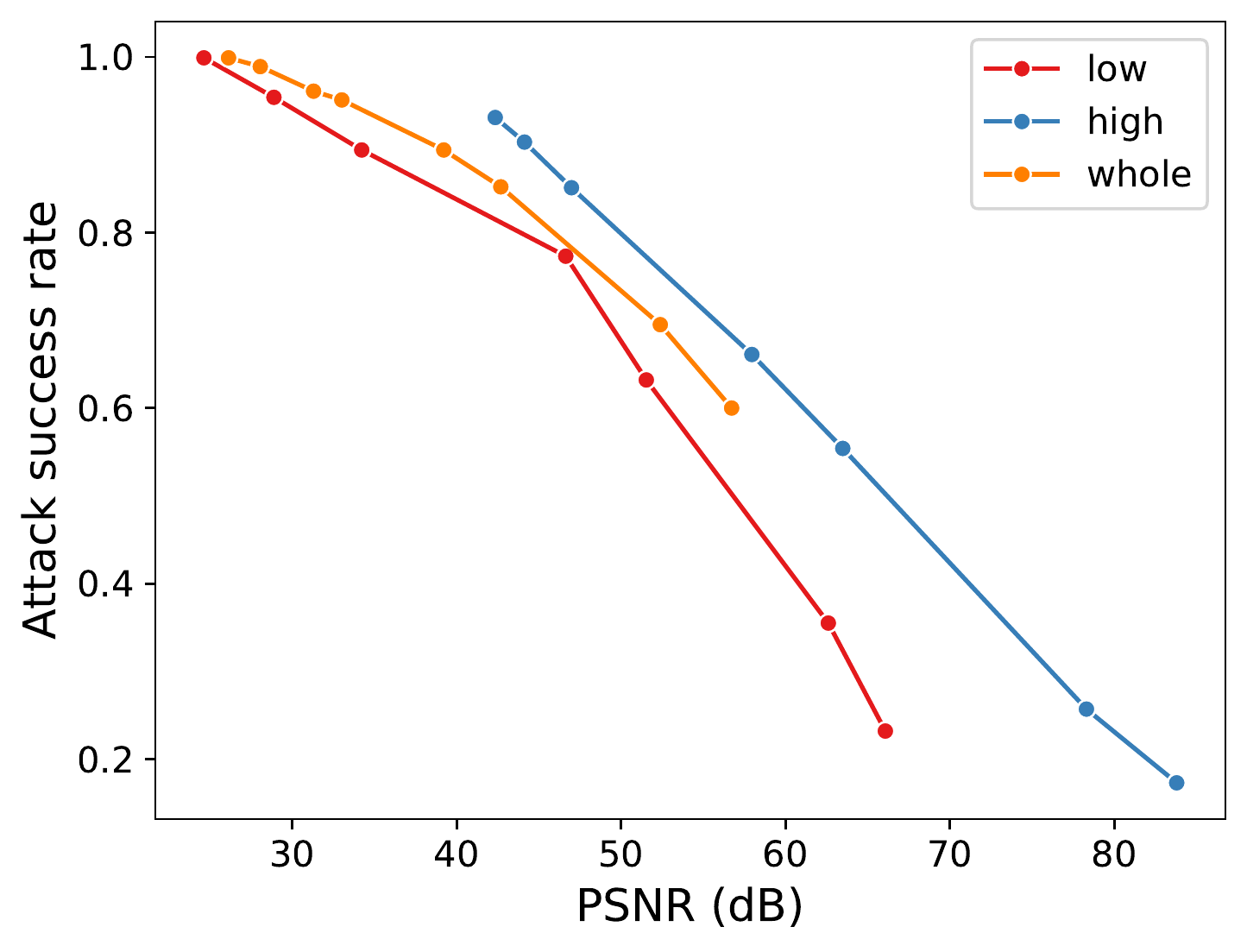}
\caption{ResNet50}
\label{figure6:a}
\end{subfigure}
\hfill
\begin{subfigure}[t]{0.2\textwidth}
\includegraphics[width=\textwidth]{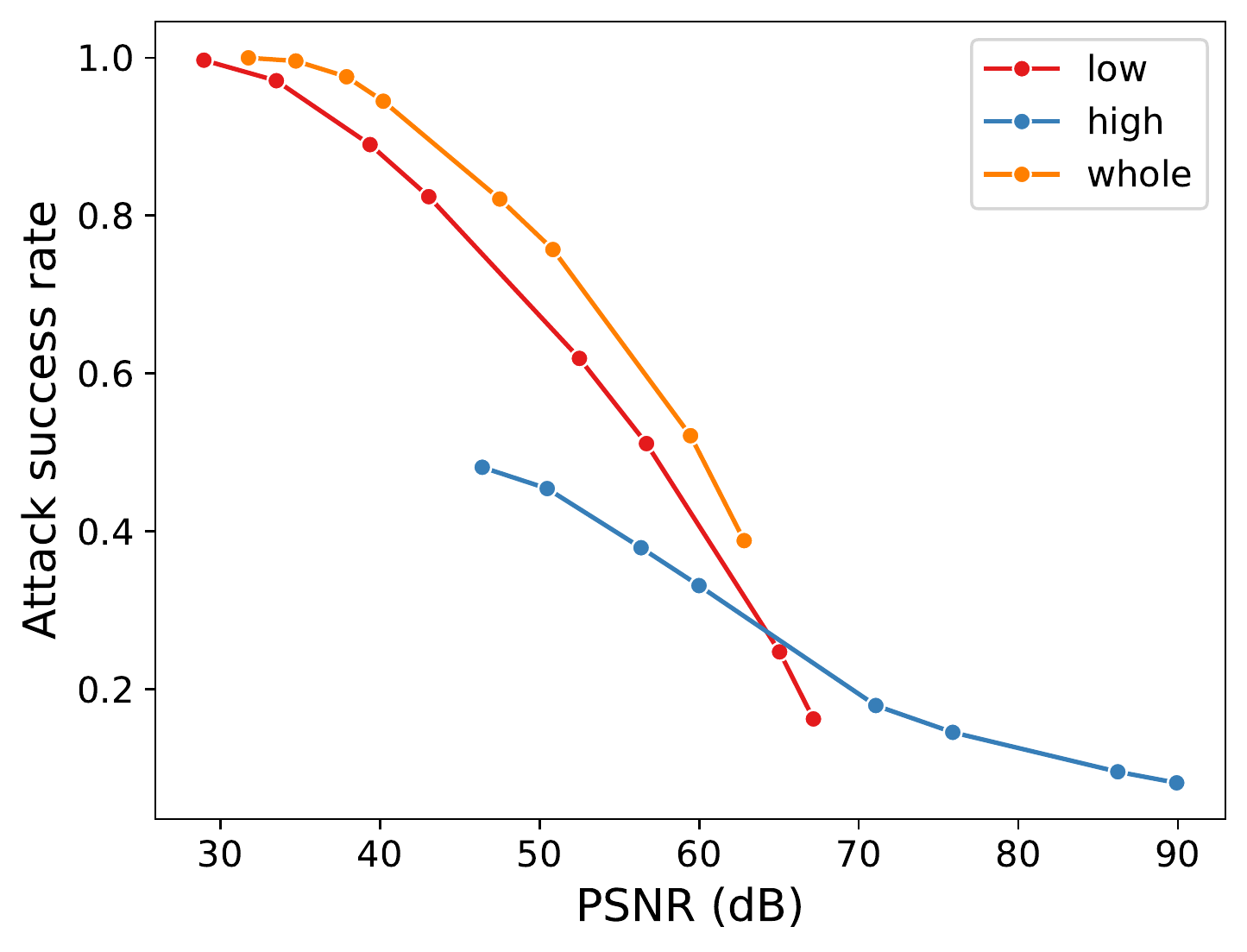}
\caption{ViT-B}
\label{figure6:b}
\end{subfigure}
\caption{Results of the phase attack when the perturbation is restricted to reside only in the low or high frequency band. The case without restriction is also shown as `whole.'}
\label{figure6}
\end{figure}

We examine the case where the perturbation is applied to a limited frequency band. Figure \ref{figure6} compares the phase attack when the phase perturbation is restricted to only the low frequency band (regions 1 and 2 in Figure \ref{figure5}(b)) or the high frequency band (region 10 in Figure \ref{figure5}(b)). Perturbing only the high frequency band is more effective than perturbing the whole band for ResNet50, as it is particularly vulnerable to high frequency perturbations. However, for ViT-B, perturbing only the high frequency band is the least effective and the case without restriction (i.e., `whole') implements the strongest attack because effective perturbations need to be applied to low-intermediate frequency regions as shown in Figure \ref{figure5}(d).

\subsubsection{Linearity of Models and Attacks}
We further analyze the vulnerability of Transformers to the phase attack in the viewpoint of the linearity of models and attacks. A noticeable difference between the attacks, as shown in Eq.~\ref{eq:2}, is whether the input is \textit{linearly} perturbed or not.
If a classifier $f$ is linear, its input-output relationship will be linear, i.e.,
\begin{equation}
    g(X + \epsilon \cdot \delta) = g(X) + \epsilon \cdot g(\delta),
    \label{eq:linearity}
\end{equation}
where $X$ is an input image, $\delta$ is a perturbation, $\epsilon$ is a scalar, and $g$ is the output at the penultimate layer of $f$ (i.e., output feature).
For such a classifier, $\delta$ can effectively move $g(X + \epsilon \cdot \delta)$ along the determined adversarial direction $g(\delta)$ in the feature space as intended.
In contrast, if $f$ is nonlinear, $g(X + \epsilon \cdot \delta)$ will not directly follow the direction of $g(\delta)$ in the feature space, which will weaken the attack.
This explanation coincides with \cite{i:9}, which showed that adversarial examples are a result of models being too linear.

To experimentally examine the linearity of models under different attacks, we use the method in~\cite{a:15}.
Consider $X_{\epsilon} = X + \epsilon \cdot \delta$, where $0 \leq \epsilon \leq 1$.
As $\epsilon$ increases, $X$ moves on the straight line along the direction determined by $\delta$ in the input space.
For a chosen $\epsilon$, we examine the following quantity:
\begin{align}
    \theta_{\epsilon} &=
    \pi - \mathrm{arccos}(
    \hat{g}_{\epsilon^{-}} \cdot \hat{g}_{\epsilon^{+}}
    ), \label{eq:theta_epsilon}
\end{align}
where
\begin{align}
    \hat{g}_{\epsilon^{\pm}} &=
    \frac{
        g(X_{\epsilon \pm \Delta \epsilon}) - g(X_{\epsilon})
    }{
        || g(X_{\epsilon \pm \Delta \epsilon}) - g(X_{\epsilon}) || _{2}
    }.
\end{align}
$\Delta \epsilon$ ($>0$) controls the amount of forward and backward shifts along the direction of $\delta$ in the input space.
$\hat{g}_{\epsilon^{\pm}}$ signifies the corresponding shifts in the feature space (with normalization).
$\theta_{\epsilon}$ indicates the orientation of these shifts in the feature space, determining whether the linear displacements in the input space are also maintained in the feature space.

\begin{figure}[t!]
    \centering
    \small

    \begin{tabular}{ccc}
        & \hspace{-0.0mm} Phase
        & \hspace{-1.0mm} Pixel \\

        \hspace{-2.0mm} \rotatebox[origin=l]{90}{\hspace{9.5mm}ViT-B} &
        \hspace{-4.0mm} \includegraphics[width=.3\linewidth]{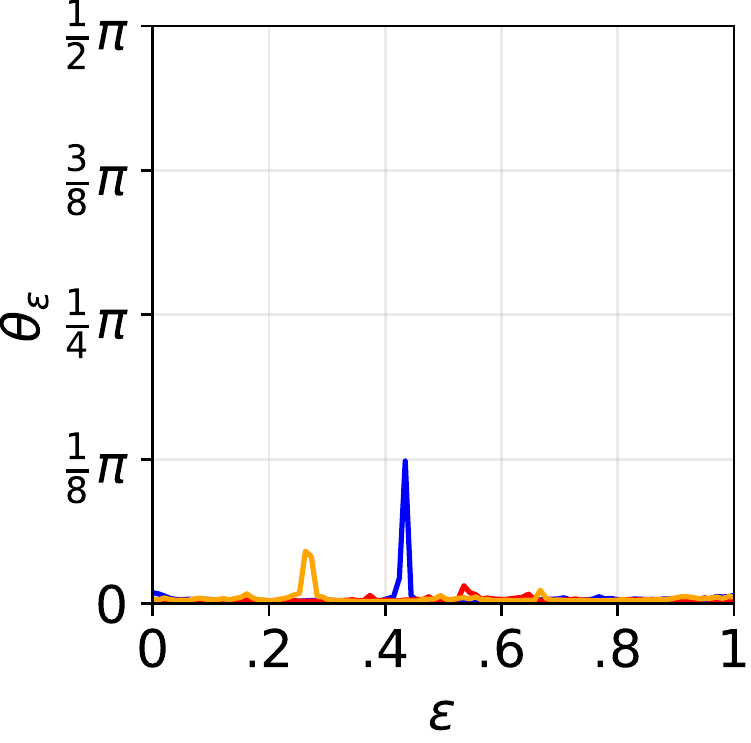} &
        \hspace{-5.0mm} \includegraphics[width=.3\linewidth]{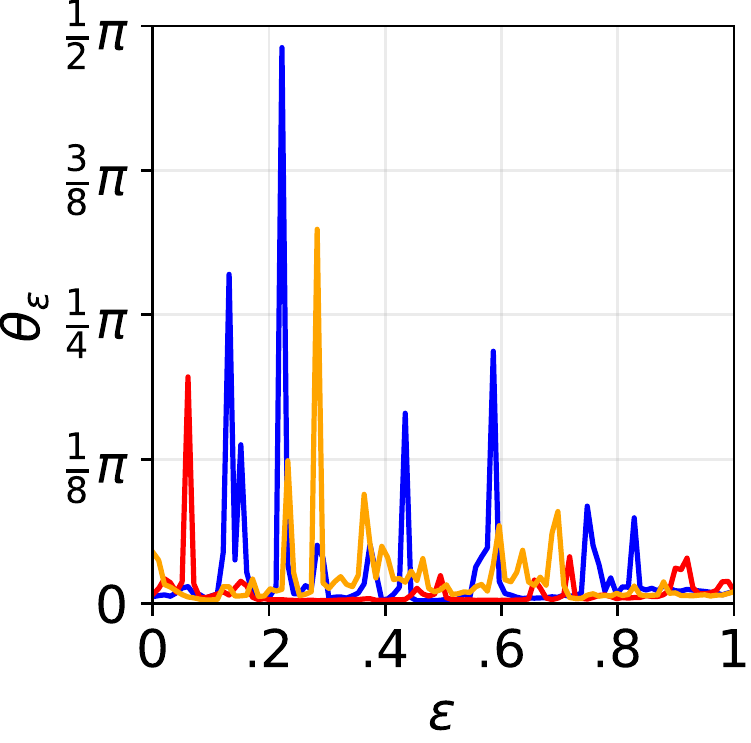} \\
        & \hspace{0.8mm} (a)
        & \hspace{-1.0mm} (b) \\

        \hspace{-2.0mm} \rotatebox[origin=l]{90}{\hspace{8.5mm}Swin-B} &
        \hspace{-4.0mm} \includegraphics[width=.3\linewidth]{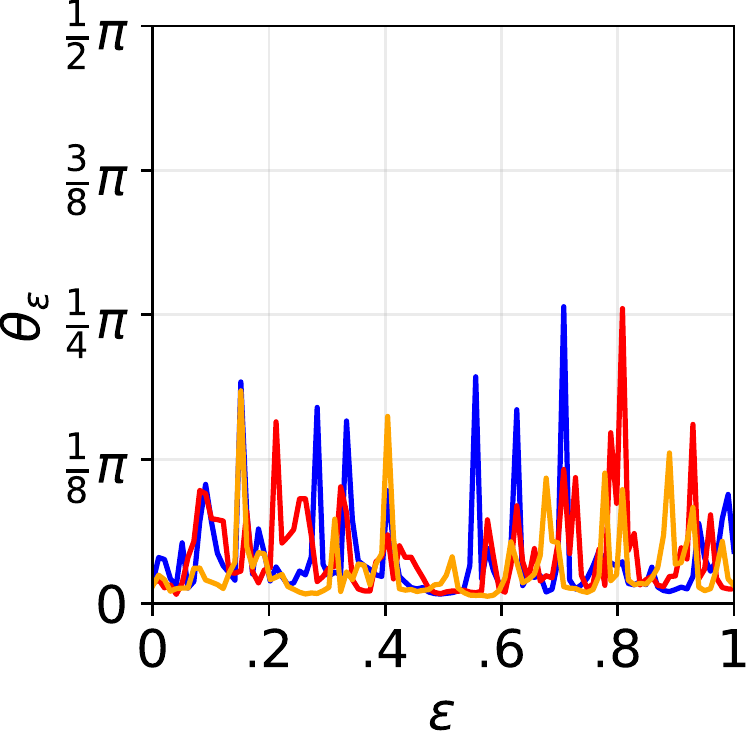} &
        \hspace{-5.0mm} \includegraphics[width=.3\linewidth]{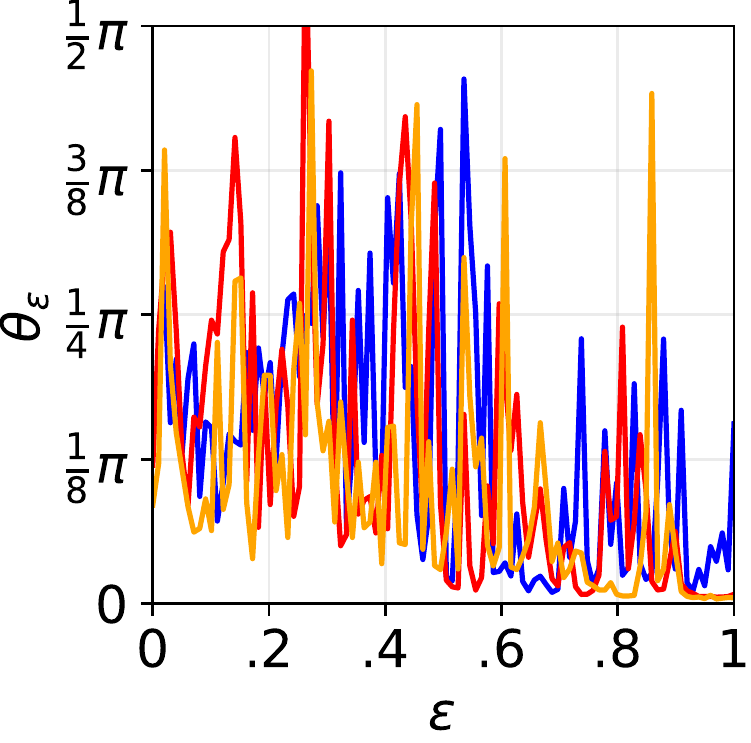} \\
        & \hspace{0.8mm} (c)
        & \hspace{-1.0mm} (d)
    \end{tabular}

    \caption{
        Direction changes of output features ($\theta_{\epsilon}$) for ViT-B and Swin-B. Different colors indicate the results of different images.
    }\label{fig:linearity}
\end{figure}

Figure~\ref{fig:linearity} shows $\theta_{\epsilon}$ of ViT-B and Swin-B for $\delta = X' - X$, where $X$ is an image, $X'$ is the attacked result of $X$ with $\lambda = 5 \times 10^3$, and $\Delta \epsilon = 0.001$. For the pixel attack, high peaks of $\theta_{\epsilon}$ appear as $\epsilon$ changes from $0$ to $1$ (Figures~\ref{fig:linearity}(b) and \ref{fig:linearity}(d)), i.e., the models are relatively nonlinear (as also shown in~\cite{a:15}) and thus are robust to the pixel attack that is linear. However, for the phase attack, which perturbs the input in a nonlinear manner, $\theta_{\epsilon}$ remains as small values (Figures~\ref{fig:linearity}(a) and \ref{fig:linearity}(c)), meaning that the features are more effectively moved in the feature space in the adversarial direction and thus become more vulnerable.

\begin{figure}[t]
\begin{subfigure}[t]{0.22\textwidth}
\includegraphics[width=\textwidth]{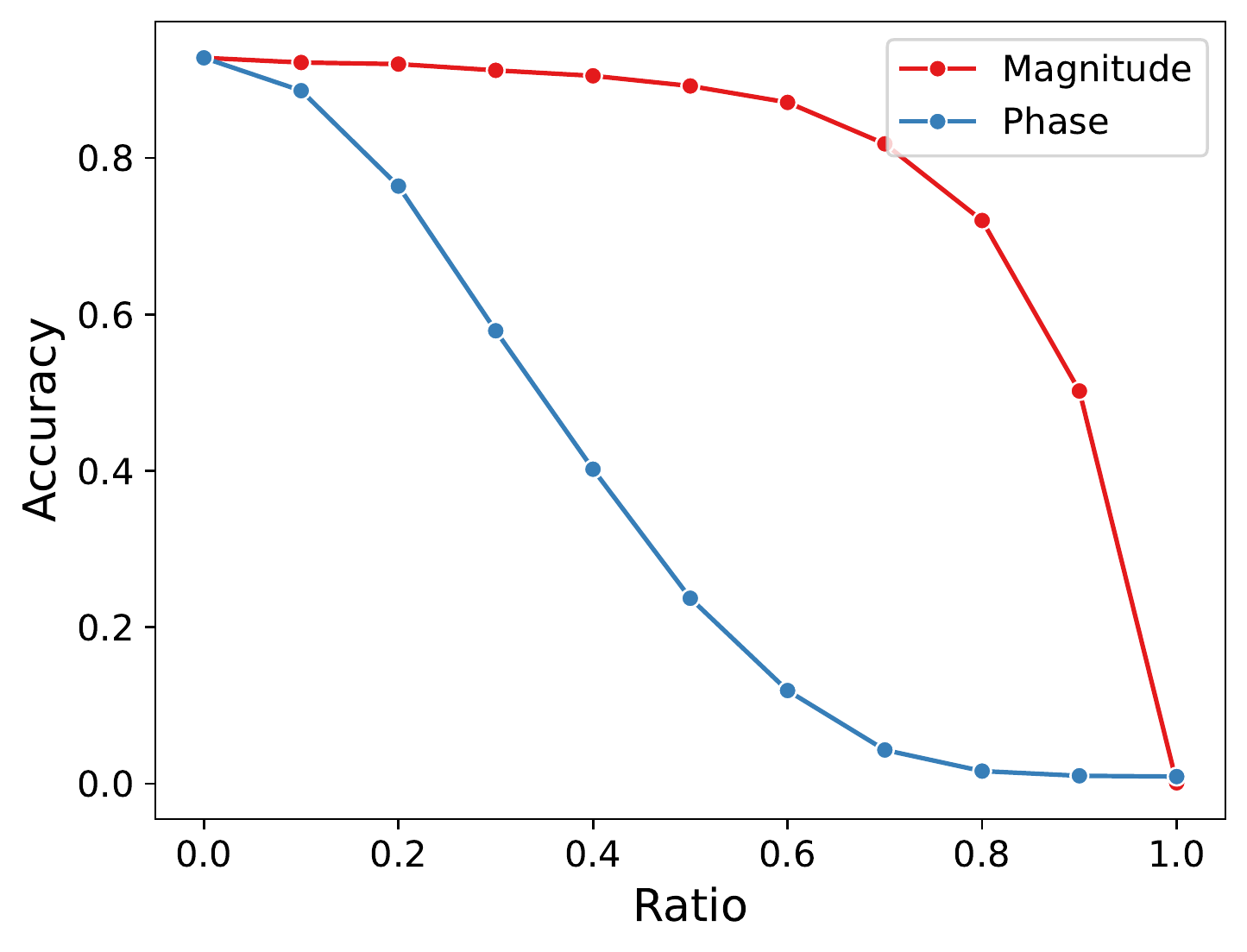}
\caption{ResNet50}
\label{figure7:a}
\end{subfigure}
\hfill
\begin{subfigure}[t]{0.22\textwidth}
\includegraphics[width=\textwidth]{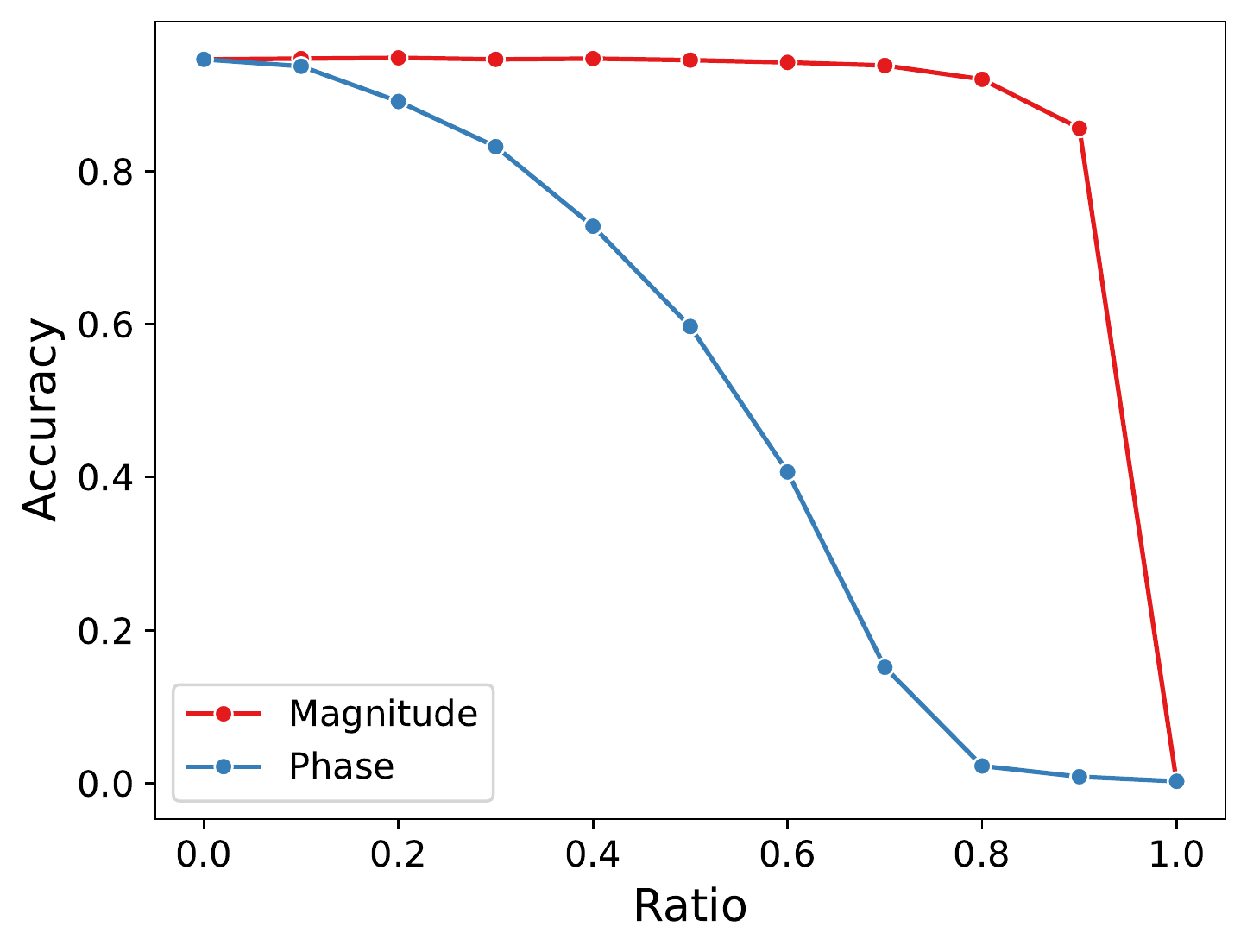}
\caption{ViT-B}
\label{figure7:b}
\end{subfigure}

\caption{Classification accuracy with respect to the ratio of reduction in the magnitude or phase spectrum.}
\label{figure7}
\end{figure}

\subsection{Dependence on Magnitude and Phase}
\label{phase_mag}
\noindent In most results above, the magnitude attack appears to be weaker than the phase attack. We investigate this phenomenon further.
\subsubsection{Sensitivity to Reduced Magnitude and Phase}
We conduct an experiment to evaluate the impact of gradually reducing the magnitude or phase spectrum on the classification accuracy without any attack applied (i.e., $M'=M\times(1-r)$ or $\phi'=\phi\times(1-r)$, where $r\in\{0, 0.1, ..., 1\}$). The results, as shown in Figure \ref{figure7}, demonstrate that both ResNet50 and ViT-B are more sensitive to phase reduction than magnitude reduction. Notably, even with 90\% of the magnitude reduced, ViT-B still achieves an accuracy of 85.6\%. These results suggest that the corruption of the phase spectrum has a more significant impact on model performance than that of the magnitude, which explains the higher vulnerability of the models to the phase attack.

\begin{table}
\begin{center}
\begin{tabular}{c|ccc}
\hline\hline
Model & Phase ($\%$) & Magnitude ($\%$) & Else ($\%$) \\
\hline
ResNet50 & 2.04 & 0.11 & 97.85 \\
ViT-B & 33.92 & 0.26 & 65.82 \\
\hline
\hline
\end{tabular}
\caption{Proportions of magnitude-phase-recombined images that are classified to the classes of the magnitude or phase images, or some other classes.}
\label{table1}
\vspace{-1em}
\end{center}
\end{table}

\subsubsection{Magnitude-Phase Recombination}
Inspired by \cite{i:4}, we conduct an experiment where the magnitude component of one image and the phase component of another image are recombined in the frequency domain and the classification result of this recombined image from a model is tested. For all possible image pairs, Table \ref{table1} shows the proportion of the images that are classified as the class of the magnitude image or phase image, or none of the two classes. For ResNet50, most of the cases (97.85\%) do not follow either the classes of the magnitude or the phase, while for the rest, more images follow the phase classes (2.04\%) than the magnitude classes (0.11\%). For ViT-B, however, a considerable amount of recombined images (33.92\%) are classified as the class of the phase images, while only 0.26\% of the images follow the classes of the magnitude images. These results highlight the relative importance of the phase information, providing an explanation on the particular vulnerability of Transformers to the phase attack and the higher strength of the phase attack than the magnitude attack for Transformers.

\section{Conclusion}
\label{conclusion}
\noindent We comparatively investigated the adversarial robustness of CNNs and Transformers using the unified attack framework with consideration of the relationship between the attack strength and ASR. Our study provides a unique contribution to the field by revealing that the vulnerability of models to adversarial attacks is highly dependent on the type of attack and the frequency regions where the perturbations are injected. Specifically, we found that Transformers are more vulnerable to the phase and magnitude attacks that mainly inject perturbations in the low frequency regions, while CNNs are more vulnerable to the pixel attack that injects perturbations mainly in the high frequency regions. We provided an explanation for this difference in the viewpoint of linearity of the models and attacks.

Our results provide insights into the underlying mechanisms of adversarial attacks and highlight the importance of considering the frequency domain when evaluating and improving the robustness of deep learning models. Furthermore, we observed that the phase information plays a more important role in classification for both CNNs and Transformers than the magnitude information, and reliance on the phase information is more prominent in Transformers.

As a future work, it would be interesting to explore the components that make ViTs vulnerable to the phase attack, based on which robust ViT structures could be developed.

\section*{Acknowledgements}
This work was supported by Artificial Intelligence Graduate School Program, Yonsei University under Grant 2020-0-01361.

\bibliographystyle{IEEEtran}
\bibliography{IEEEabrv,./references.bib}

\begin{thebibliography}{10}
\providecommand{\url}[1]{#1}
\csname url@samestyle\endcsname
\providecommand{\newblock}{\relax}
\providecommand{\bibinfo}[2]{#2}
\providecommand{\BIBentrySTDinterwordspacing}{\spaceskip=0pt\relax}
\providecommand{\BIBentryALTinterwordstretchfactor}{4}
\providecommand{\BIBentryALTinterwordspacing}{\spaceskip=\fontdimen2\font plus
\BIBentryALTinterwordstretchfactor\fontdimen3\font minus
  \fontdimen4\font\relax}
\providecommand{\BIBforeignlanguage}[2]{{%
\expandafter\ifx\csname l@#1\endcsname\relax
\typeout{** WARNING: IEEEtran.bst: No hyphenation pattern has been}%
\typeout{** loaded for the language `#1'. Using the pattern for}%
\typeout{** the default language instead.}%
\else
\language=\csname l@#1\endcsname
\fi
#2}}
\providecommand{\BIBdecl}{\relax}
\BIBdecl

\bibitem{i:5}
A.~Dosovitskiy, L.~Beyer, A.~Kolesnikov, D.~Weissenborn, X.~Zhai,
  T.~Unterthiner, M.~Dehghani, M.~Minderer, G.~Heigold, S.~Gelly \emph{et~al.},
  ``An image is worth 16x16 words: transformers for image recognition at
  scale,'' in \emph{Proceedings of the 9th International Conference on Learning
  Representations}, 2021.

\bibitem{a:7}
N.~Akhtar and A.~Mian, ``Threat of adversarial attacks on deep learning in
  computer vision: A survey,'' \emph{IEEE Access}, vol.~6, pp.
  14\,410--14\,430, 2018.

\bibitem{a:1}
M.~M. Naseer, K.~Ranasinghe, S.~H. Khan, M.~Hayat, F.~Shahbaz~Khan, and M.-H.
  Yang, ``Intriguing properties of vision transformers,'' in \emph{Advances in
  Neural Information Processing Systems}, vol.~34, 2021, pp. 23\,296--23\,308.

\bibitem{a:6}
P.~Benz, S.~Ham, C.~Zhang, A.~Karjauv, and I.~S. Kweon, ``Adversarial
  robustness comparison of vision transformer and {MLP}-mixer to {CNN}s,'' in
  \emph{Proceedings of the 32nd British Machine Vision Conference}, 2021.

\bibitem{m:2}
R.~Shao, Z.~Shi, J.~Yi, P.-Y. Chen, and C.-J. Hsieh, ``On the adversarial
  robustness of vision transformers,'' \emph{arXiv preprint arXiv:2103.15670},
  2021.

\bibitem{m:3}
A.~Aldahdooh, W.~Hamidouche, and O.~Deforges, ``Reveal of vision transformers
  robustness against adversarial attacks,'' \emph{arXiv preprint
  arXiv:2106.03734}, 2021.

\bibitem{a:2}
Y.~Bai, J.~Mei, A.~L. Yuille, and C.~Xie, ``Are transformers more robust than
  {CNN}s?'' in \emph{Advances in Neural Information Processing Systems},
  vol.~34, 2021, pp. 26\,831--26\,843.

\bibitem{i:1}
K.~Mahmood, R.~Mahmood, and M.~Van~Dijk, ``On the robustness of vision
  transformers to adversarial examples,'' in \emph{Proceedings of the IEEE/CVF
  International Conference on Computer Vision}, 2021, pp. 7838--7847.

\bibitem{i:2}
S.~Bhojanapalli, A.~Chakrabarti, D.~Glasner, D.~Li, T.~Unterthiner, and
  A.~Veit, ``Understanding robustness of transformers for image
  classification,'' in \emph{Proceedings of the IEEE/CVF International
  Conference on Computer Vision}, 2021, pp. 10\,231--10\,241.

\bibitem{i:13}
N.~Park and S.~Kim, ``How do vision transformers work?'' in \emph{Proceedings
  of the 10th International Conference on Learning Representations}, 2022.

\bibitem{i:15}
H.~Wang, X.~Wu, Z.~Huang, and E.~P. Xing, ``High-frequency component helps
  explain the generalization of convolutional neural networks,'' in
  \emph{Proceedings of the IEEE/CVF Conference on Computer Vision and Pattern
  Recognition}, 2020, pp. 8684--8694.

\bibitem{m:6}
J.~Jo and Y.~Bengio, ``Measuring the tendency of {CNN}s to learn surface
  statistical regularities,'' \emph{arXiv preprint arXiv:1711.11561}, 2017.

\bibitem{i:8}
Z.~Liu, Y.~Lin, Y.~Cao, H.~Hu, Y.~Wei, Z.~Zhang, S.~Lin, and B.~Guo, ``Swin
  transformer: Hierarchical vision transformer using shifted windows,'' in
  \emph{Proceedings of the IEEE/CVF International Conference on Computer
  Vision}, 2021, pp. 10\,012--10\,022.

\bibitem{i:6}
H.~Touvron, M.~Cord, M.~Douze, F.~Massa, A.~Sablayrolles, and H.~J{\'e}gou,
  ``Training data-efficient image transformers \& distillation through
  attention,'' in \emph{Proceedings of the 38th International Conference on
  Machine Learning}, 2021, pp. 10\,347--10\,357.

\bibitem{i:23}
L.~Yuan, Y.~Chen, T.~Wang, W.~Yu, Y.~Shi, Z.-H. Jiang, F.~E. Tay, J.~Feng, and
  S.~Yan, ``Tokens-to-token {ViT}: Training vision transformers from scratch on
  {ImageNet},'' in \emph{Proceedings of the IEEE/CVF International Conference
  on Computer Vision}, 2021, pp. 558--567.

\bibitem{i:24}
W.~Wang, E.~Xie, X.~Li, D.-P. Fan, K.~Song, D.~Liang, T.~Lu, P.~Luo, and
  L.~Shao, ``Pyramid vision transformer: A versatile backbone for dense
  prediction without convolutions,'' in \emph{Proceedings of the IEEE/CVF
  International Conference on Computer Vision}, 2021, pp. 568--578.

\bibitem{a:8}
K.~Han, A.~Xiao, E.~Wu, J.~Guo, C.~Xu, and Y.~Wang, ``Transformer in
  transformer,'' in \emph{Advances in Neural Information Processing Systems},
  vol.~34, 2021, pp. 15\,908--15\,919.

\bibitem{a:9}
A.~Ali, H.~Touvron, M.~Caron, P.~Bojanowski, M.~Douze, A.~Joulin, I.~Laptev,
  N.~Neverova, G.~Synnaeve, J.~Verbeek \emph{et~al.}, ``{XCIT}:
  Cross-covariance image transformers,'' in \emph{Advances in Neural
  Information Processing Systems}, vol.~34, 2021, pp. 20\,014--20\,027.

\bibitem{i:9}
I.~J. Goodfellow, J.~Shlens, and C.~Szegedy, ``Explaining and harnessing
  adversarial examples,'' in \emph{Proceedings of the 3rd International
  Conference on Learning Representations}, 2015.

\bibitem{i:12}
A.~Madry, A.~Makelov, L.~Schmidt, D.~Tsipras, and A.~Vladu, ``Towards deep
  learning models resistant to adversarial attacks,'' in \emph{Proceedings of
  the 6th International Conference on Learning Representations}, 2018.

\bibitem{i:11}
N.~Carlini and D.~Wagner, ``Towards evaluating the robustness of neural
  networks,'' in \emph{Proceedings of the IEEE Symposium on Security and
  Privacy}, 2017, pp. 39--57.

\bibitem{i:3}
X.-C. Li, X.-Y. Zhang, F.~Yin, and C.-L. Liu, ``F-mixup: Attack {CNN}s from
  {Fourier} perspective,'' in \emph{Proceedings of the 25th International
  Conference on Pattern Recognition}, 2020, pp. 541--548.

\bibitem{i:21}
C.~Guo, J.~S. Frank, and K.~Q. Weinberger, ``Low frequency adversarial
  perturbation,'' in \emph{Proceedings of the Uncertainty in Artificial
  Intelligence}, 2020, pp. 1127--1137.

\bibitem{i:22}
Y.~Sharma, G.~W. Ding, and M.~A. Brubaker, ``On the effectiveness of low
  frequency perturbations,'' in \emph{Proceedings of the 28th International
  Joint Conference on Artificial Intelligence}, 2019, pp. 3389--3396.

\bibitem{i:33}
R.~Duan, Y.~Chen, D.~Niu, Y.~Yang, A.~K. Qin, and Y.~He, ``Advdrop: Adversarial
  attack to {DNN}s by dropping information,'' in \emph{Proceedings of the
  IEEE/CVF International Conference on Computer Vision}, 2021, pp. 7506--7515.

\bibitem{a:14}
A.~Agarwal, N.~Ratha, M.~Vatsa, and R.~Singh, ``Crafting adversarial
  perturbations via transformed image component swapping,'' \emph{IEEE
  Transactions on Image Processing}, vol.~31, pp. 7338--7349, 2022.

\bibitem{i:27}
Z.~Wen, ``Fourier attack--a more efficient adversarial attack method,'' in
  \emph{Proceedings of the 6th International Conference on Control Engineering
  and Artificial Intelligence}, 2022, pp. 125--130.

\bibitem{i:25}
S.~Paul and P.-Y. Chen, ``Vision transformers are robust learners,'' in
  \emph{Proceedings of the AAAI Conference on Artificial Intelligence},
  vol.~36, 2022, pp. 2071--2081.

\bibitem{a:15}
J.~Kim, J.~Park, S.~Kim, and J.-S. Lee, ``Curved representation space of vision
  transformers,'' \emph{arXiv preprint arXiv:2210.05742}, 2022.

\bibitem{i:35}
A.~Liu, S.~Tang, S.~Liang, R.~Gong, B.~Wu, X.~Liu, and D.~Tao, ``Exploring the
  relationship between architectural design and adversarially robust
  generalization,'' in \emph{Proceedings of the IEEE/CVF Conference on Computer
  Vision and Pattern Recognition}, 2023, pp. 4096--4107.

\bibitem{i:29}
Y.~Fu, S.~Zhang, S.~Wu, C.~Wan, and Y.~Lin, ``Patch-fool: Are vision
  transformers always robust against adversarial perturbations?'' in
  \emph{Proceedings of the International Conference on Learning
  Representations}, 2022.

\bibitem{i:31}
D.~Karmon, D.~Zoran, and Y.~Goldberg, ``Lavan: Localized and visible
  adversarial noise,'' in \emph{Proceedings of International Conference on
  Machine Learning}, 2018, pp. 2507--2515.

\bibitem{i:30}
J.~Gu, V.~Tresp, and Y.~Qin, ``Are vision transformers robust to patch
  perturbations?'' in \emph{Proceedings of the European Conference on Computer
  Vision}, 2022, pp. 404--421.

\bibitem{m:11}
R.~Wightman, ``{NIPS} 2017 adversarial competition (pytorch),''
  \url{https://github.com/rwightman/pytorch-nips2017-adversarial}, 2017.

\bibitem{a:4}
A.~Paszke, S.~Gross, F.~Massa, A.~Lerer, J.~Bradbury, G.~Chanan, T.~Killeen,
  Z.~Lin, N.~Gimelshein, L.~Antiga \emph{et~al.}, ``{PyTorch}: An imperative
  style, high-performance deep learning library,'' in \emph{Advances in Neural
  Information Processing Systems}, vol.~32, 2019.

\bibitem{a:11}
O.~Russakovsky, J.~Deng, H.~Su, J.~Krause, S.~Satheesh, S.~Ma, Z.~Huang,
  A.~Karpathy, A.~Khosla, M.~Bernstein, A.~C. Berg, and L.~Fei-Fei,
  ``{ImageNet} large scale visual recognition challenge,'' \emph{International
  Journal of Computer Vision}, vol. 115, no.~3, pp. 211--252, 2015.

\bibitem{m:10}
R.~Wightman, ``{PyTorch} image models,''
  \url{https://github.com/rwightman/pytorch-image-models}, 2019.

\bibitem{i:26}
J.~Deng, W.~Dong, R.~Socher, L.-J. Li, K.~Li, and L.~Fei-Fei, ``{ImageNet}: A
  large-scale hierarchical image database,'' in \emph{Proceedings of the
  IEEE/CVF Conference on Computer Vision and Pattern Recognition}, 2009, pp.
  248--255.

\bibitem{a:12}
Z.~Wang, E.~P. Simoncelli, and A.~C. Bovik, ``Multiscale structural similarity
  for image quality assessment,'' in \emph{Proceedings of the Asilomar
  Conference on Signals, Systems \& Computers}, vol.~2, 2003, pp. 1398--1402.

\bibitem{a:5}
H.~Z. Nafchi, A.~Shahkolaei, R.~Hedjam, and M.~Cheriet, ``Mean deviation
  similarity index: Efficient and reliable full-reference image quality
  evaluator,'' \emph{IEEE Access}, vol.~4, pp. 5579--5590, 2016.

\bibitem{i:18}
R.~Zhang, P.~Isola, A.~A. Efros, E.~Shechtman, and O.~Wang, ``The unreasonable
  effectiveness of deep features as a perceptual metric,'' in \emph{Proceedings
  of the IEEE/CVF conference on Computer Vision and Pattern Recognition}, 2018,
  pp. 586--595.

\bibitem{i:4}
G.~Chen, P.~Peng, L.~Ma, J.~Li, L.~Du, and Y.~Tian, ``Amplitude-phase
  recombination: Rethinking robustness of convolutional neural networks in
  frequency domain,'' in \emph{Proceedings of the IEEE/CVF International
  Conference on Computer Vision}, 2021, pp. 458--467.

\bibitem{m:12}
\BIBentryALTinterwordspacing
S.~Kastryulin, D.~Zakirov, and D.~Prokopenko, ``{PyTorch Image Quality}:
  Metrics and measure for image quality assessment,'' 2019, open-source
  software available at https://github.com/photosynthesis-team/piq. [Online].
  Available: \url{https://github.com/photosynthesis-team/piq}
\BIBentrySTDinterwordspacing

\bibitem{i:32}
S.~Abnar and W.~Zuidema, ``Quantifying attention flow in transformers,'' in
  \emph{Proceedings of the 58th Annual Meeting of the Association for
  Computational Linguistics}, 2020, pp. 4190--4197.

\end{thebibliography}

\appendices
\section{Results using Other Image Quality Metrics}
\begin{figure*}
\centering
    \begin{subfigure}[t]{0.33\textwidth}
    \includegraphics[width=\textwidth]{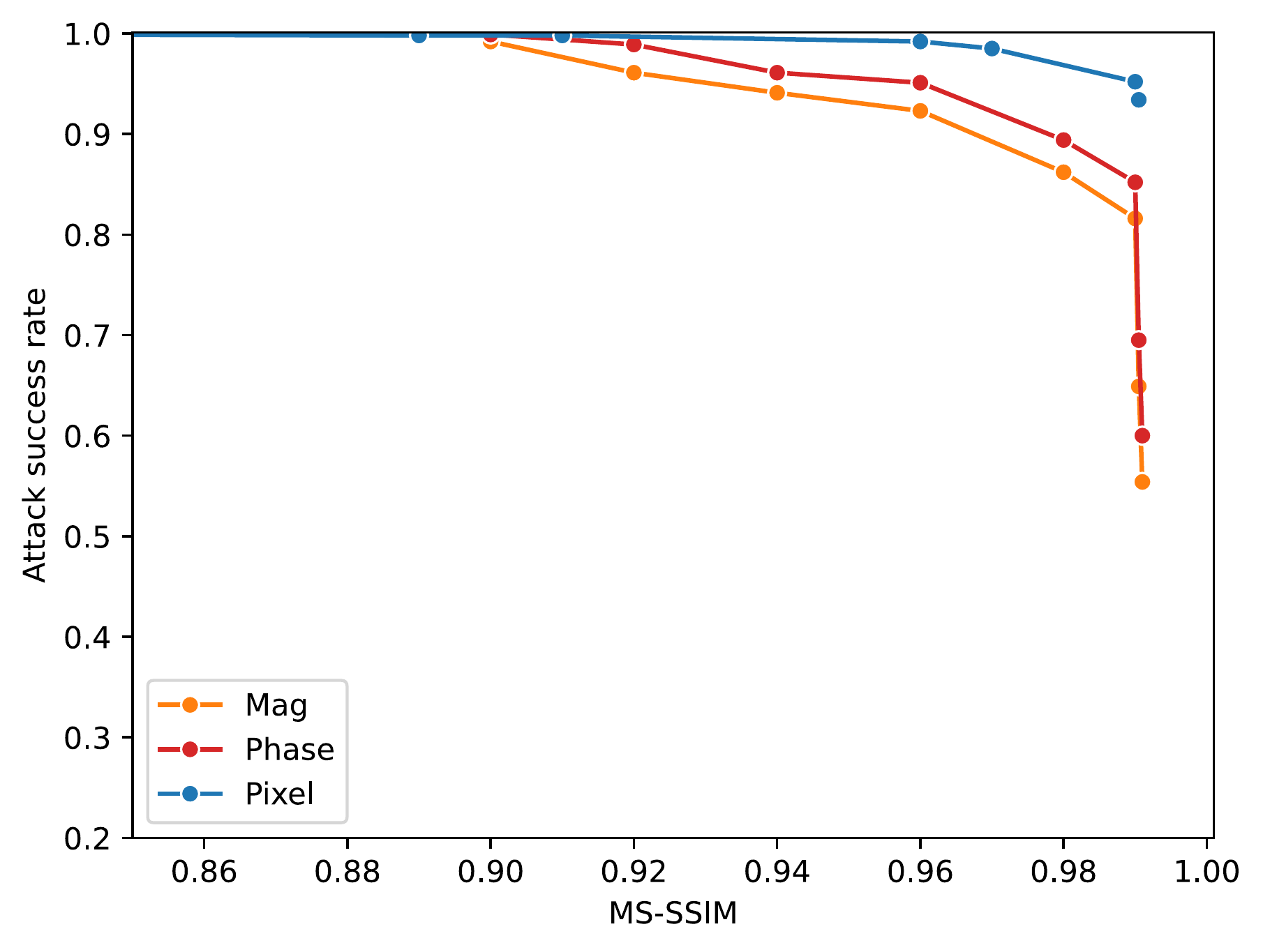}
    \caption{ResNet50}
    \label{figure8:a}
    \end{subfigure}
    \hfill
    \begin{subfigure}[t]{0.33\textwidth}
    \includegraphics[width=\textwidth]{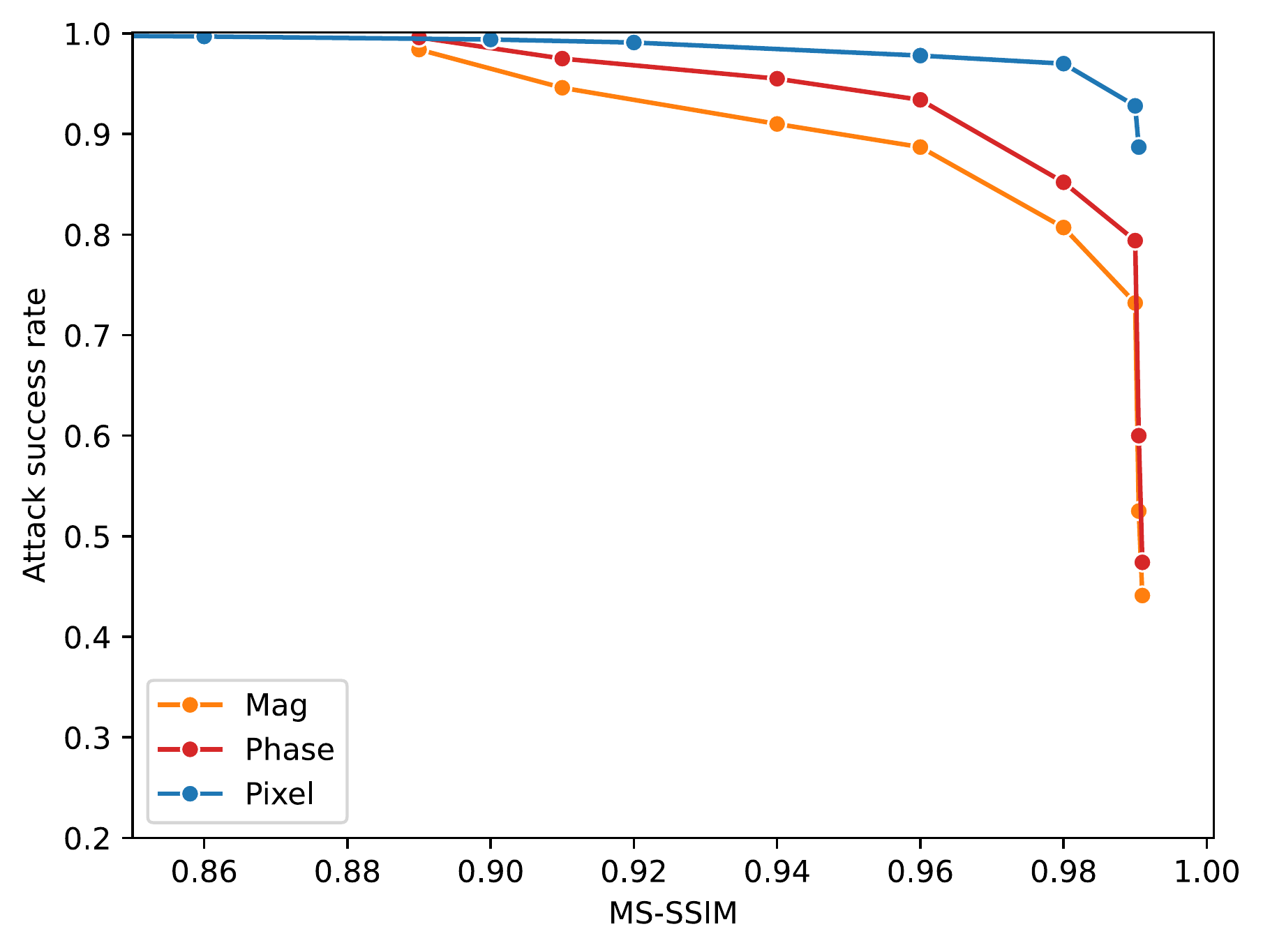}
    \caption{ResNet152}
    \label{figure8:b}
    \end{subfigure}
    \hfill
    \begin{subfigure}[t]{0.33\textwidth}
    \includegraphics[width=\textwidth]{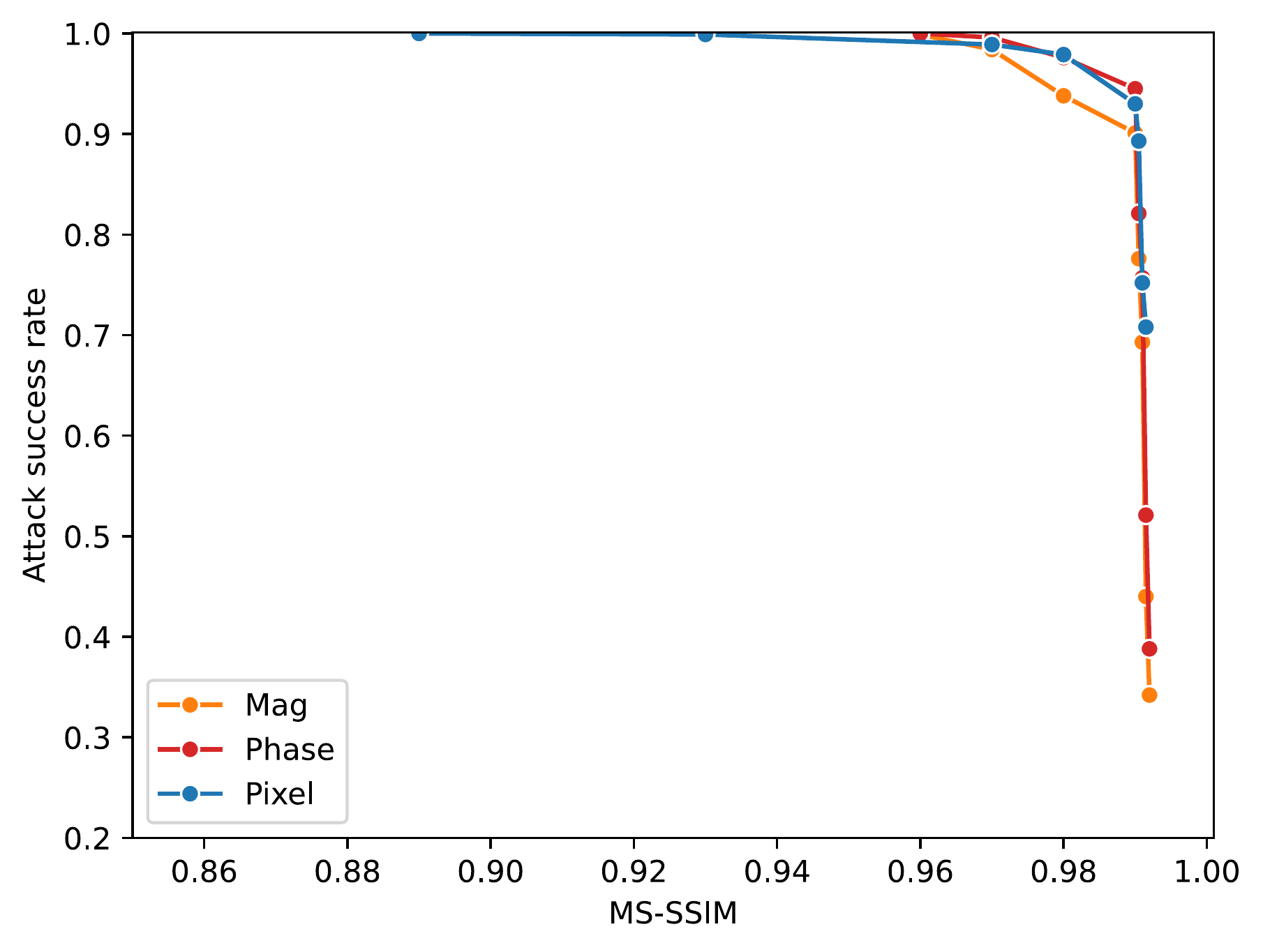}
    \caption{ViT-B}
    \label{figure8:c}
    \end{subfigure}
    \hfill
    \begin{subfigure}[t]{0.33\textwidth}
    \includegraphics[width=\textwidth]{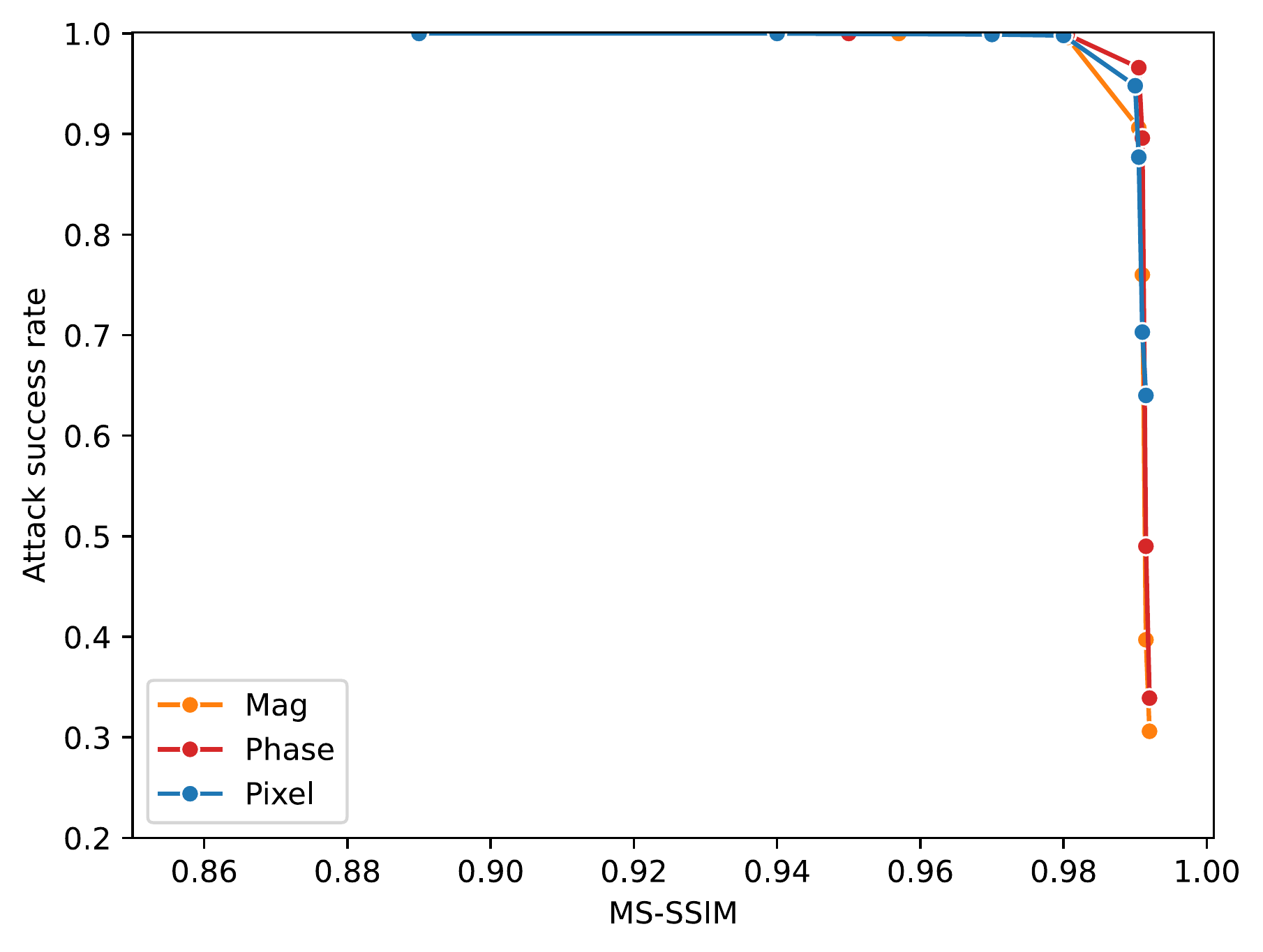}
    \caption{ViT-B-1k}
    \label{figure8:d}
    \end{subfigure}
    \hfill 
    \begin{subfigure}[t]{0.33\textwidth}
    \includegraphics[width=\textwidth]{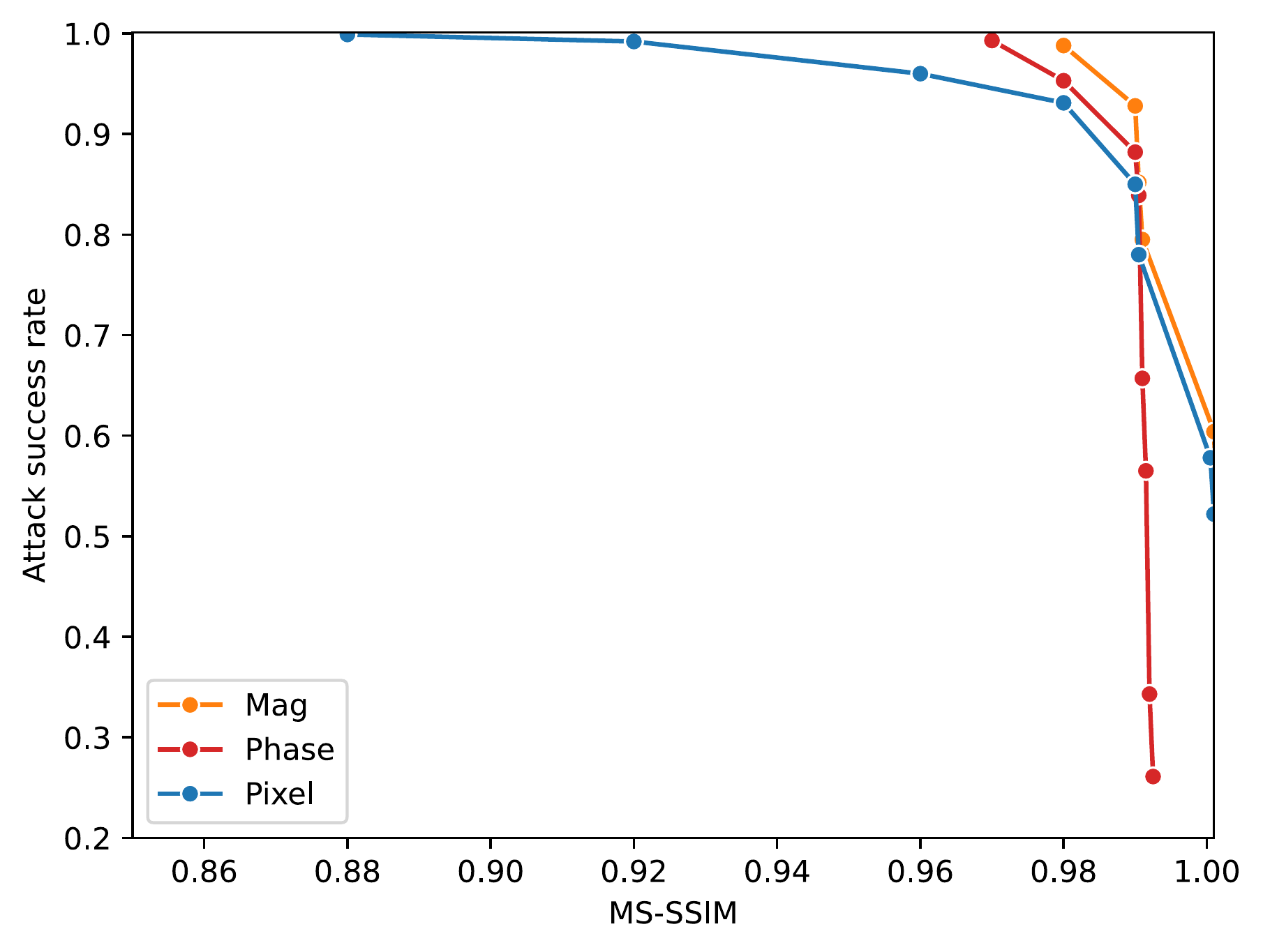}
    \caption{ViT-L}
    \label{figure8:e}
    \end{subfigure}
    \hfill
    \begin{subfigure}[t]{0.33\textwidth}
    \includegraphics[width=\textwidth]{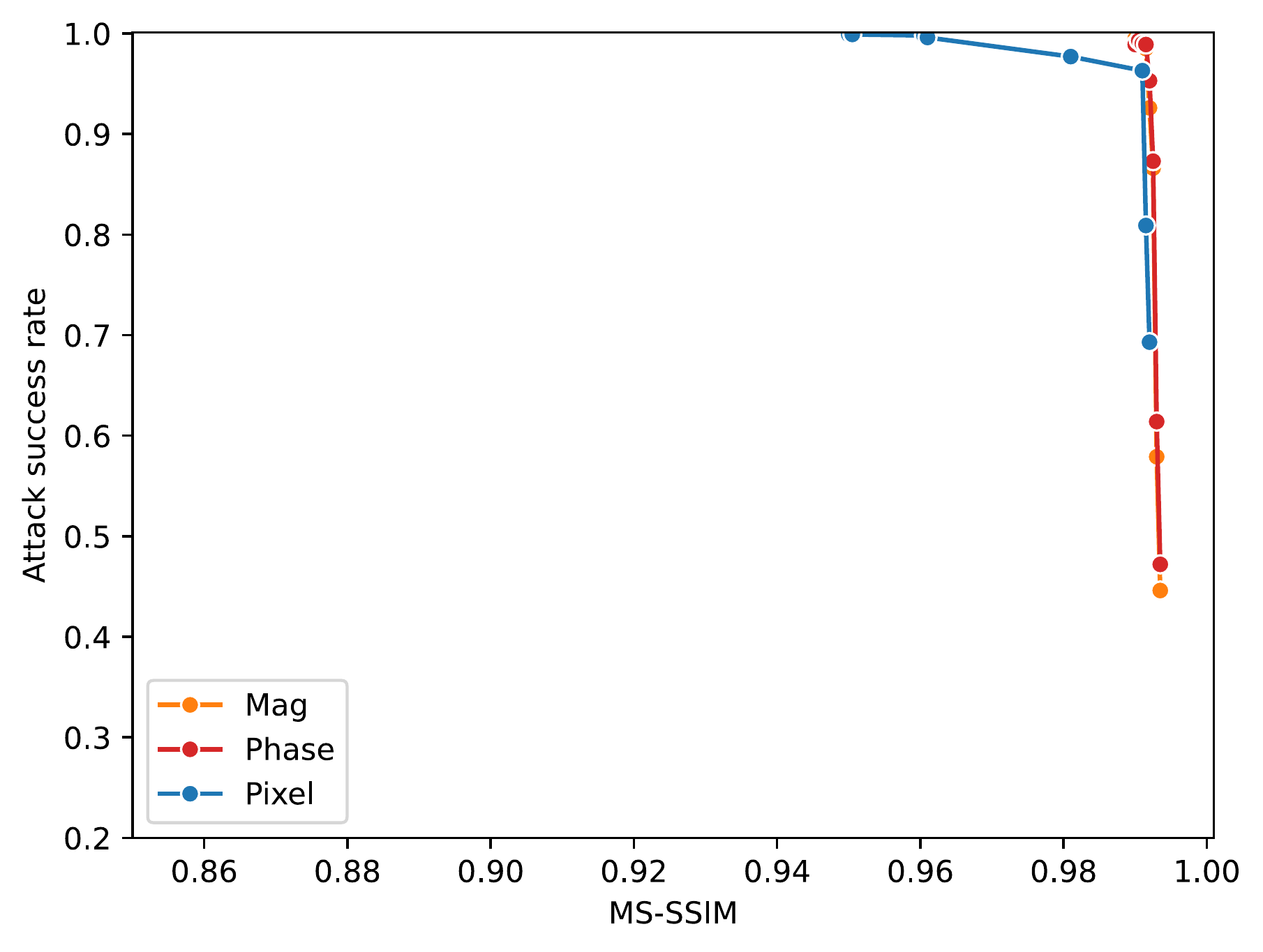}
    \caption{Swin-B}
    \label{figure8:f}
    \end{subfigure}
    \hfill  
    \begin{subfigure}[t]{0.33\textwidth}
    \includegraphics[width=\textwidth]{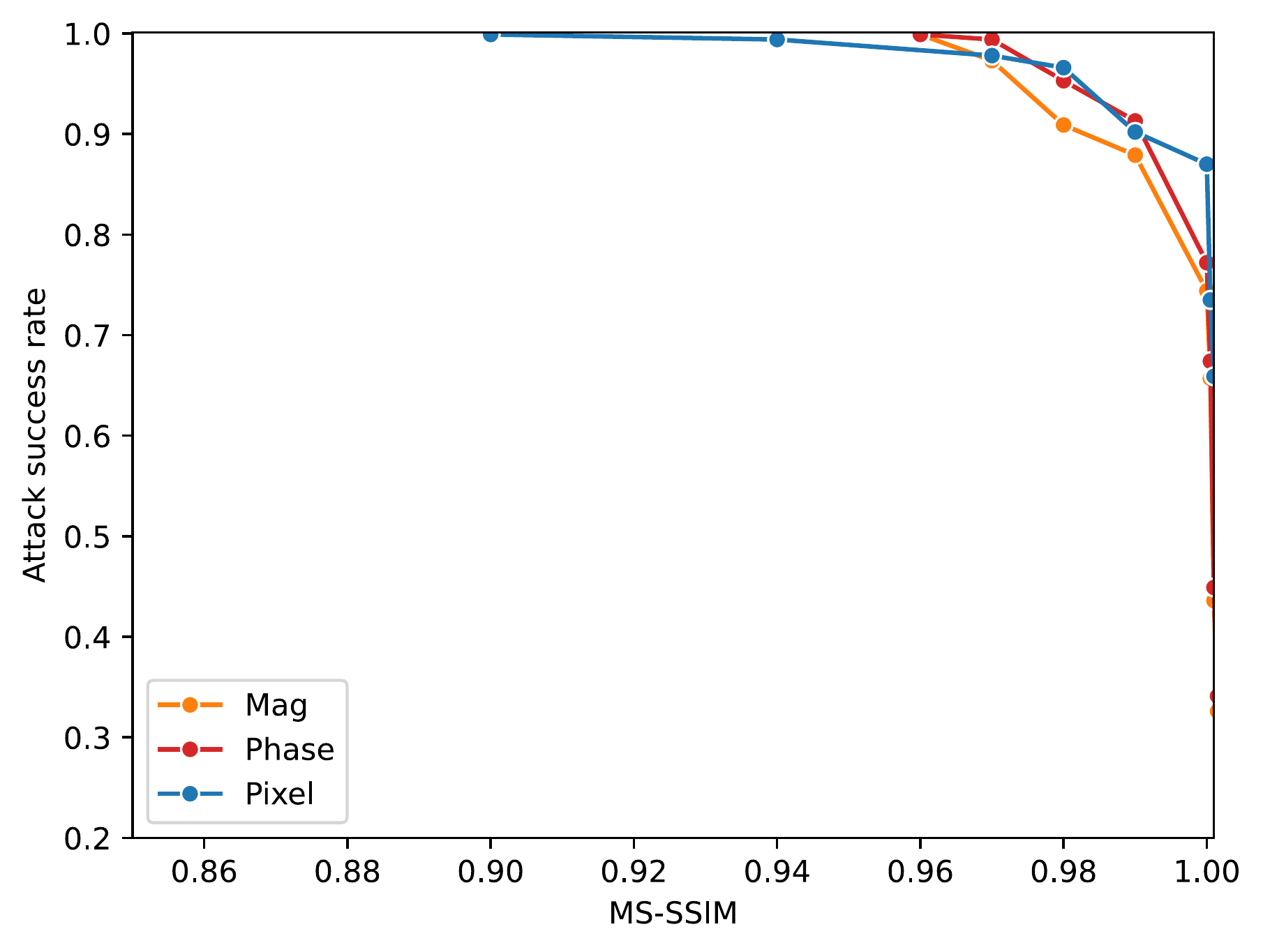}
    \caption{DeiT-S}
    \label{figure8:g}
    \end{subfigure}
    \begin{subfigure}[t]{0.33\textwidth}
    \includegraphics[width=\textwidth]{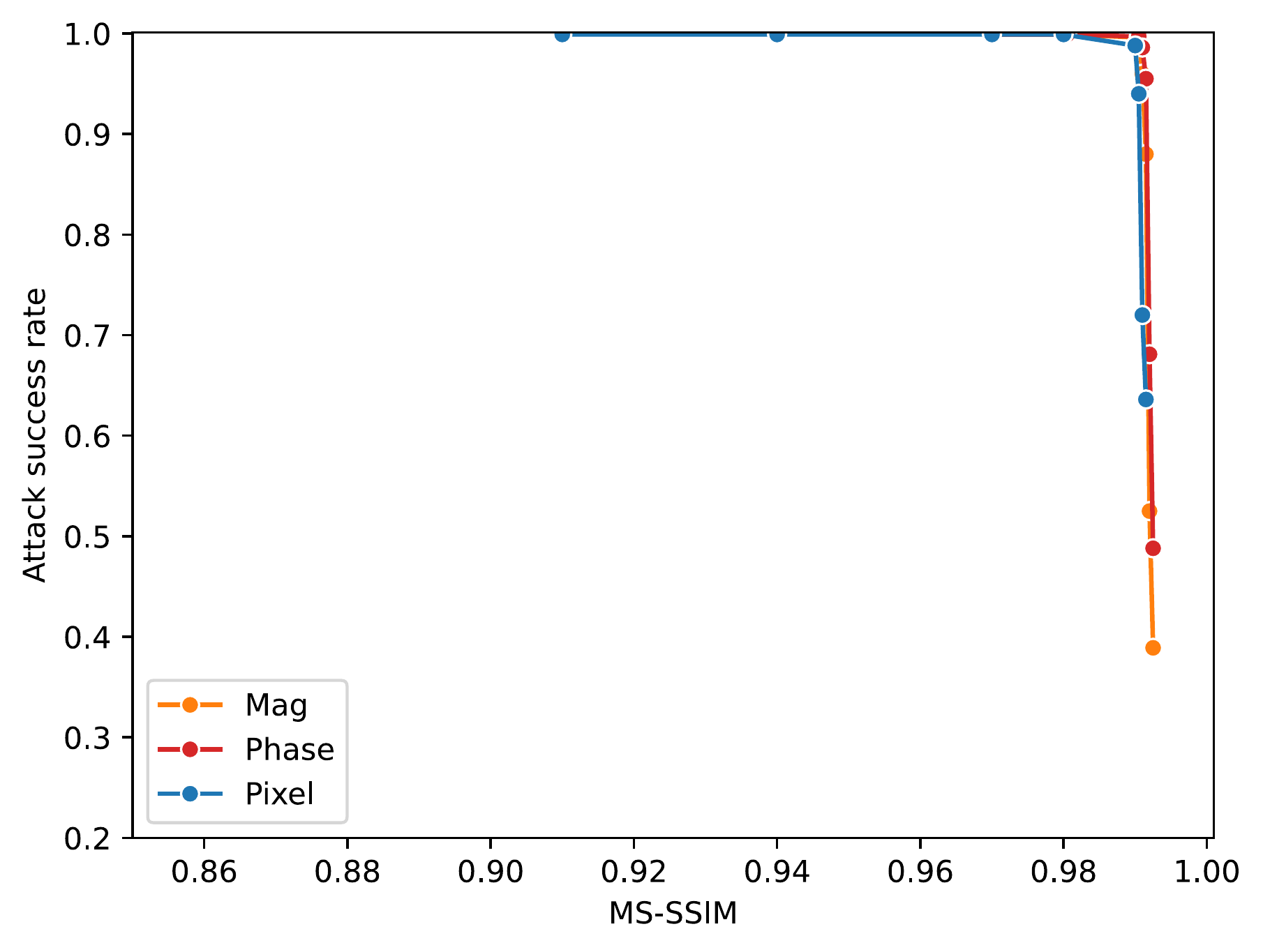}
    \caption{DeiT-S with no distillation}
    \label{figure8:h}
    \end{subfigure}
\caption{Comparison of different attacks for each model using MS-SSIM.}
\label{figure8}
\end{figure*}

\begin{figure*}
\centering
    \begin{subfigure}[t]{0.33\textwidth}
    \includegraphics[width=\textwidth]{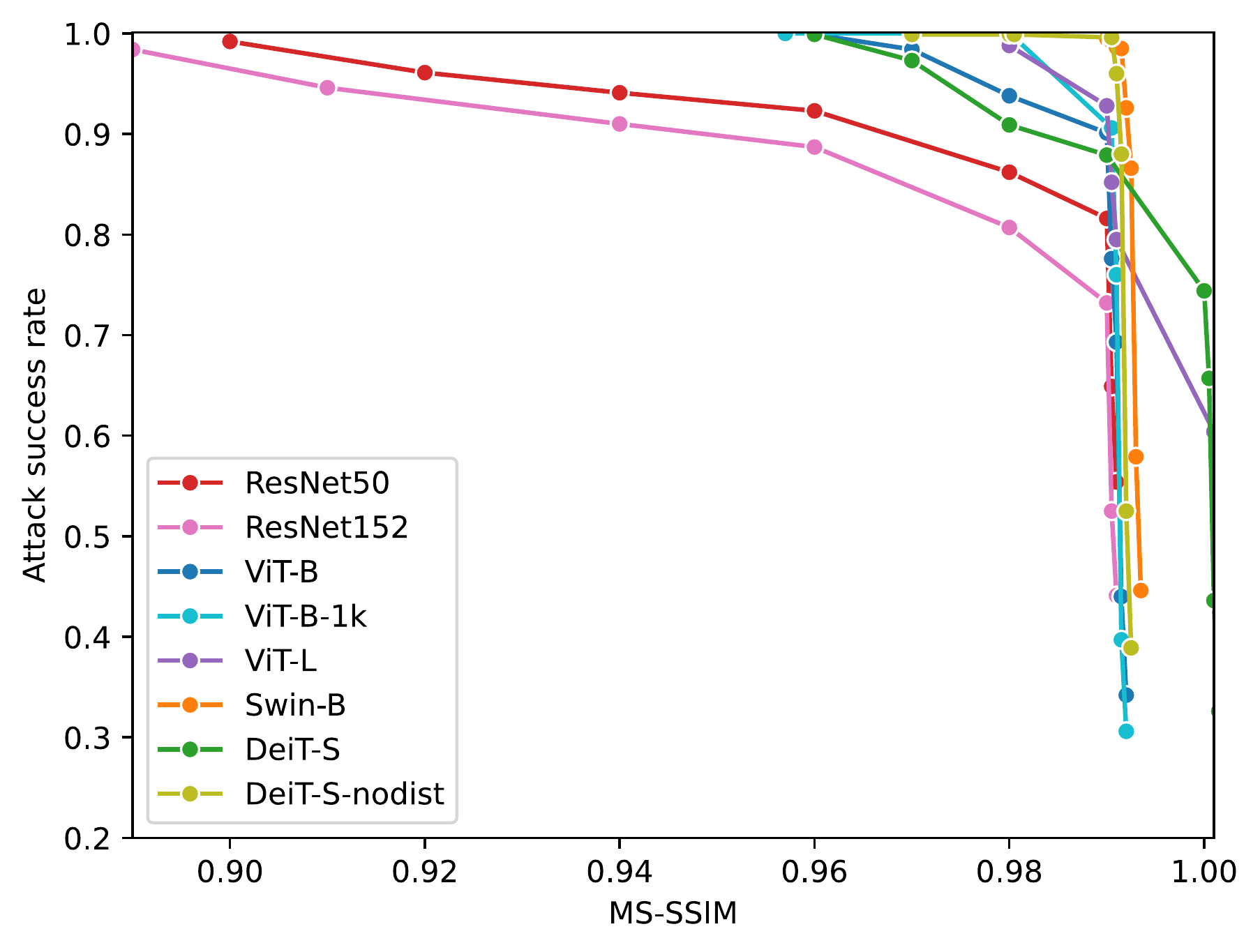}
    \caption{Magnitude attack}\label{figure9:a}
    \end{subfigure}
    \hfill
    \begin{subfigure}[t]{0.33\textwidth}
    \includegraphics[width=\textwidth]{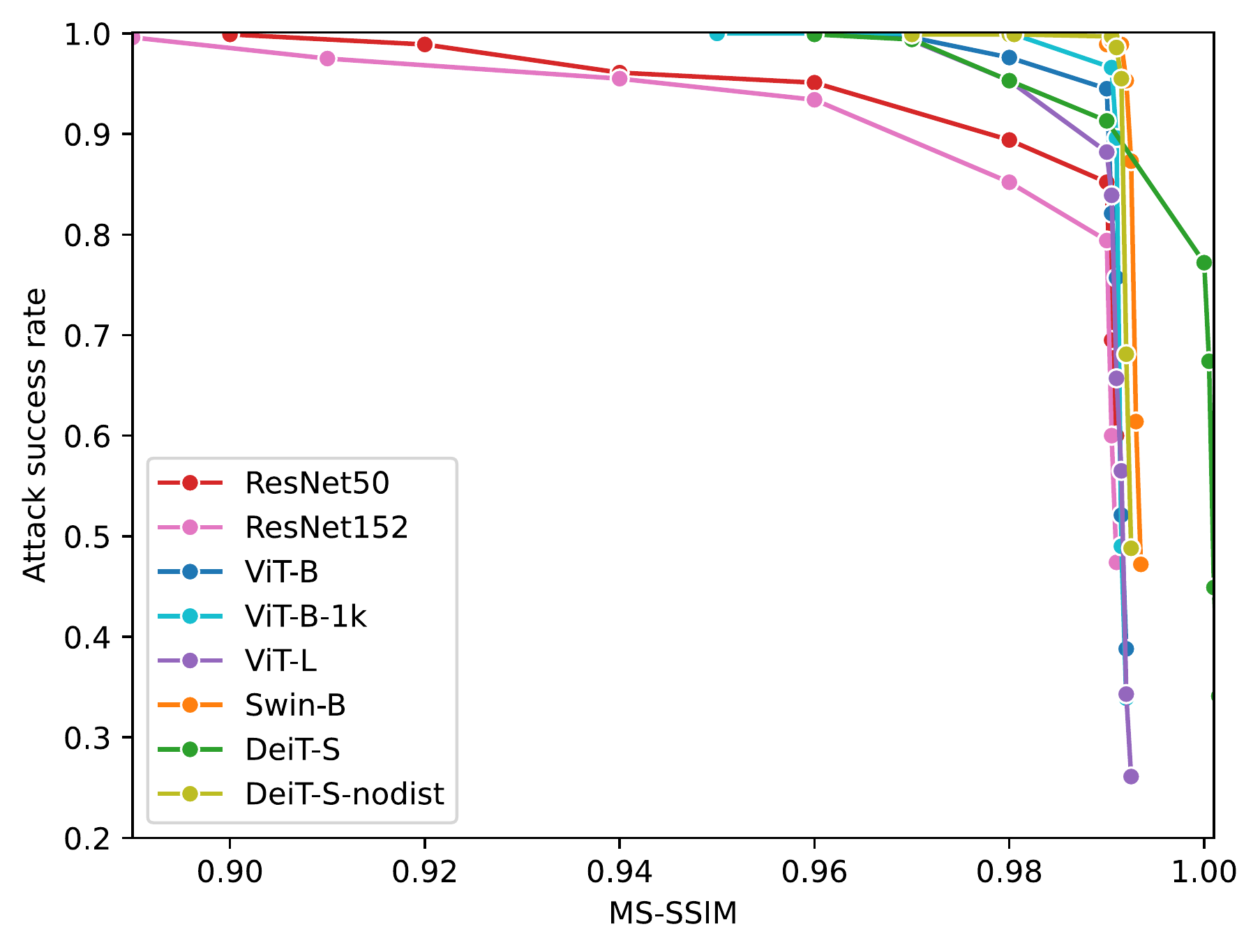}
    \caption{Phase attack}\label{figure9:b}
    \end{subfigure}
    \hfill
    \begin{subfigure}[t]{0.33\textwidth}
    \includegraphics[width=\textwidth]{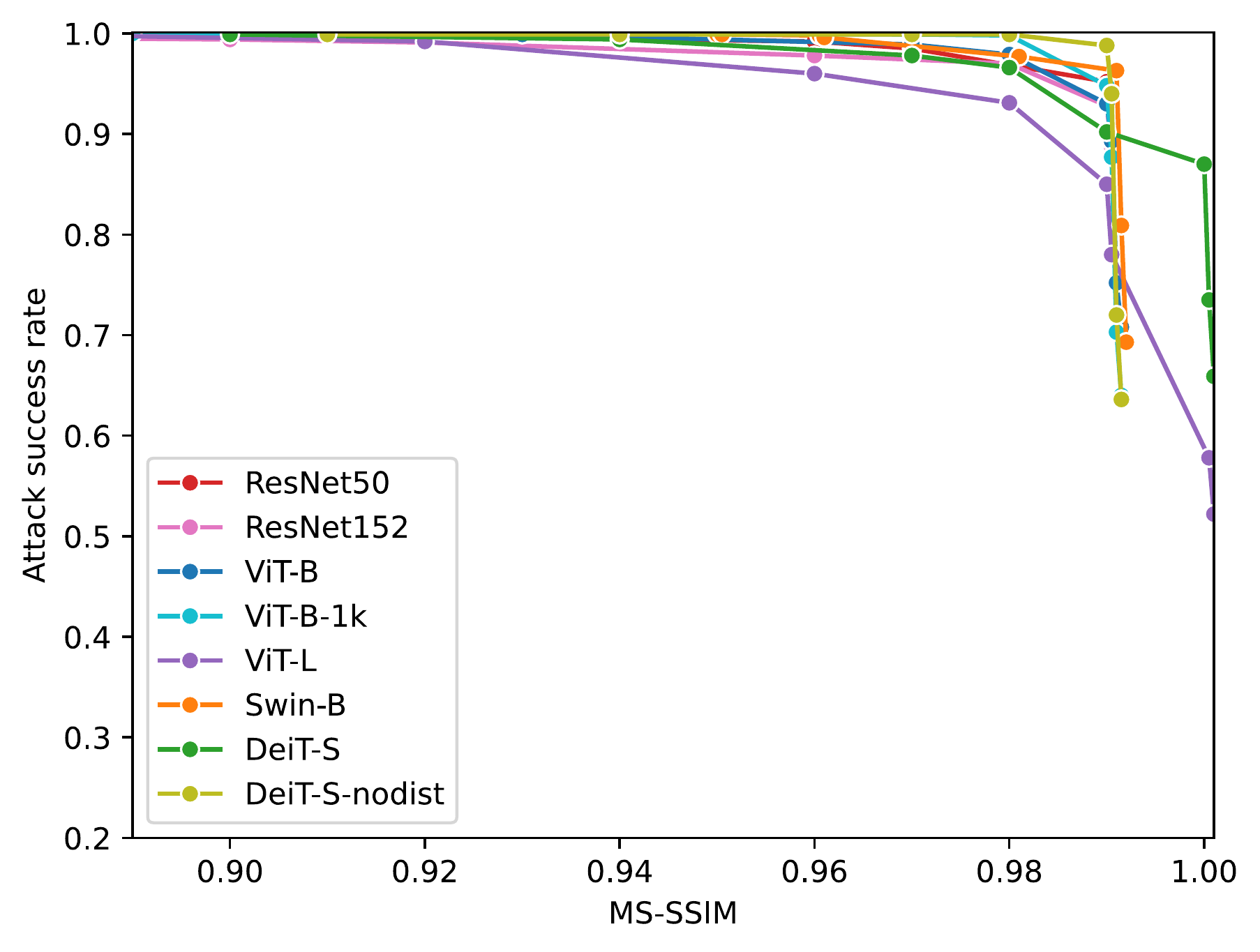}
    \caption{Pixel attack}\label{figure9:c}
    \end{subfigure}
\caption{Comparison of different models for each attack type using MS-SSIM.}
\label{figure9}
\end{figure*}

\begin{figure*}
\centering
    \begin{subfigure}[t]{0.33\textwidth}
    \includegraphics[width=\textwidth]{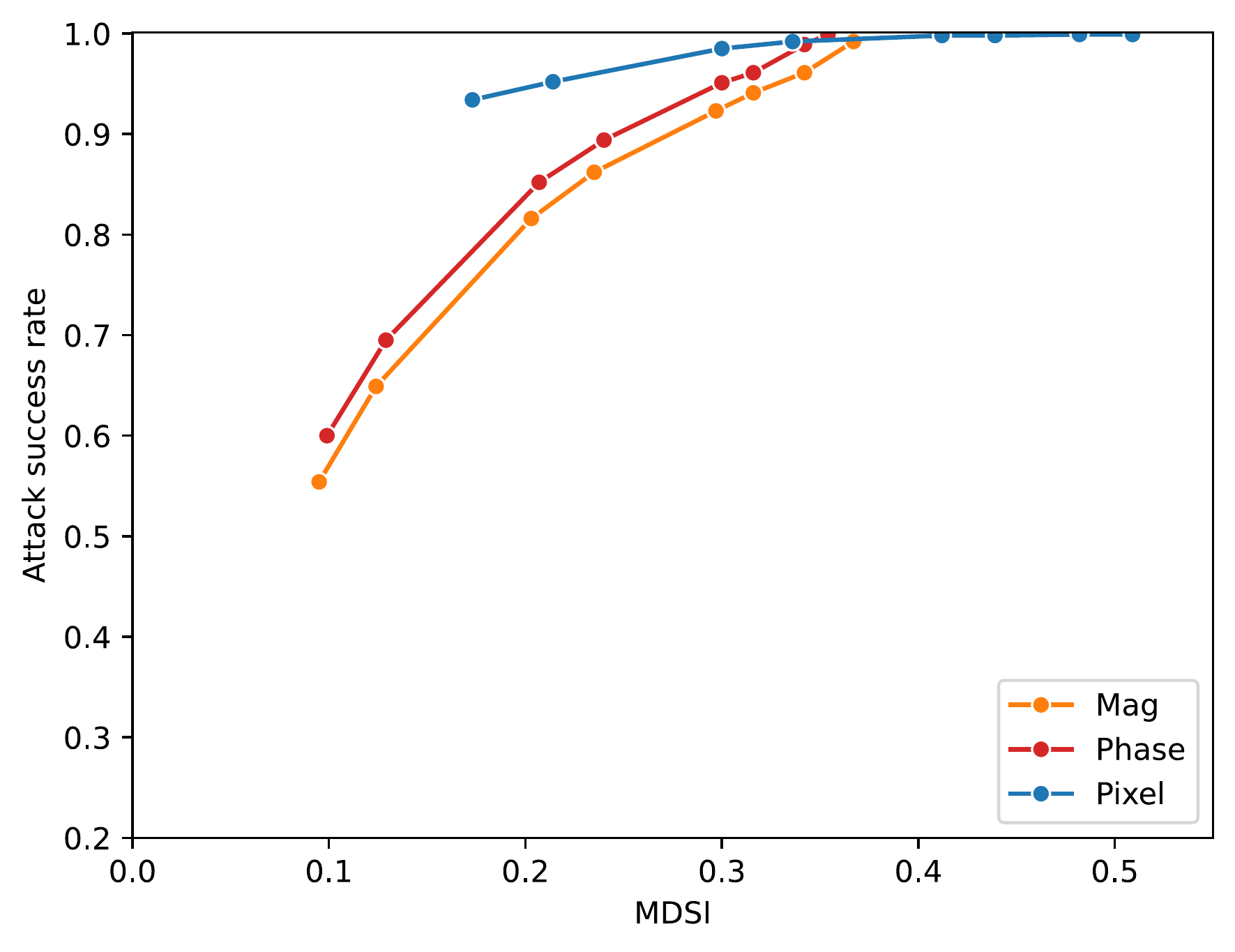}
    \caption{ResNet50}
    \label{figure10:a}
    \end{subfigure}
    \hfill
    \begin{subfigure}[t]{0.33\textwidth}
    \includegraphics[width=\textwidth]{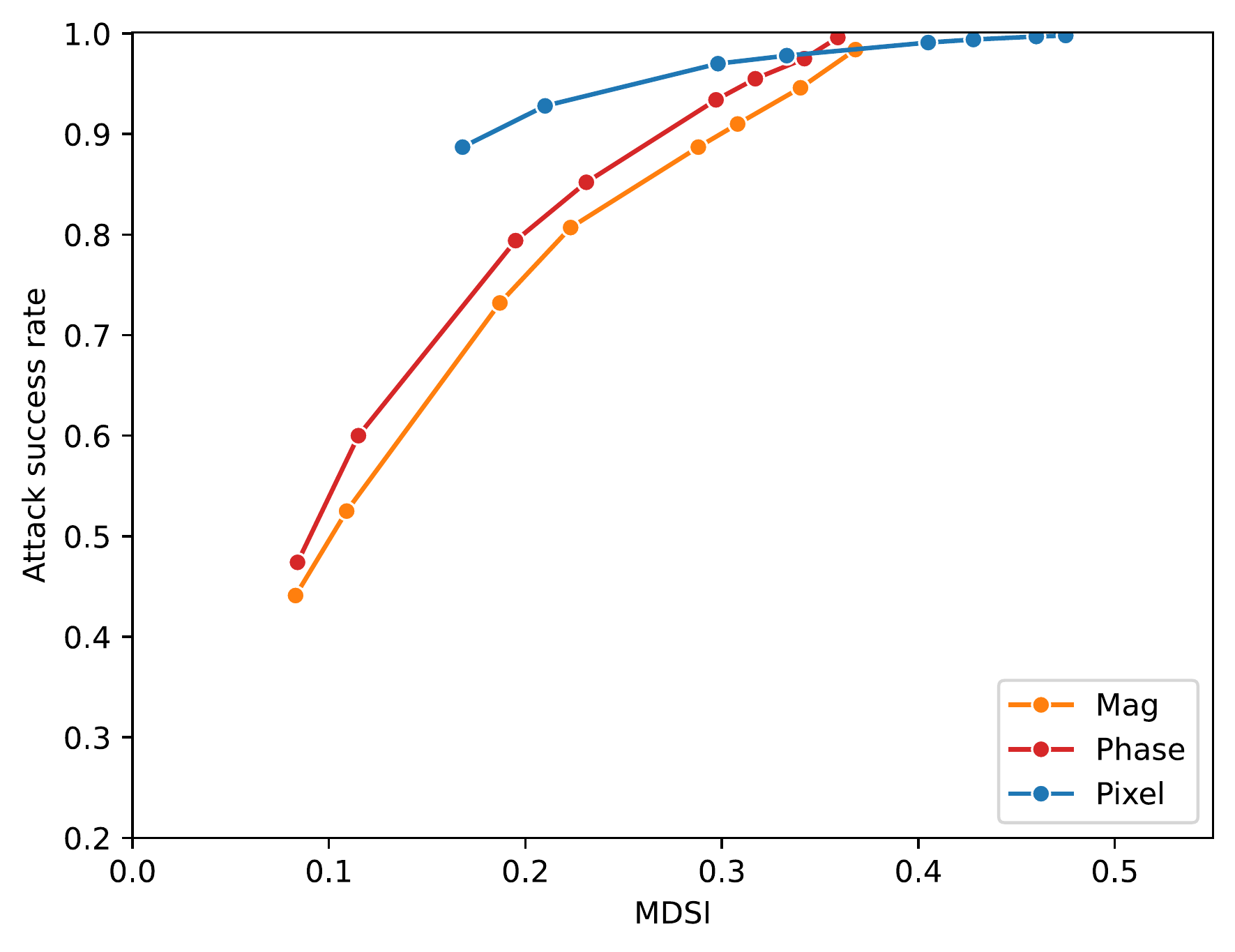}
    \caption{ResNet152}
    \label{figure10:b}
    \end{subfigure}
    \hfill
    \begin{subfigure}[t]{0.33\textwidth}
    \includegraphics[width=\textwidth]{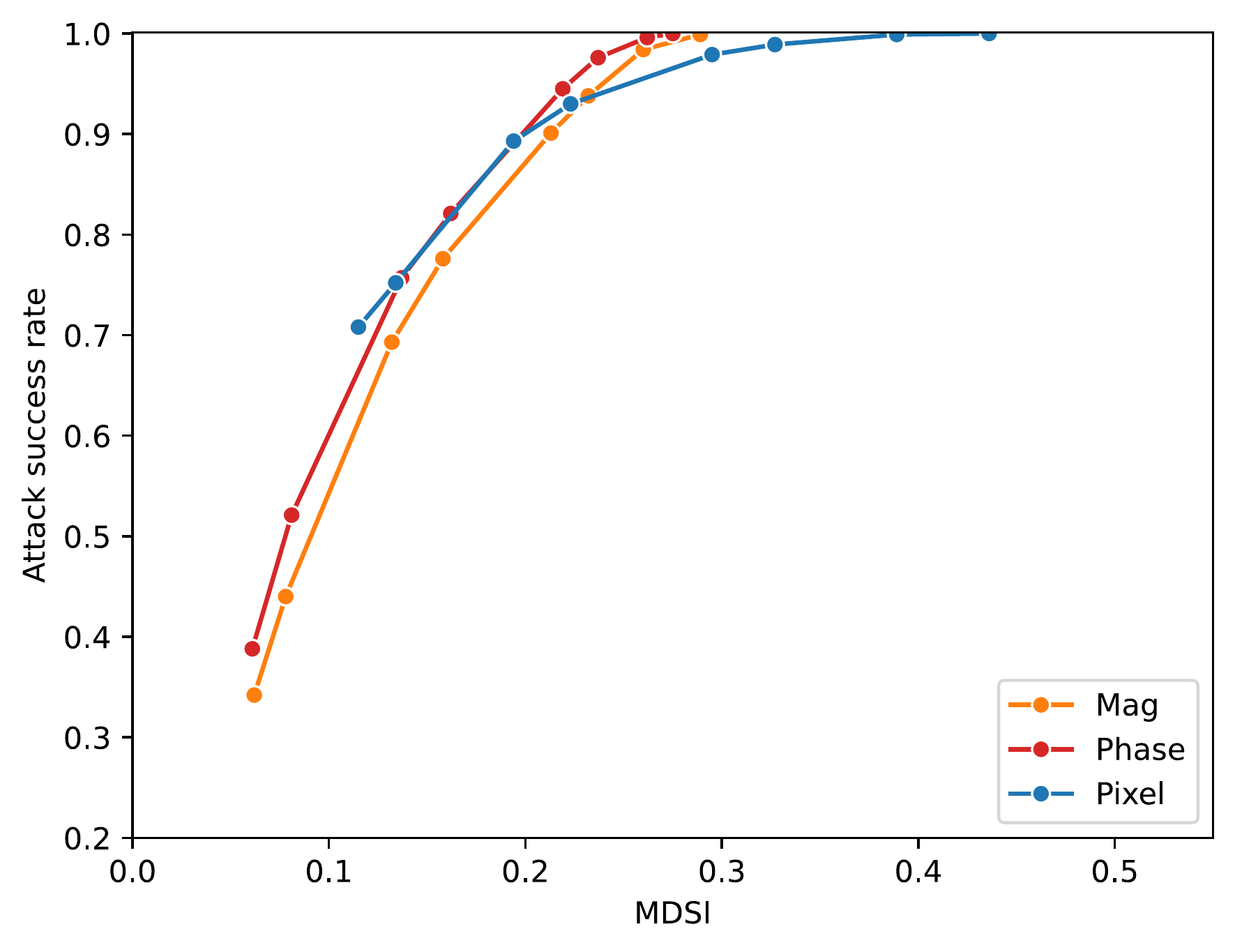}
    \caption{ViT-B}
    \label{figure10:c}
    \end{subfigure}
    \hfill
    \begin{subfigure}[t]{0.33\textwidth}
    \includegraphics[width=\textwidth]{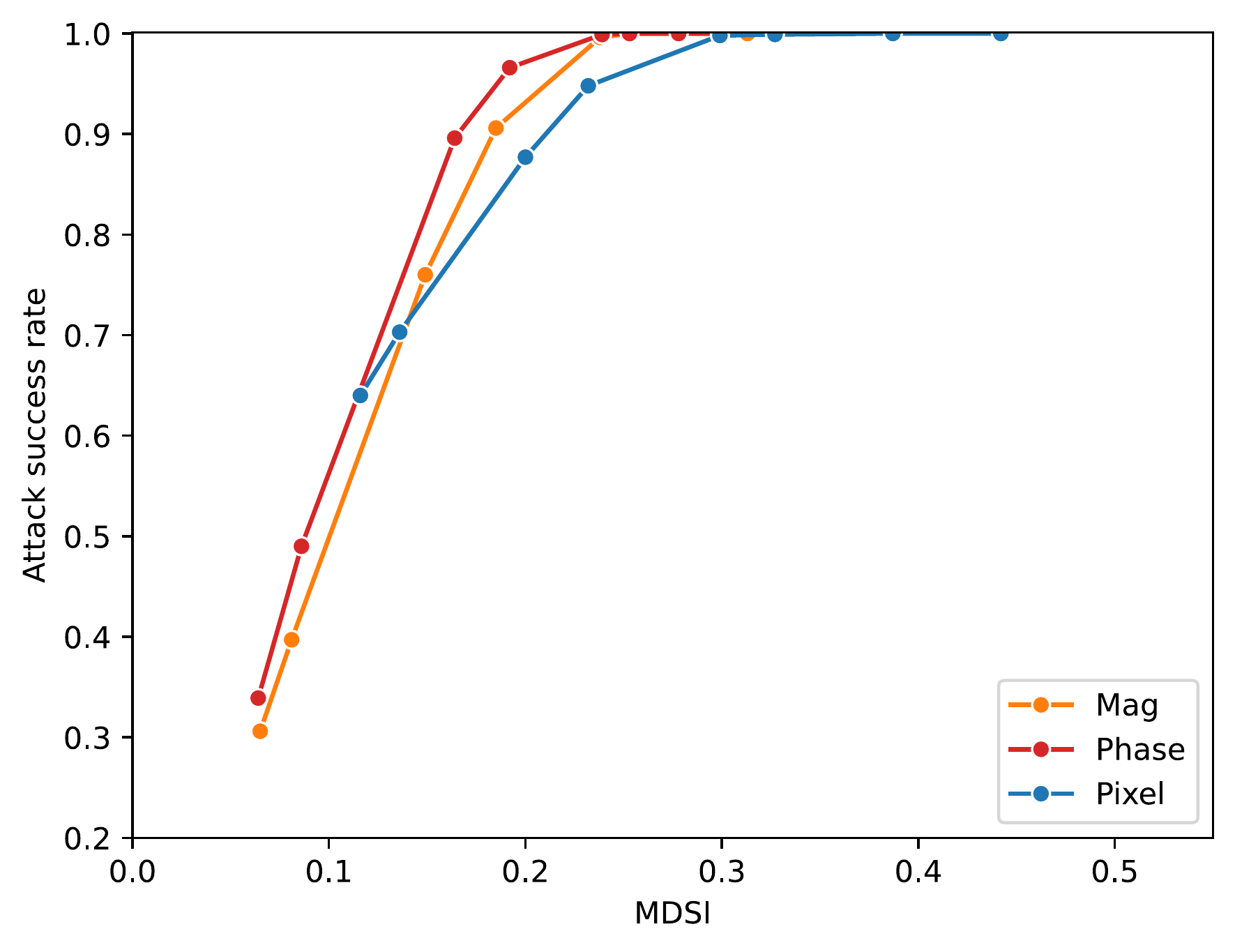}
    \caption{ViT-B-1k}
    \label{figure10:d}
    \end{subfigure}
    \hfill
    \begin{subfigure}[t]{0.33\textwidth}
    \includegraphics[width=\textwidth]{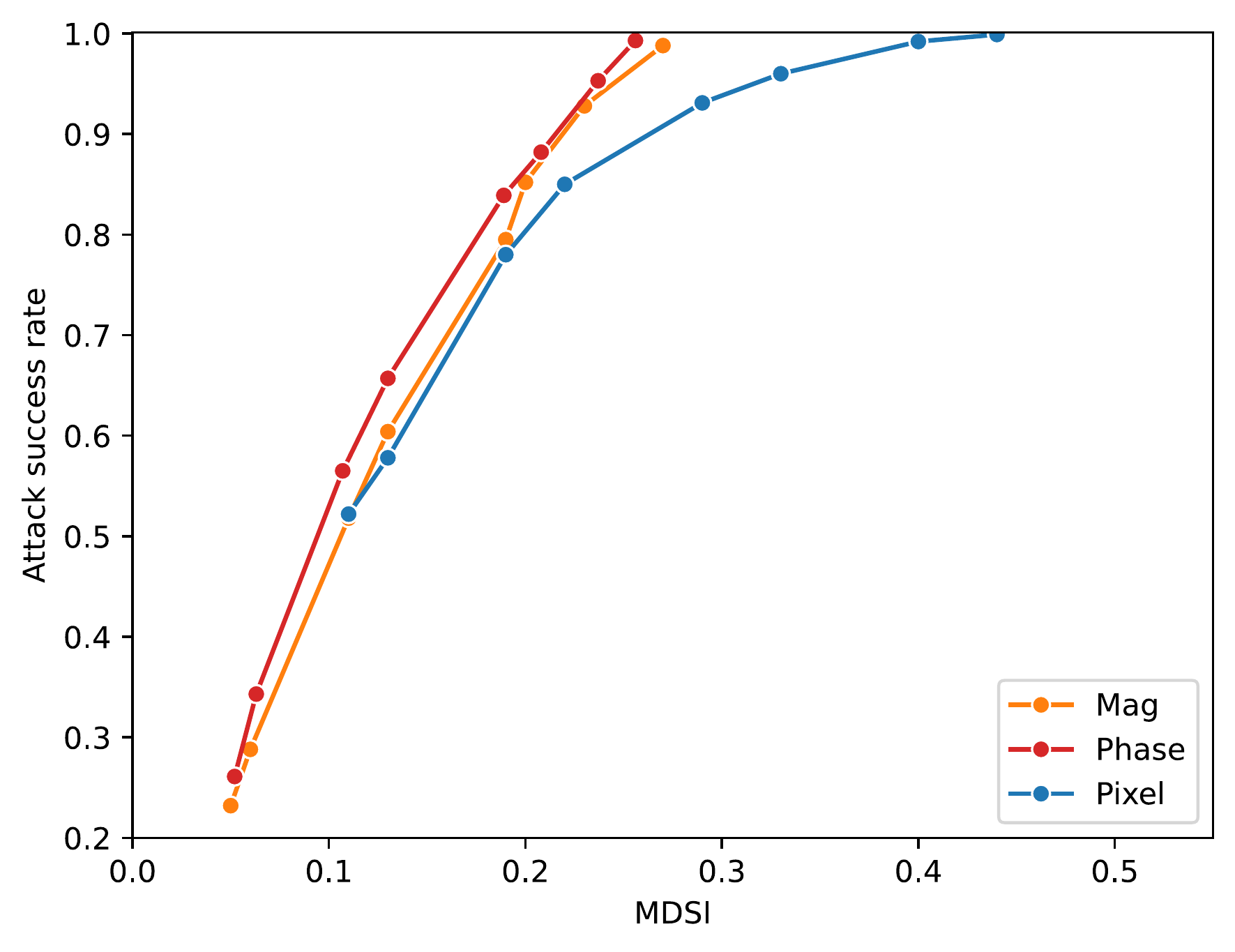}
    \caption{ViT-L}
    \label{figure10:e}
    \end{subfigure}
    \hfill
    \begin{subfigure}[t]{0.33\textwidth}
    \includegraphics[width=\textwidth]{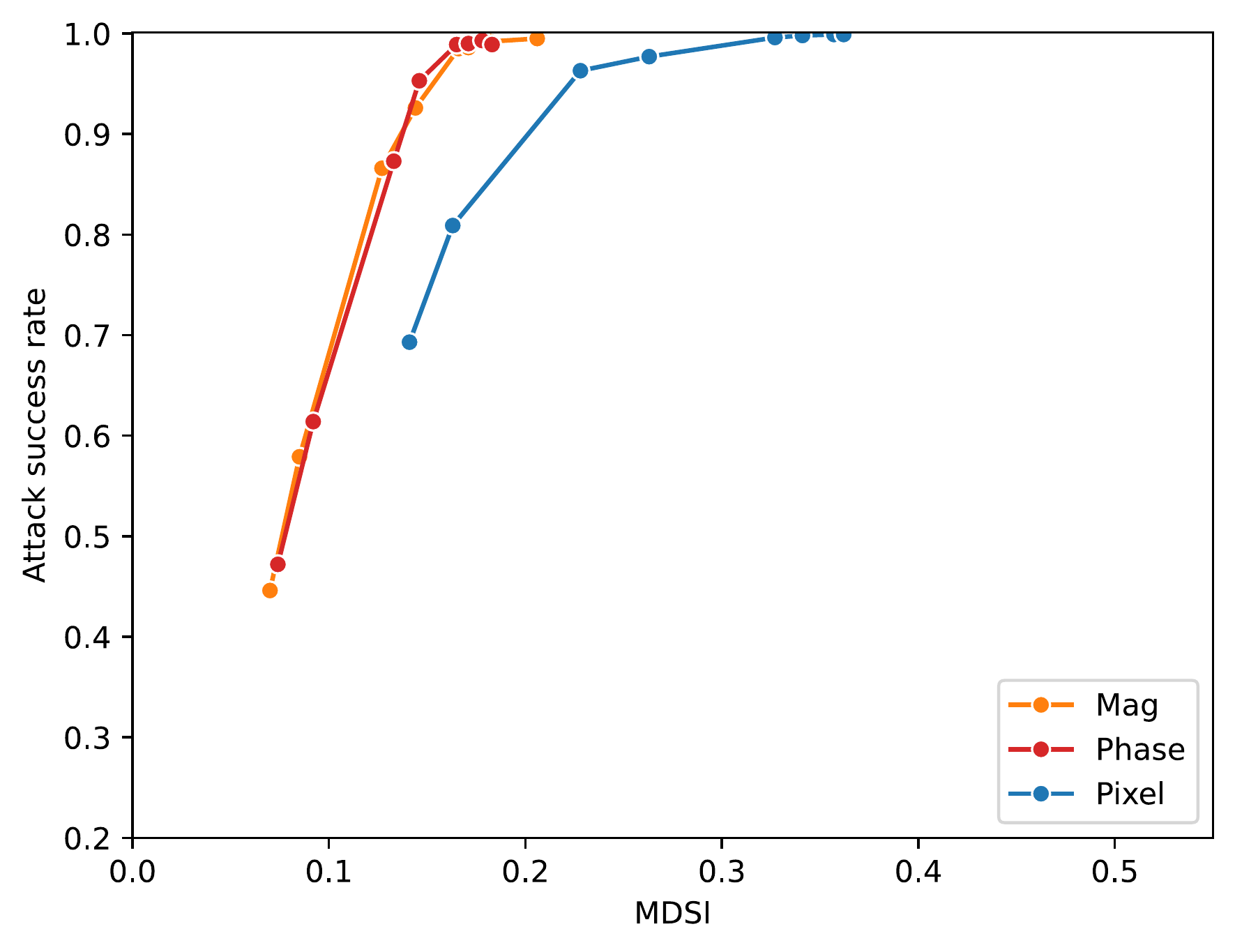}
    \caption{Swin-B}
    \label{figure10:f}
    \end{subfigure}
    \hfill
    \begin{subfigure}[t]{0.33\textwidth}
    \includegraphics[width=\textwidth]{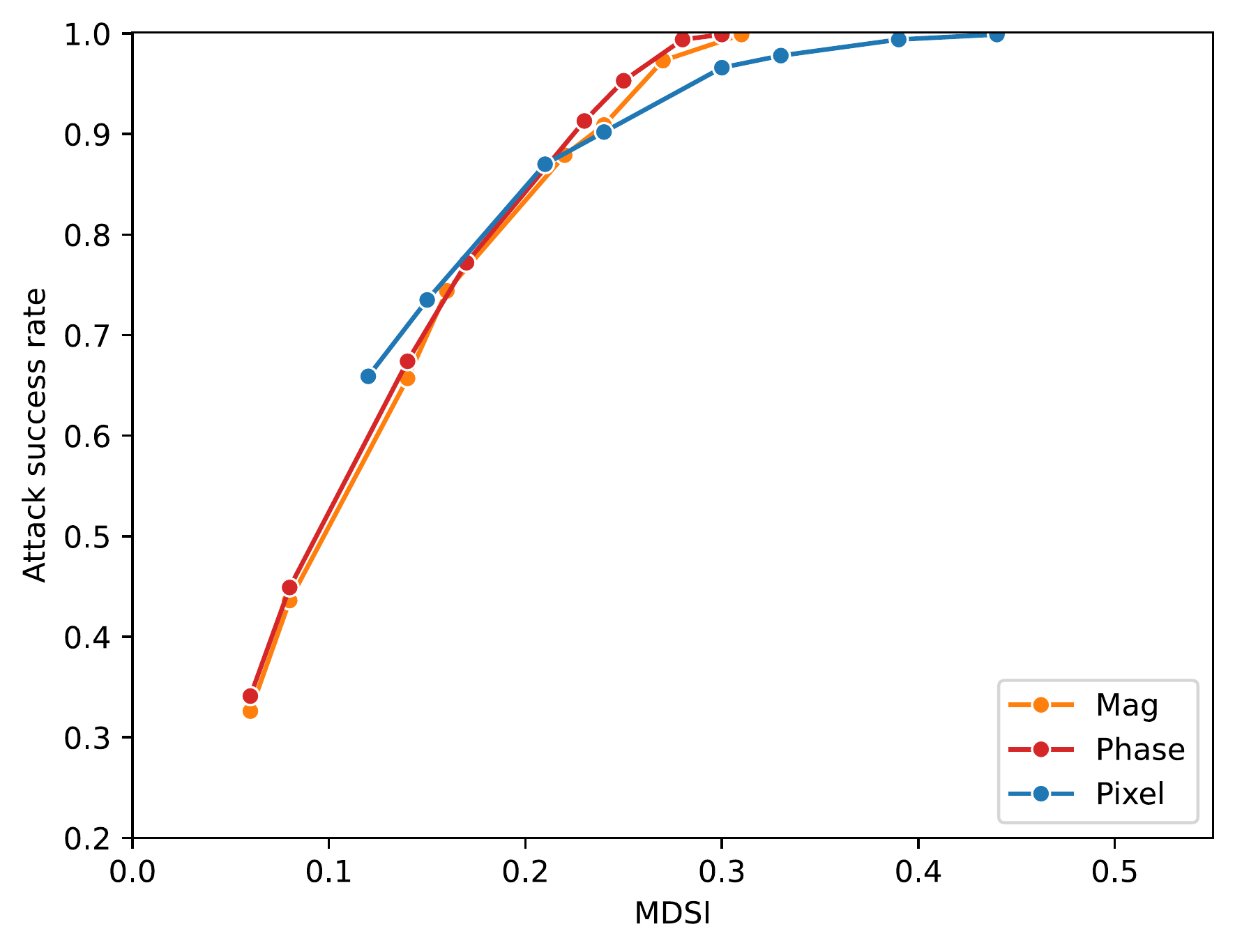}
    \caption{DeiT-S}
    \label{figure10:g}
    \end{subfigure}
    \begin{subfigure}[t]{0.33\textwidth}
    \includegraphics[width=\textwidth]{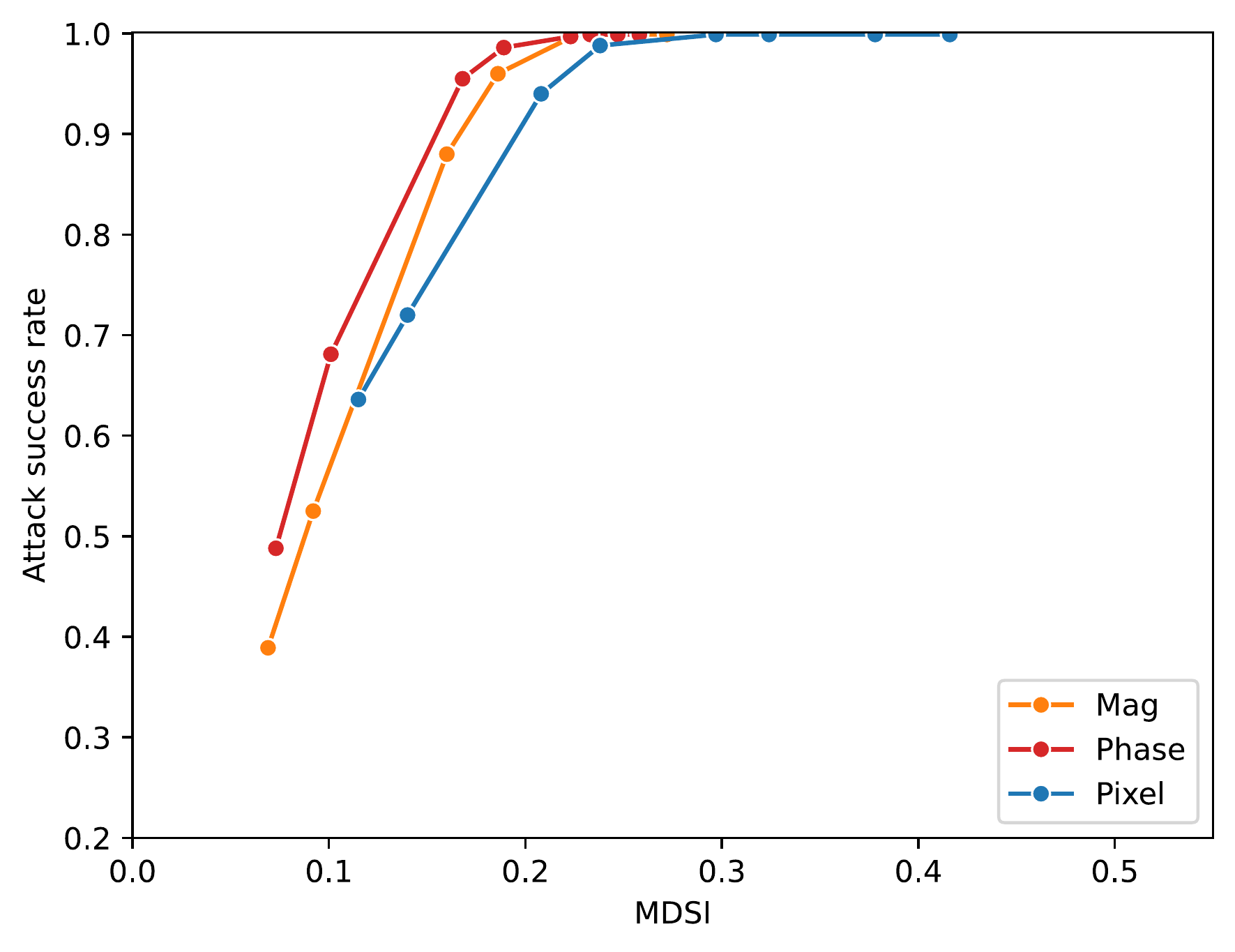}
    \caption{DeiT-S with no distillation}
    \label{figure10:h}
    \end{subfigure}
\caption{Comparison of different attacks for each model using MDSI.}
\label{figure10}
\end{figure*}

\begin{figure*}
\centering
    \begin{subfigure}[t]{0.33\textwidth}
    \includegraphics[width=\textwidth]{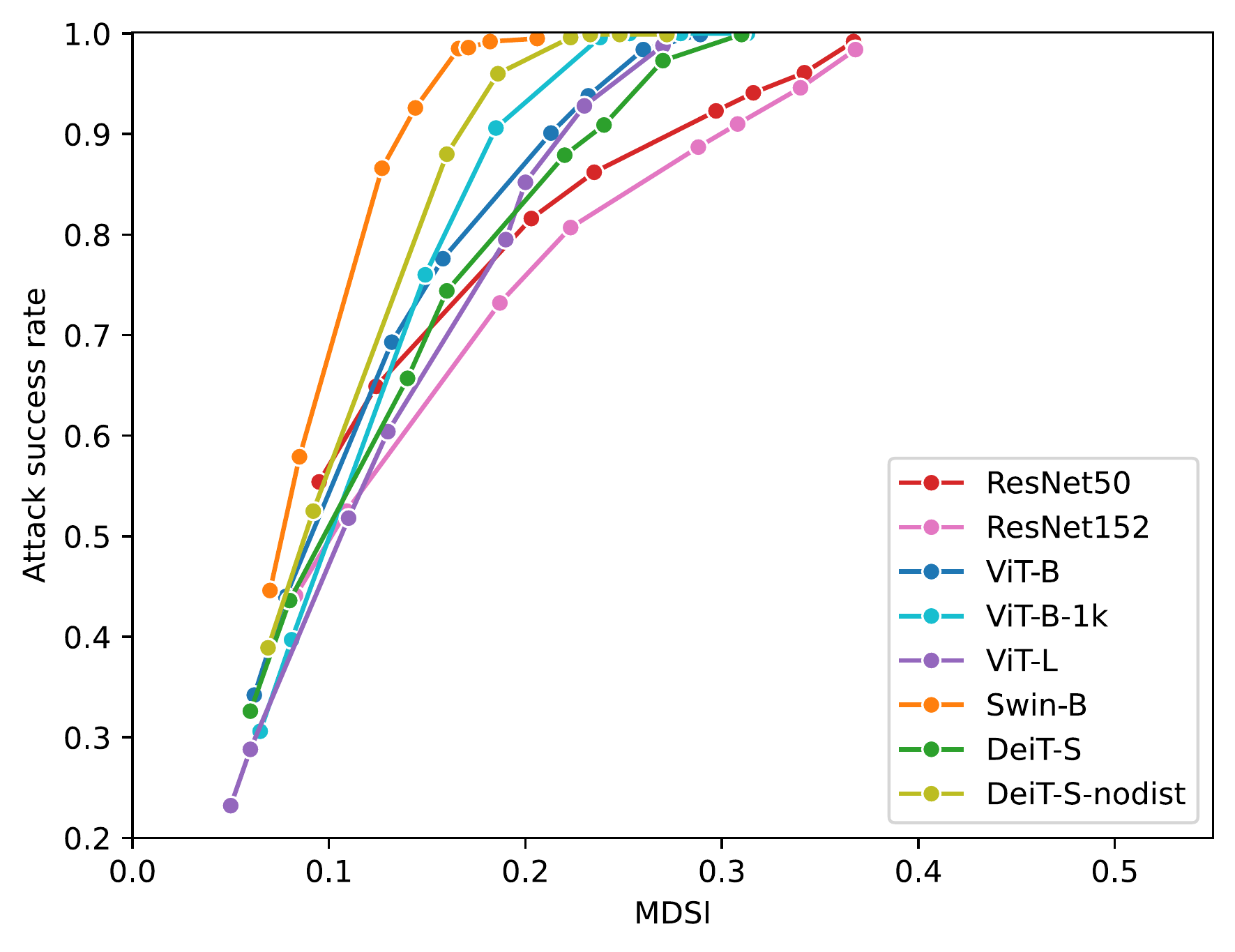}
    \caption{Magnitude attack}\label{figure11:a}
    \end{subfigure}
    \hfill
    \begin{subfigure}[t]{0.33\textwidth}
    \includegraphics[width=\textwidth]{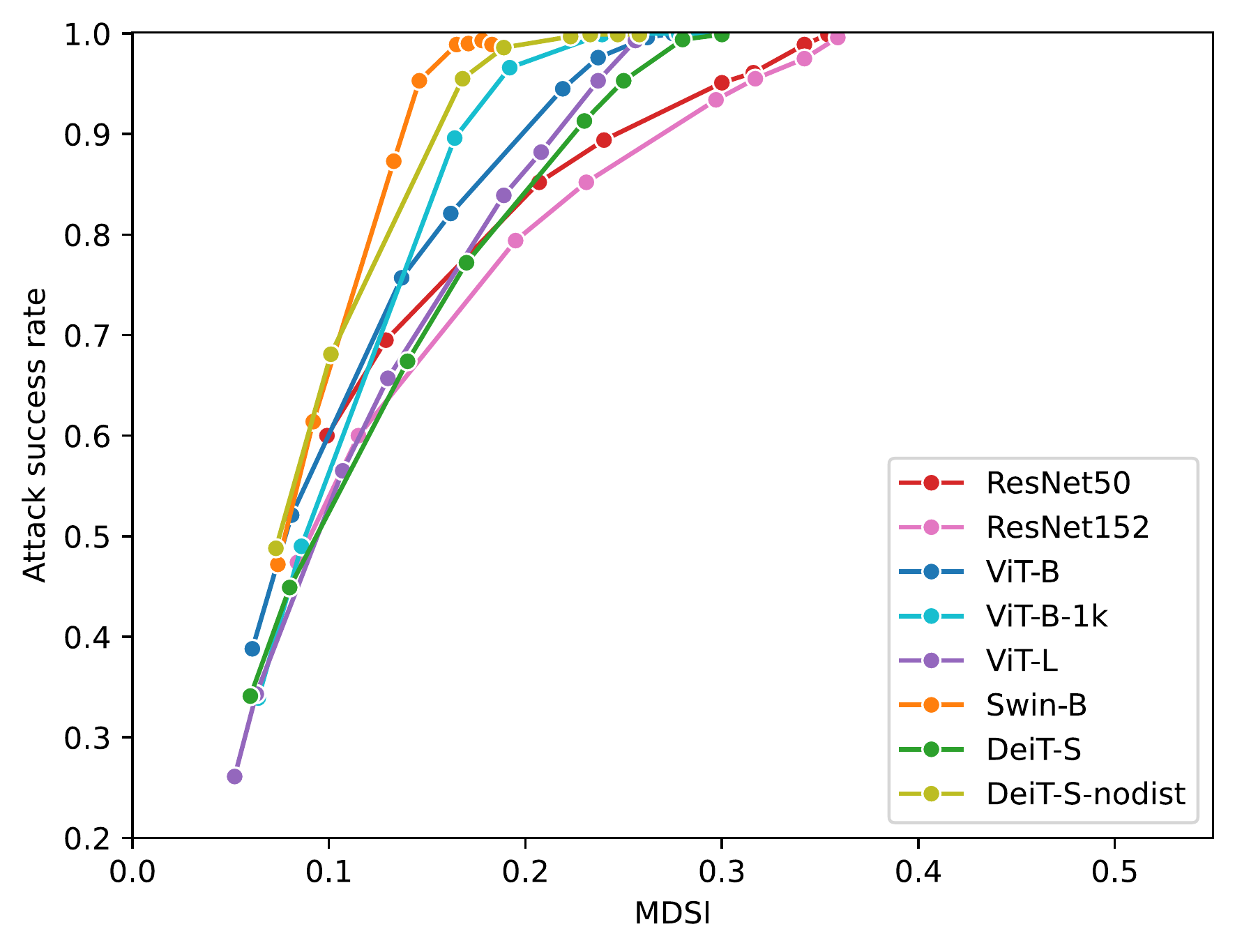}
    \caption{Phase attack}\label{figure11:b}
    \end{subfigure}
    \hfill
    \begin{subfigure}[t]{0.33\textwidth}
    \includegraphics[width=\textwidth]{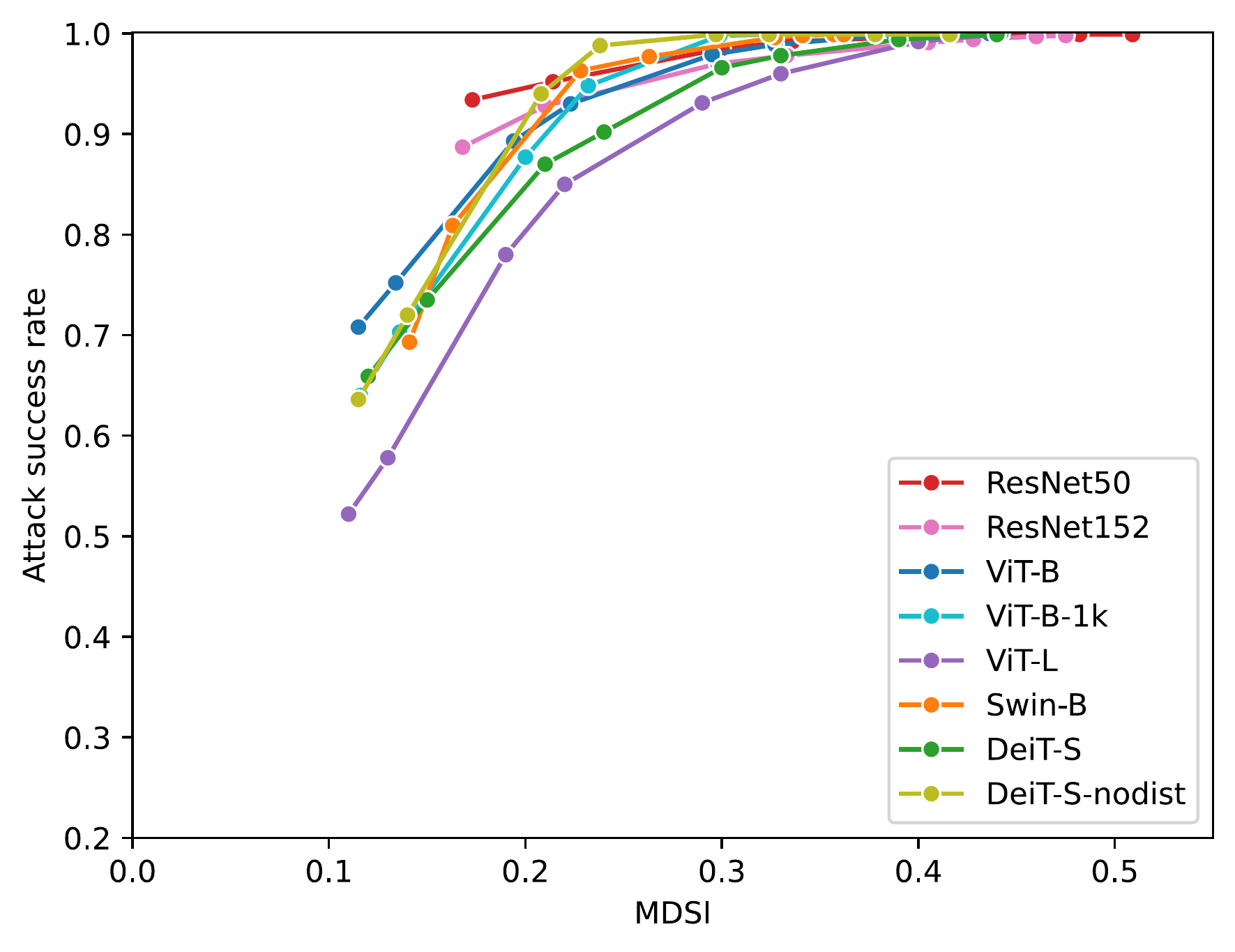}
    \caption{Pixel attack}\label{figure11:c}
    \end{subfigure}
\caption{Comparison of different models for each attack type using MDSI.}
\label{figure11}
\end{figure*}

\begin{figure*}
\centering
    \begin{subfigure}[t]{0.33\textwidth}
    \includegraphics[width=\textwidth]{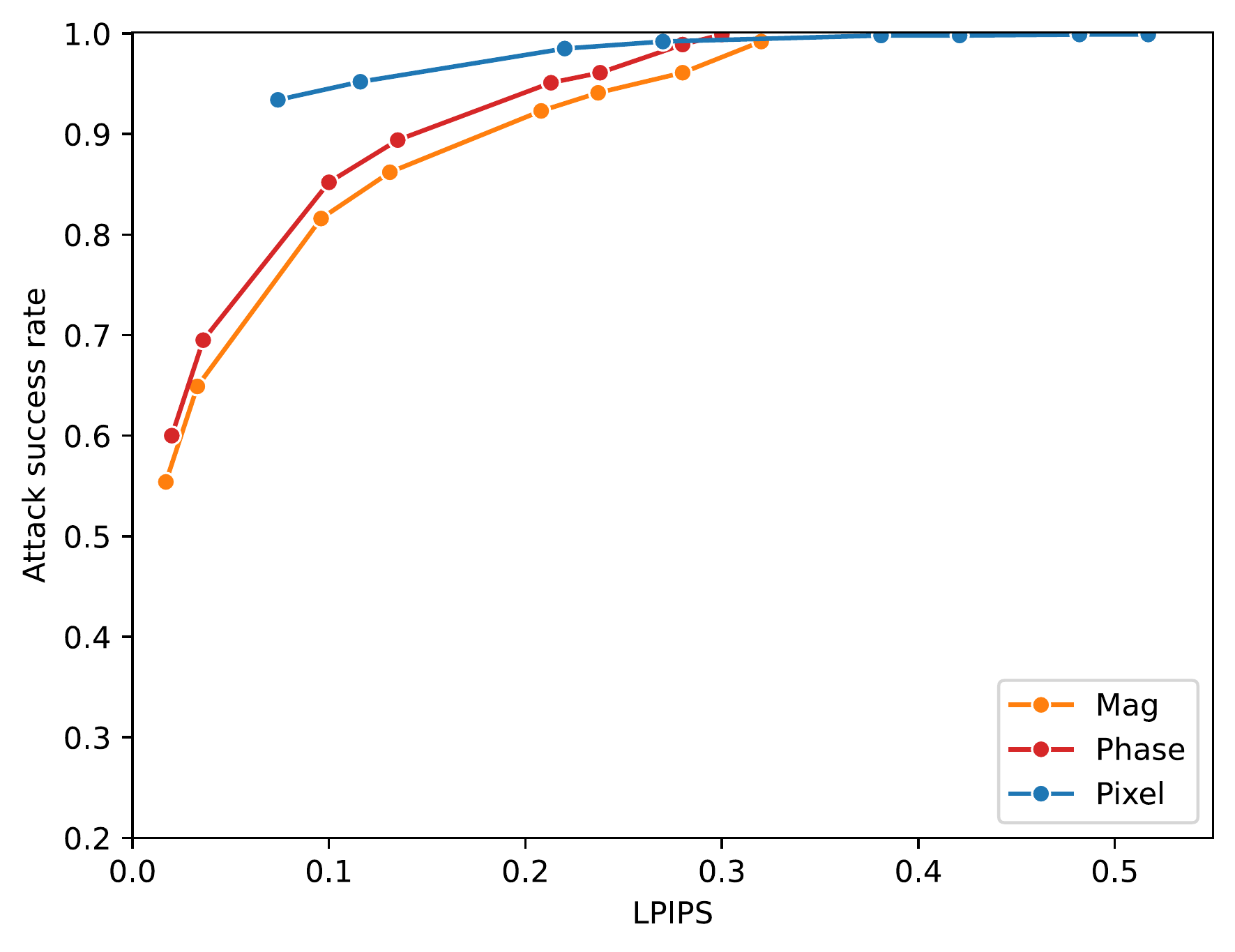}
    \caption{ResNet50}
    \label{figure12:a}
    \end{subfigure}
    \hfill
    \begin{subfigure}[t]{0.33\textwidth}
    \includegraphics[width=\textwidth]{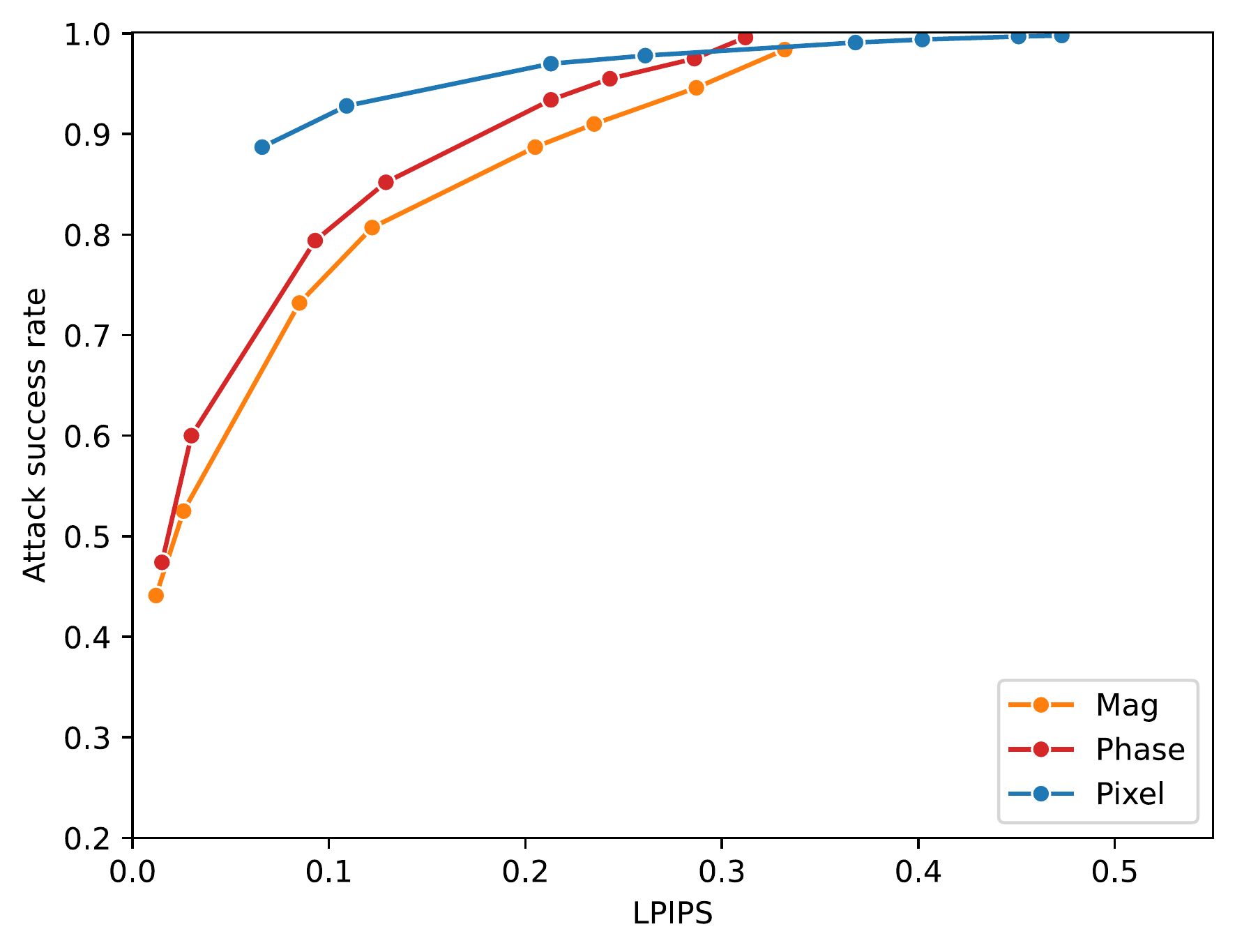}
    \caption{ResNet152}
    \label{figure12:b}
    \end{subfigure}
    \hfill
    \begin{subfigure}[t]{0.33\textwidth}
    \includegraphics[width=\textwidth]{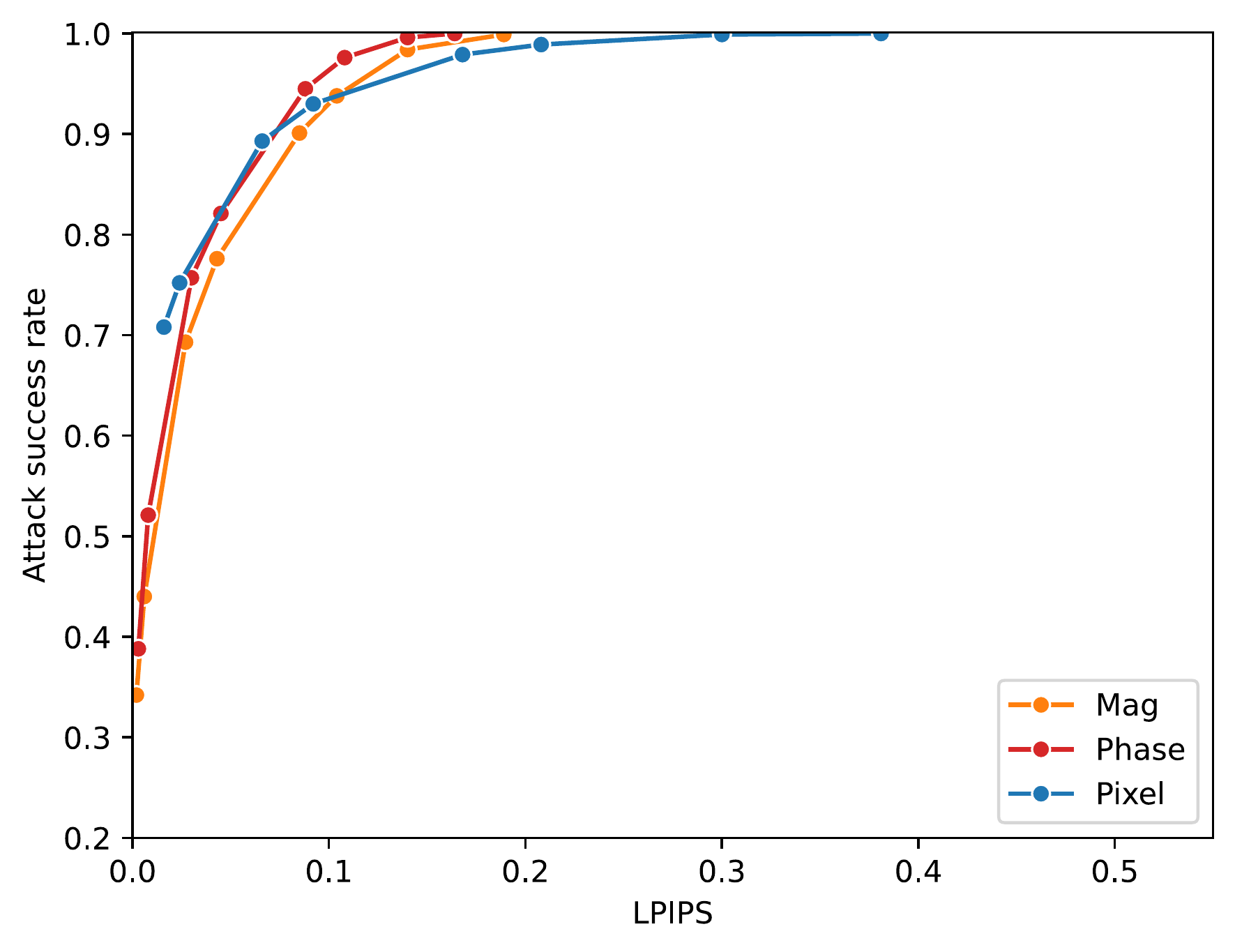}
    \caption{ViT-B}
    \label{figure12:c}
    \end{subfigure}
    \hfill
    \begin{subfigure}[t]{0.33\textwidth}
    \includegraphics[width=\textwidth]{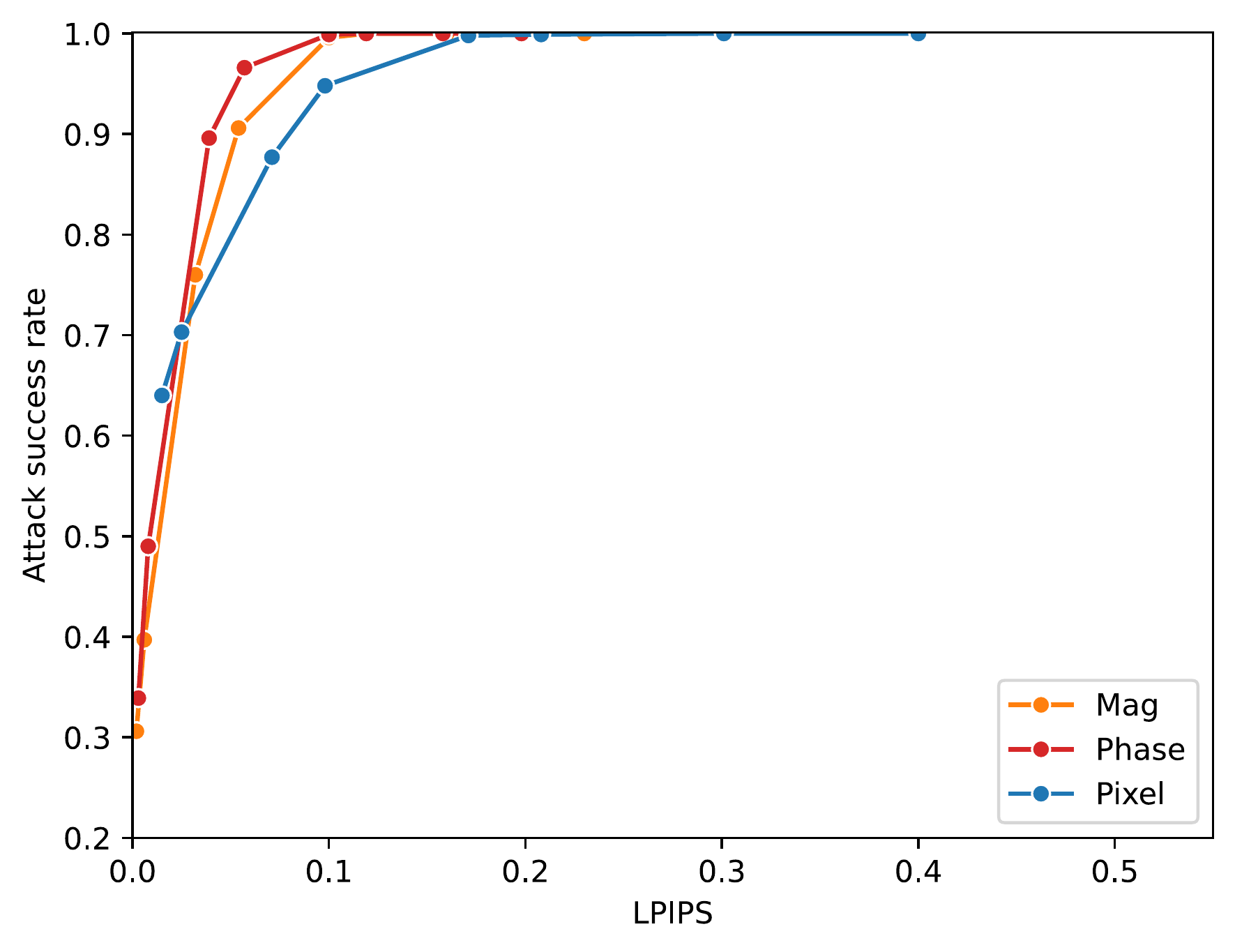}
    \caption{ViT-B-1k}
    \label{figure12:d}
    \end{subfigure}
    \hfill
    \begin{subfigure}[t]{0.33\textwidth}
    \includegraphics[width=\textwidth]{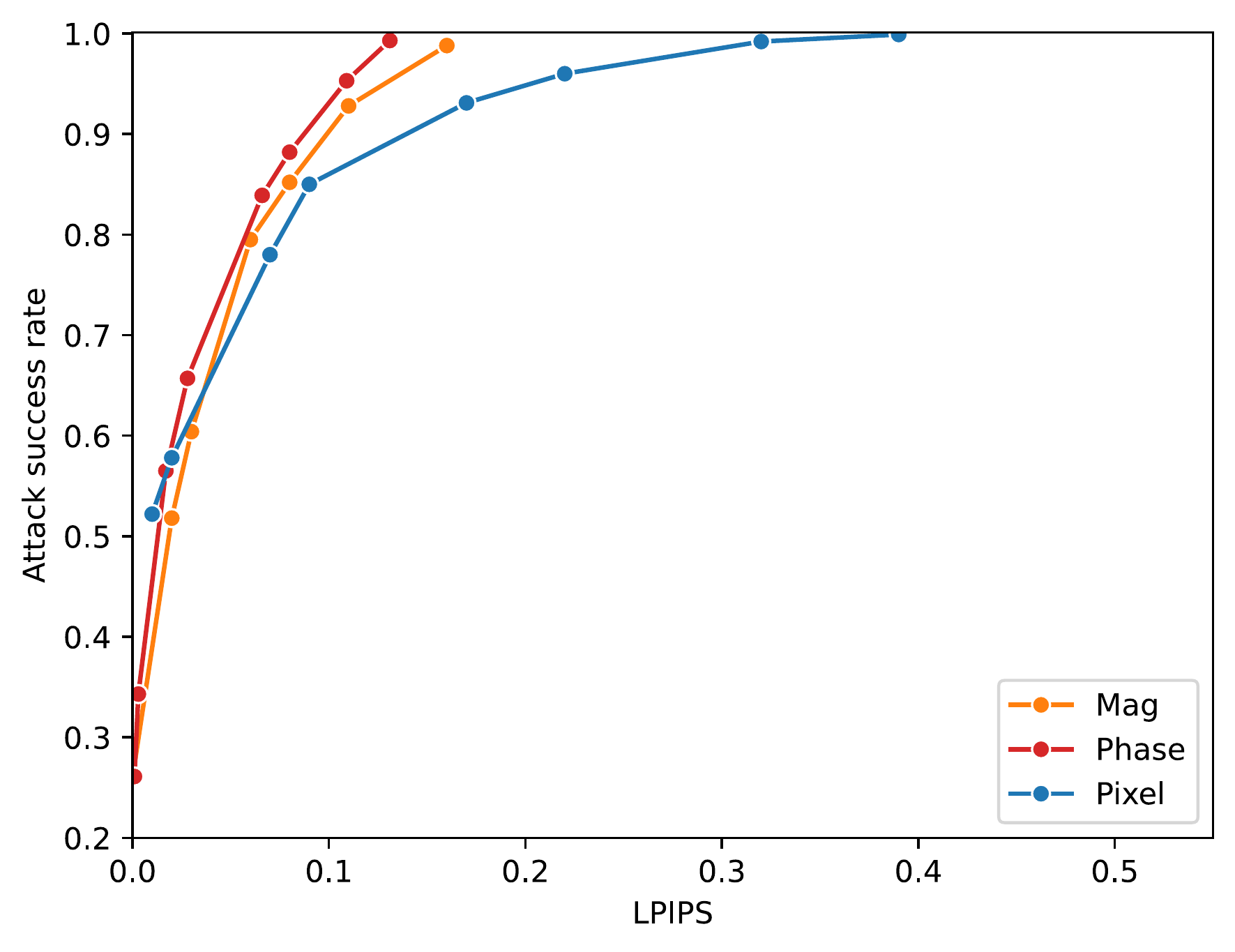}
    \caption{ViT-L}
    \label{figure12:e}
    \end{subfigure}
    \hfill
    \begin{subfigure}[t]{0.33\textwidth}
    \includegraphics[width=\textwidth]{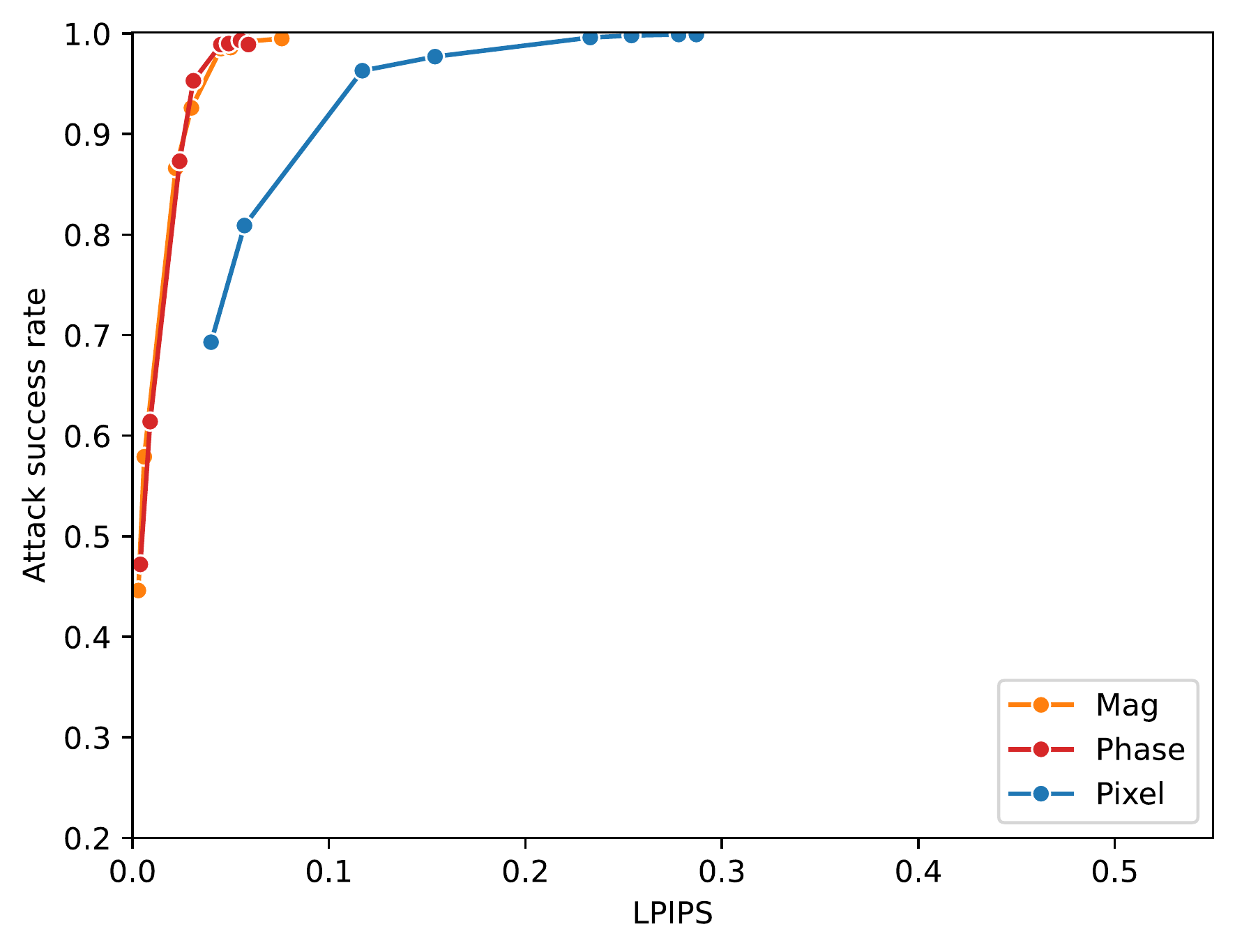}
    \caption{Swin-B}
    \label{figure12:f}
    \end{subfigure}
    \hfill
    \begin{subfigure}[t]{0.33\textwidth}
    \includegraphics[width=\textwidth]{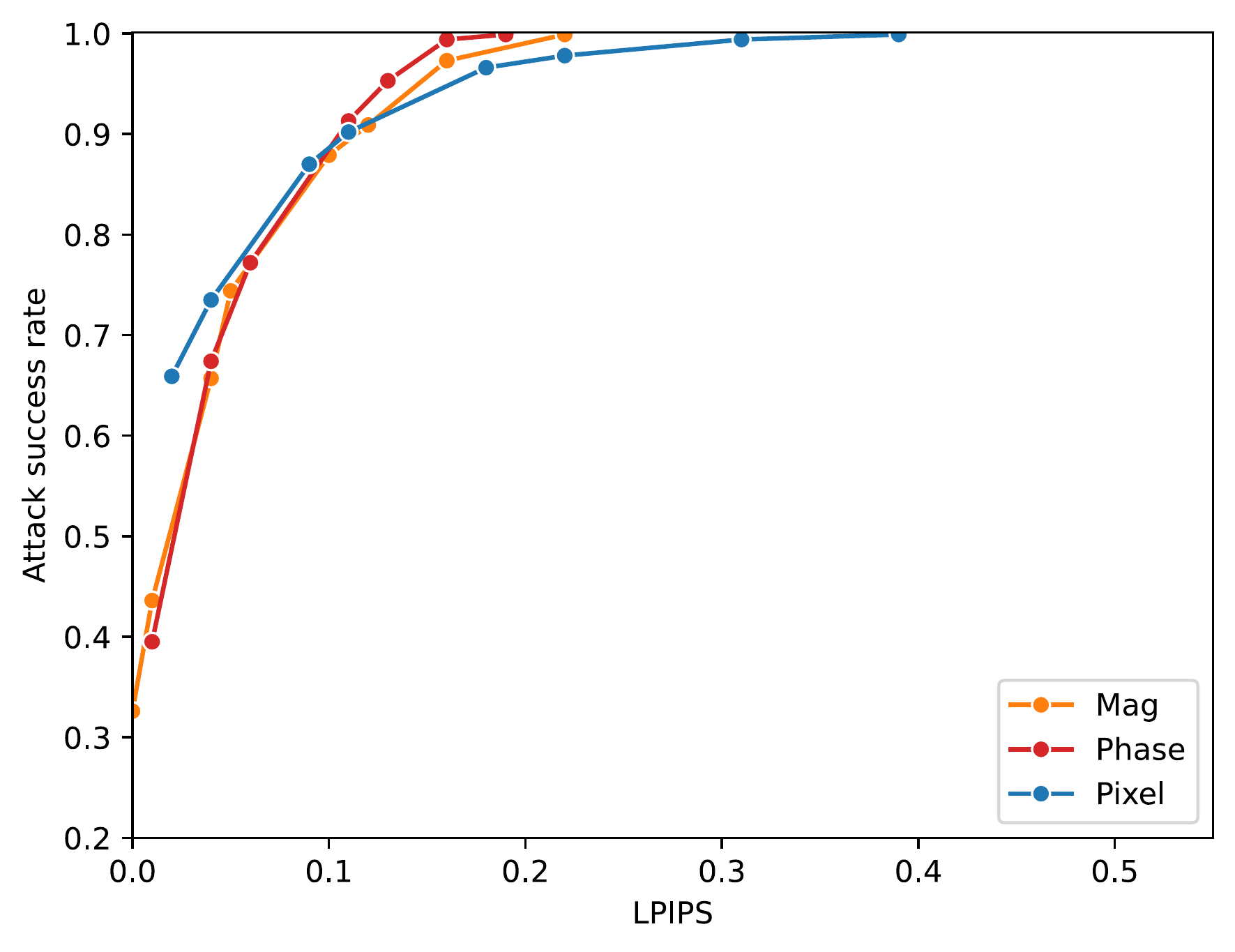}
    \caption{DeiT-S}
    \label{figure12:g}
    \end{subfigure}
    \begin{subfigure}[t]{0.33\textwidth}
    \includegraphics[width=\textwidth]{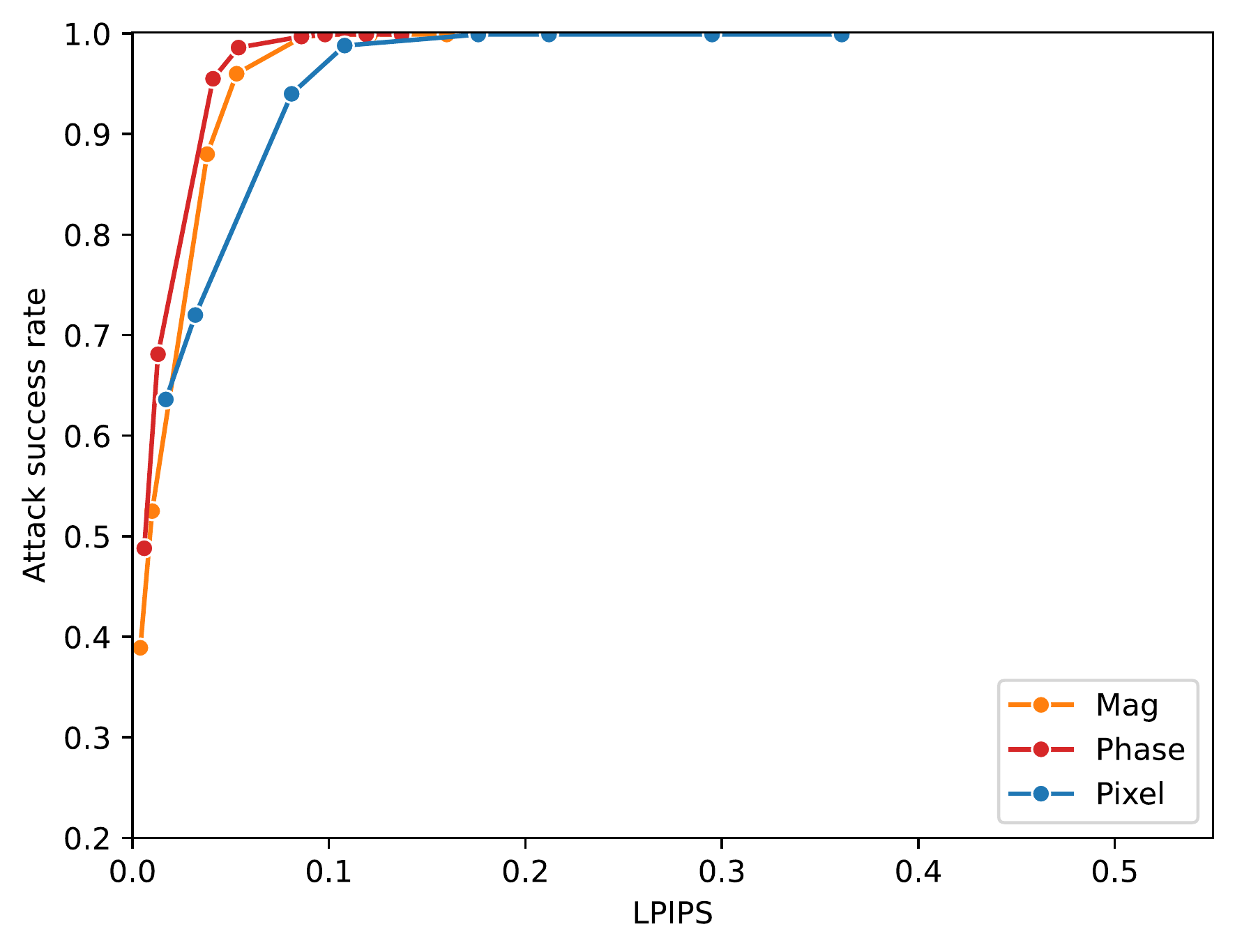}
    \caption{DeiT-S with no distillation}
    \label{figure12:h}
    \end{subfigure}
\caption{Comparison of different attacks for each model using LPIPS.}
\label{figure12}
\end{figure*}

\begin{figure*}
\centering
    \begin{subfigure}[t]{0.33\textwidth}
    \includegraphics[width=\textwidth]{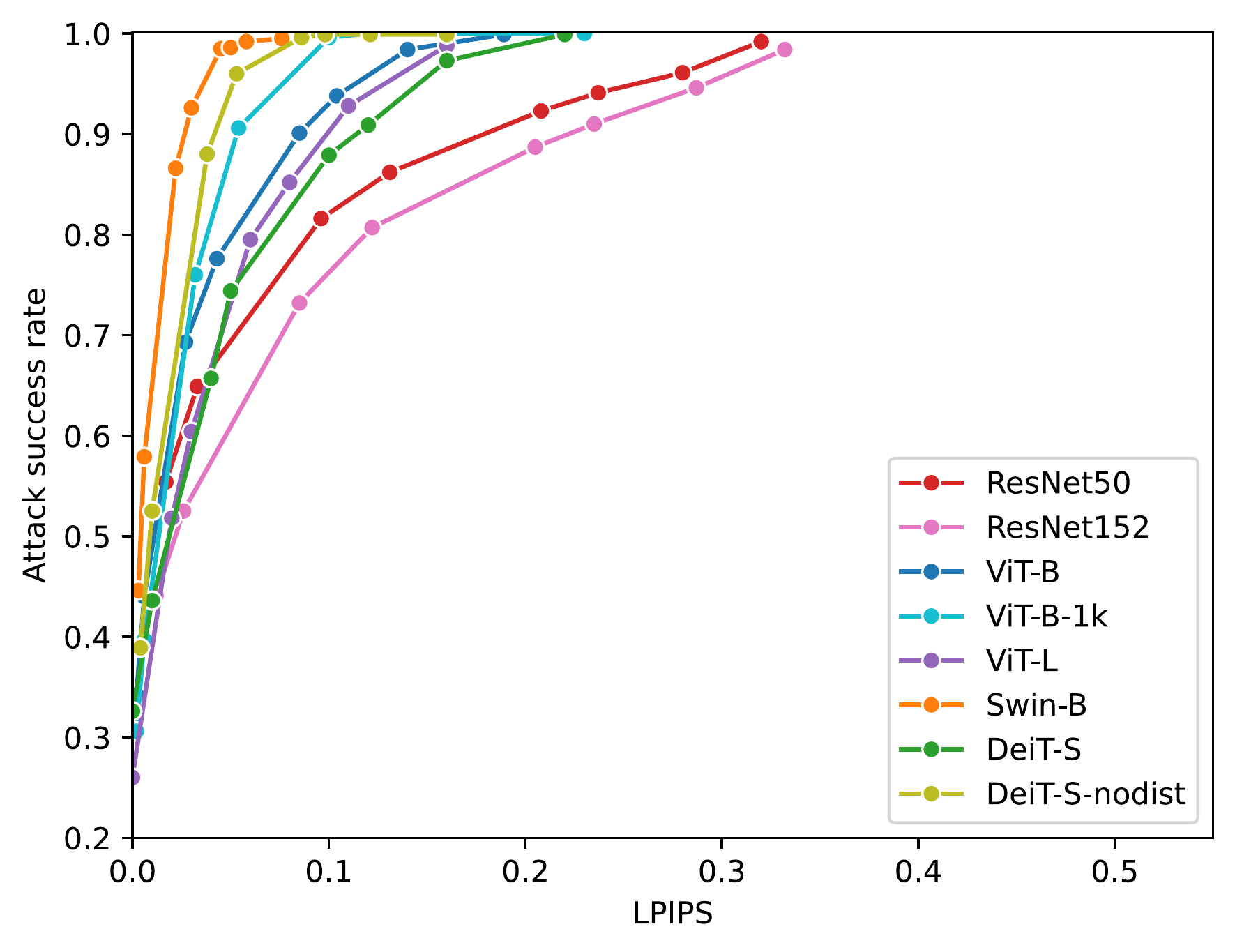}
    \caption{Magnitude attack}\label{figure13:a}
    \end{subfigure}
    \hfill
    \begin{subfigure}[t]{0.33\textwidth}
    \includegraphics[width=\textwidth]{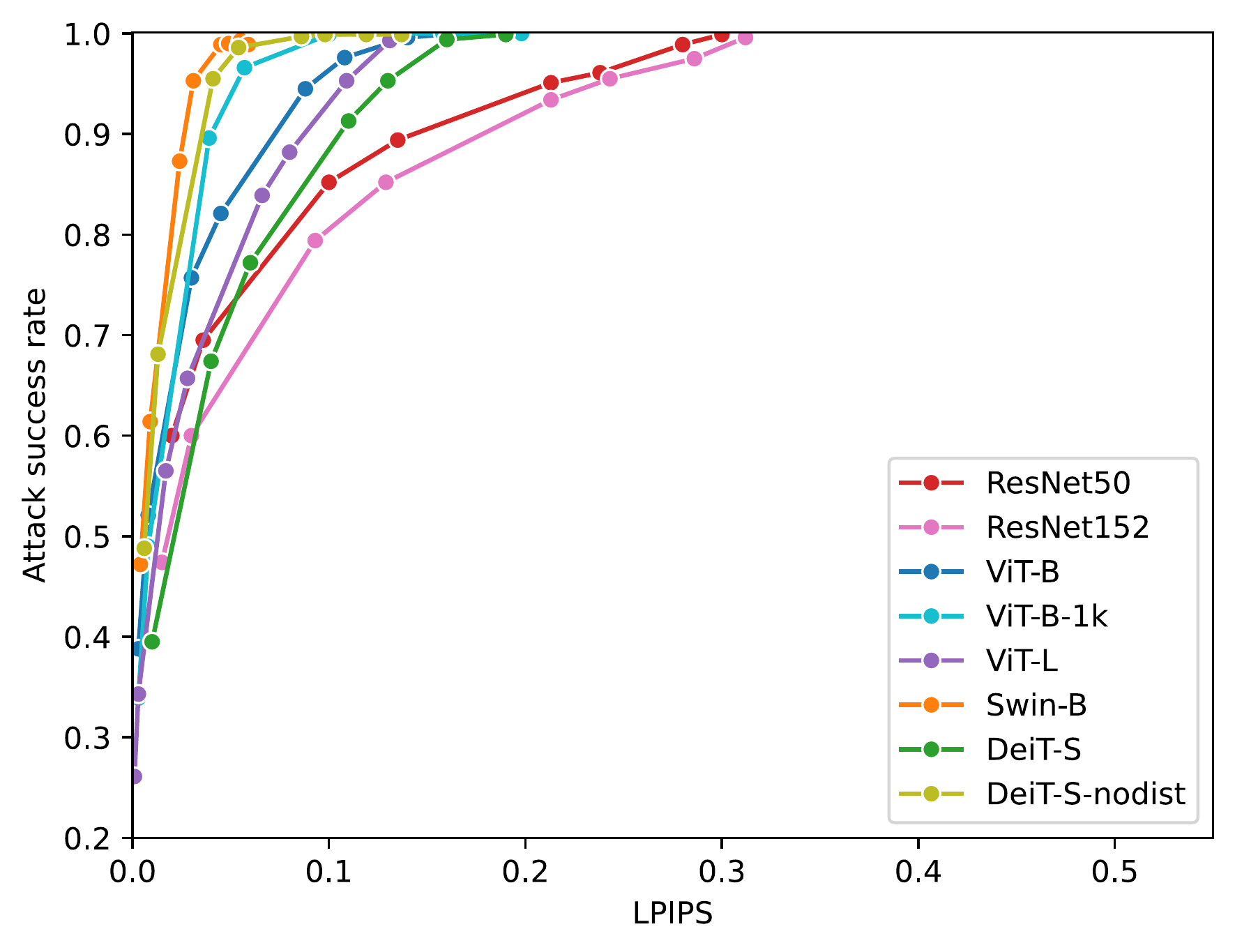}
    \caption{Phase attack}\label{figure13:b}
    \end{subfigure}
    \hfill
    \begin{subfigure}[t]{0.33\textwidth}
    \includegraphics[width=\textwidth]{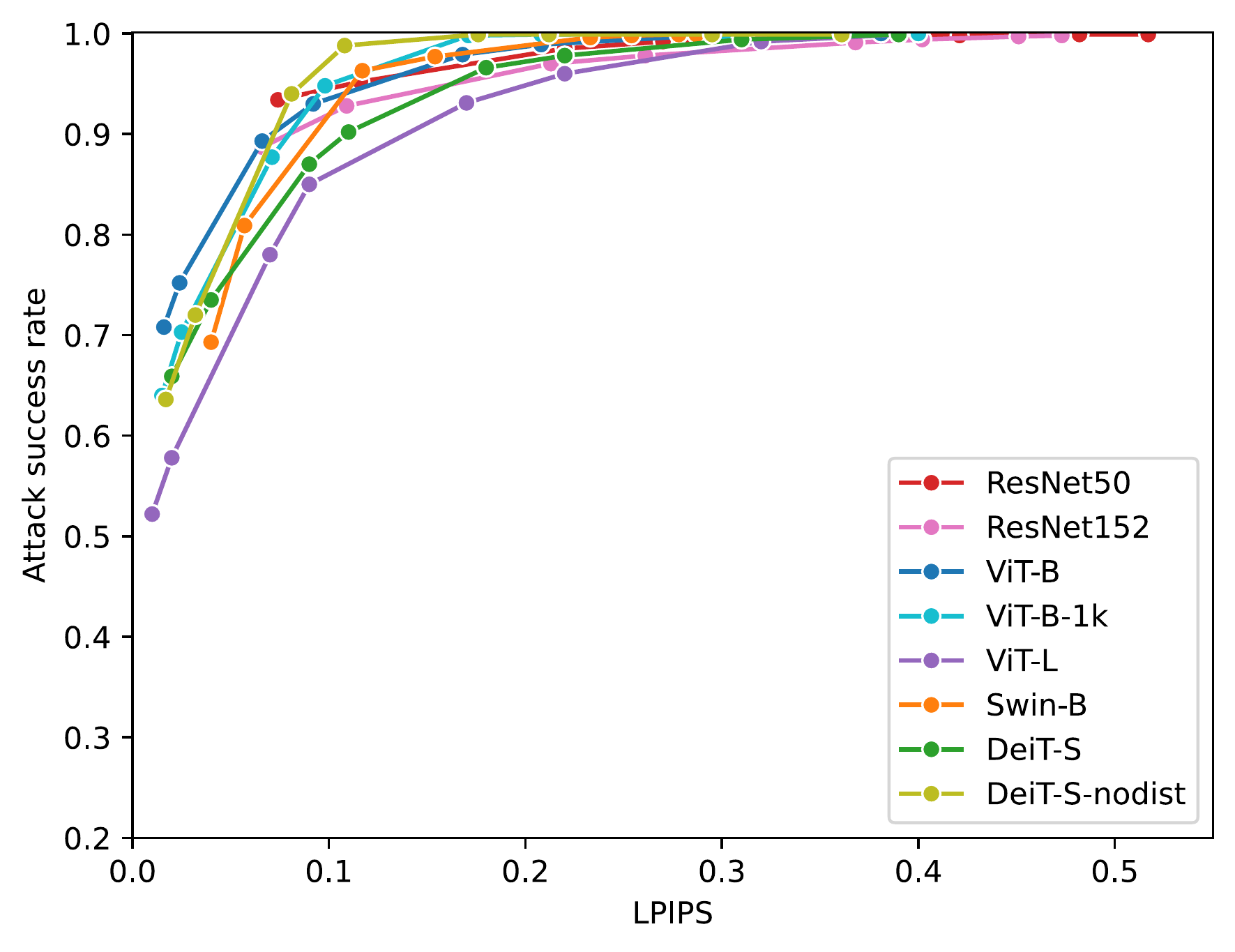}
    \caption{Pixel attack}\label{figure13:c}
    \end{subfigure}
\caption{Comparison of different models for each attack type using LPIPS.}
\label{figure13}
\end{figure*}

\noindent While we use PSNR as the image quality metric in the main paper, results using other perceptual image quality metrics are shown here. We choose the multi-scale structural similarity index measure (MS-SSIM) \cite{a:12}, which is advanced form of SSIM that is one of the most popular perceptual metrics, and the mean deviation similarity index (MDSI) \cite{a:5}, which was shown to perform the best in \cite{m:12}, and learned perceptual image patch similarity (LPIPS) \cite{i:18}, which uses the features from a pre-trained network for classification. The results are shown in Figures \ref{figure8}-\ref{figure13}. Note that a larger value of MDSI or LPIPS indicates a lower quality level. Overall, the trends remain similar to those shown in Figures 3 and 4 of the main paper. 

\section{Comparison of Models Pre-trained on ImageNet-1k}
\begin{figure*}
\centering
\begin{subfigure}[t]{0.33\textwidth}
\includegraphics[width=\textwidth]{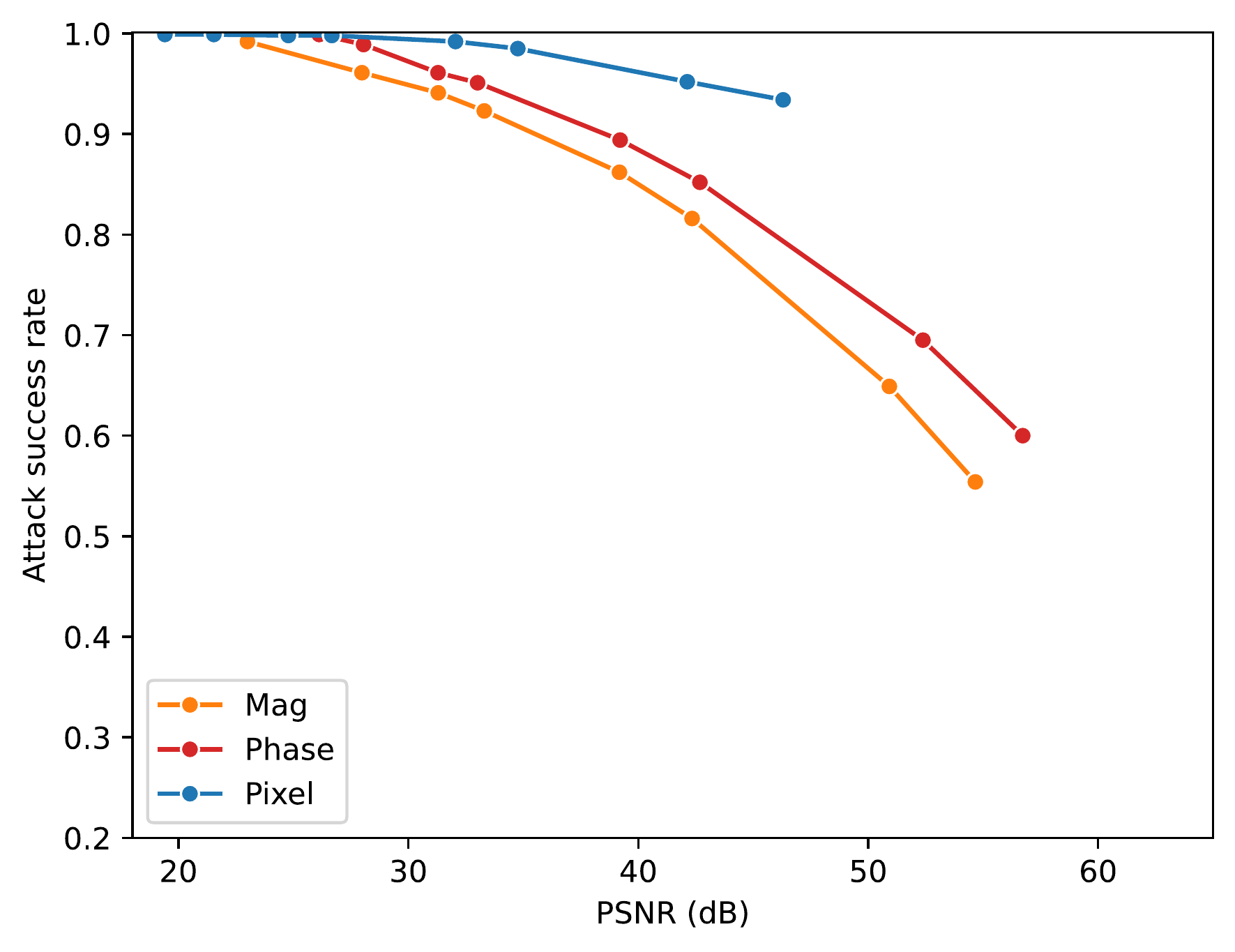}
\caption{ResNet50}\label{figure14:a}
\end{subfigure}
\hfill
\begin{subfigure}[t]{0.33\textwidth}
\includegraphics[width=\textwidth]{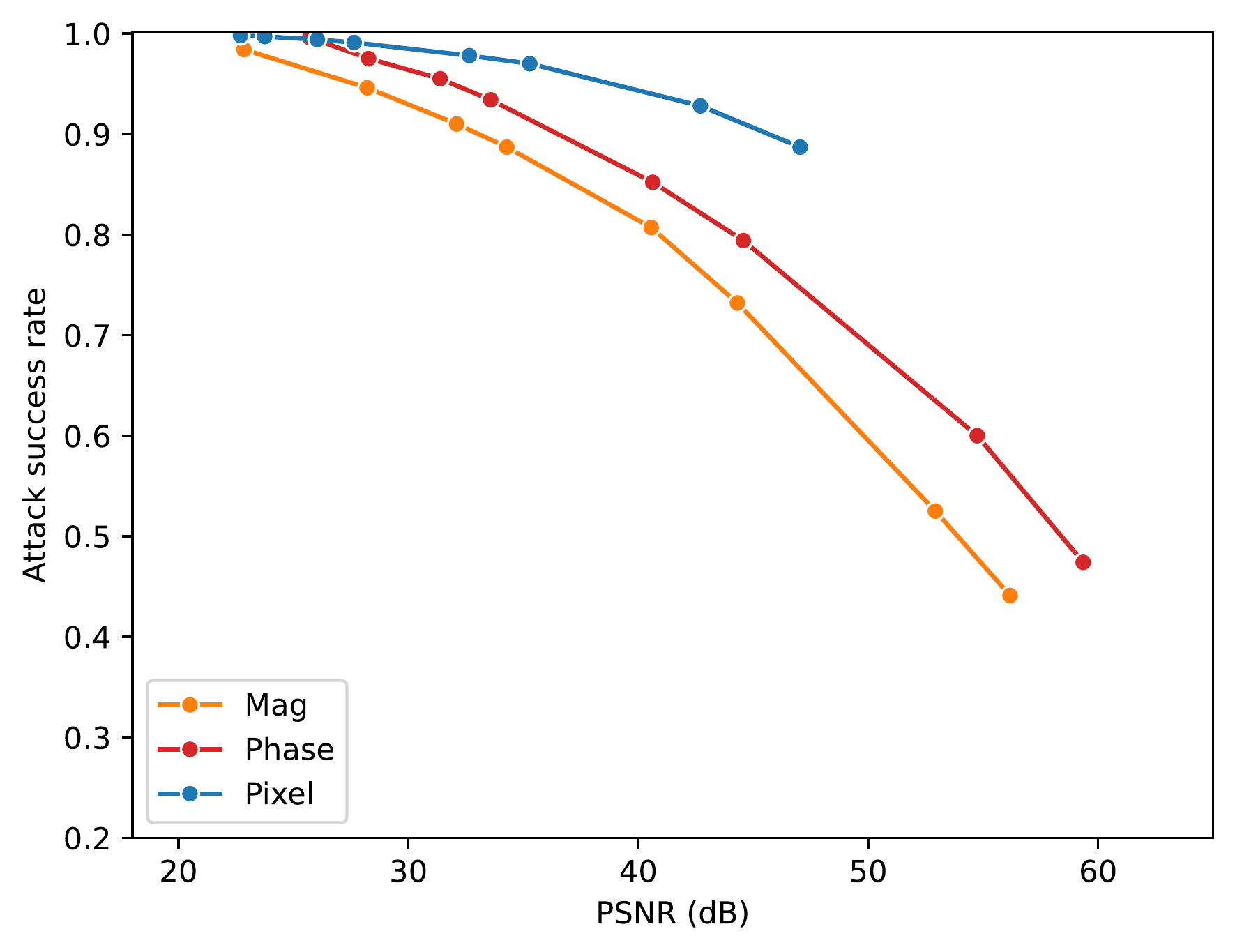}
\caption{ResNet152}\label{figure14:b}
\end{subfigure}
\hfill
\begin{subfigure}[t]{0.33\textwidth}
\includegraphics[width=\textwidth]{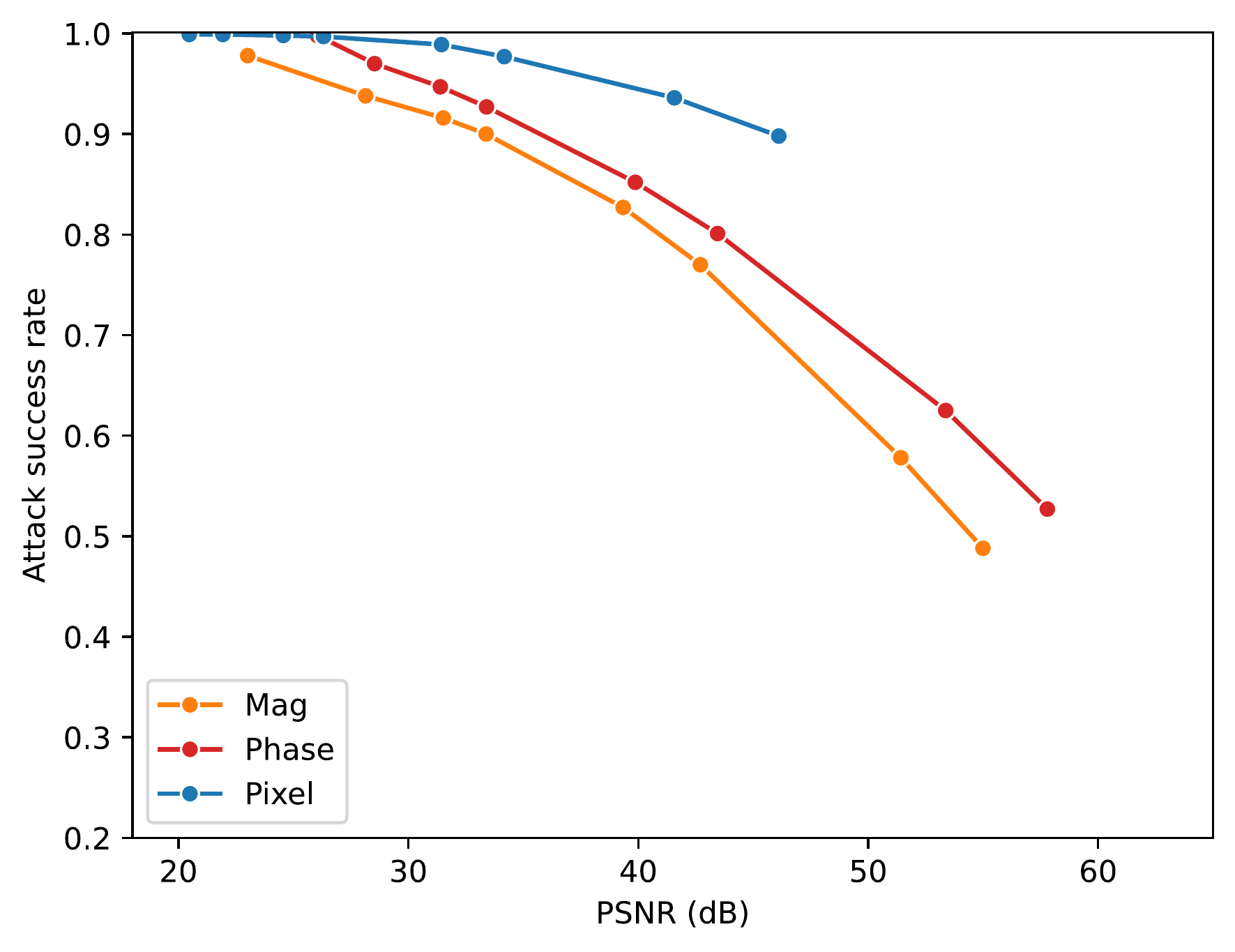}
\caption{Wide-ResNet50}\label{figure14:c}
\end{subfigure}
\hfill
\begin{subfigure}[t]{0.33\textwidth}
\includegraphics[width=\textwidth]{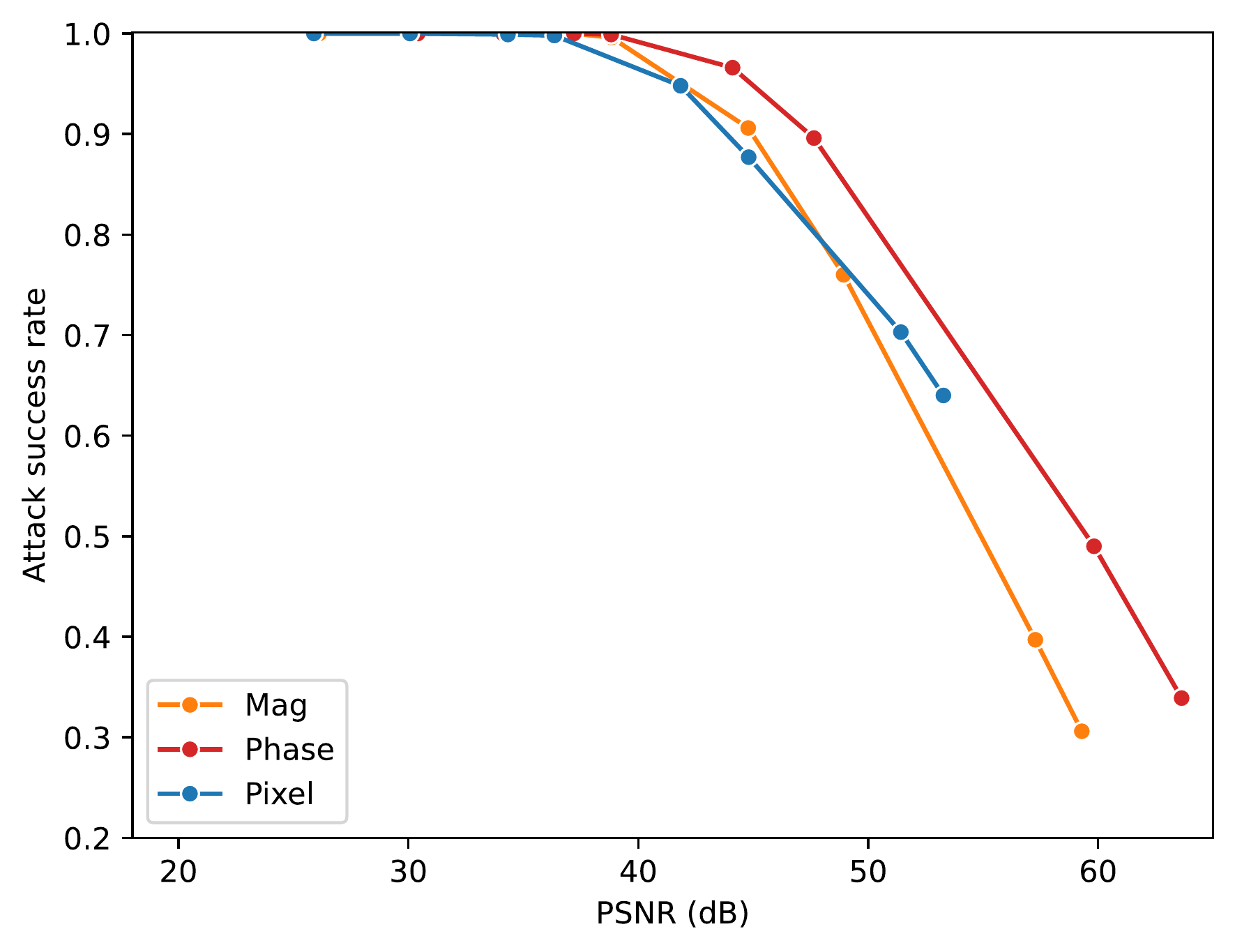}
\caption{ViT-B-1k}\label{figure14:d}
\end{subfigure}
\hfill
\begin{subfigure}[t]{0.33\textwidth}
\includegraphics[width=\textwidth]{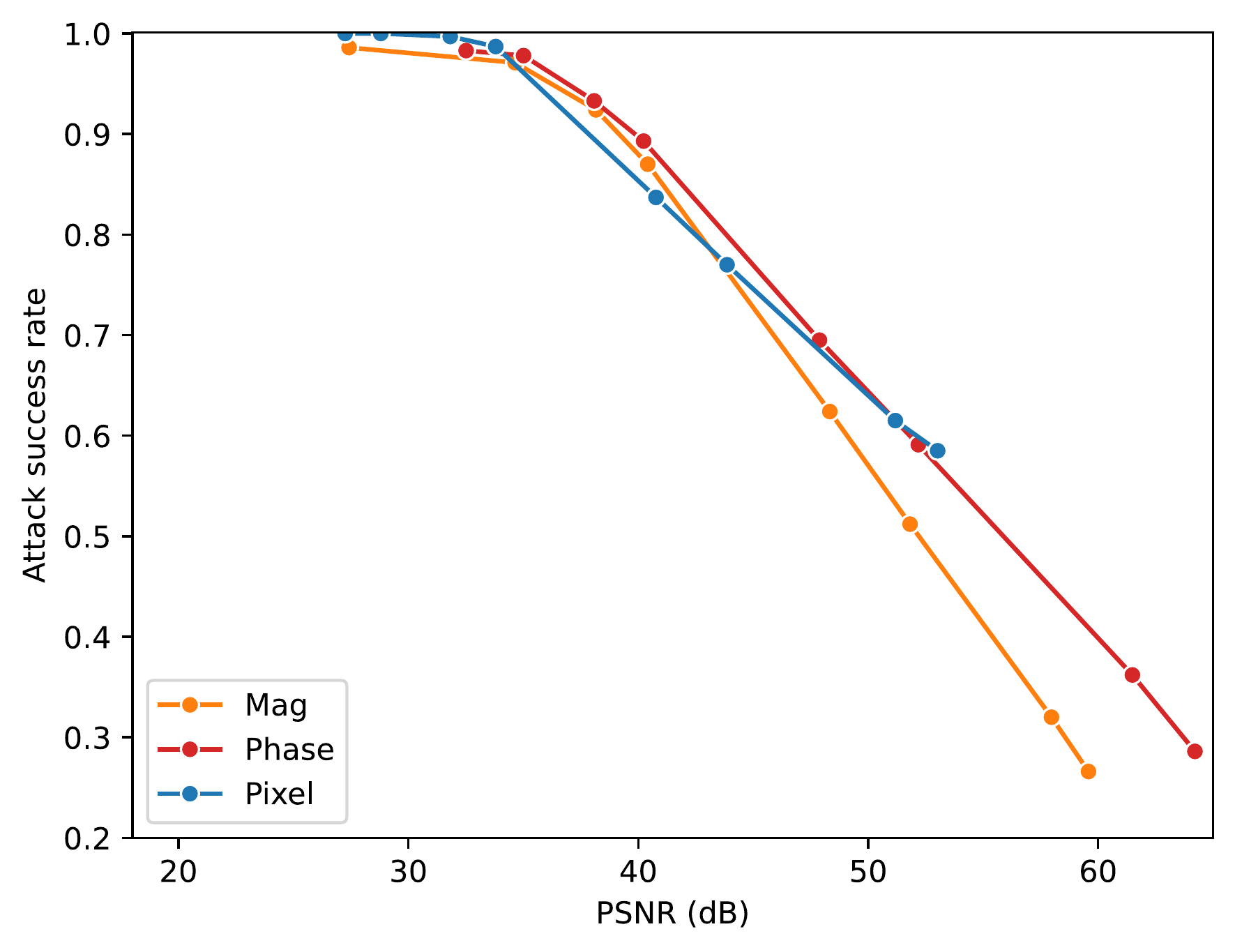}
\caption{ViT-L-1k}\label{figure14:e}
\end{subfigure}
\hfill
\begin{subfigure}[t]{0.33\textwidth}
\includegraphics[width=\textwidth]{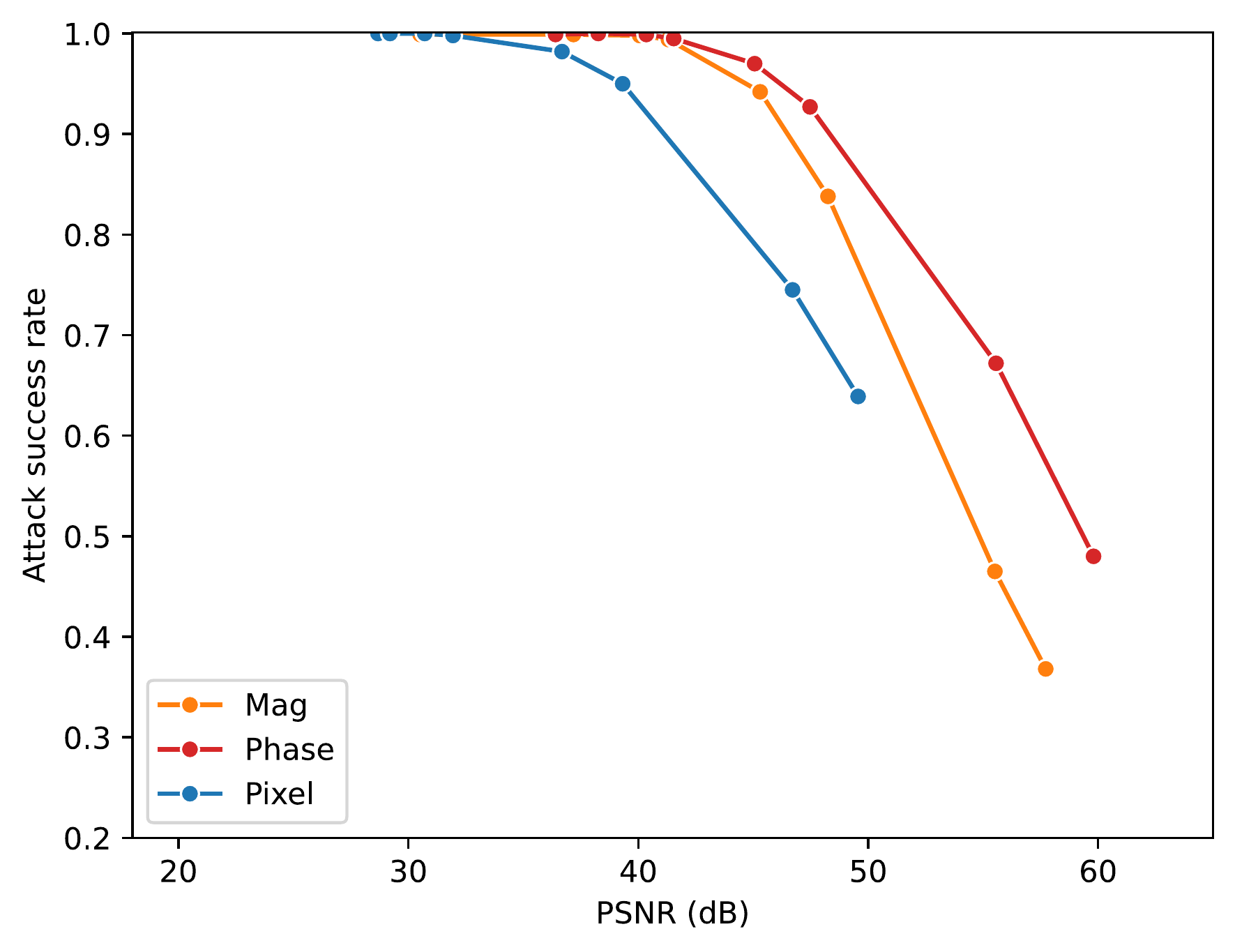}
\caption{Swin-B-1k}\label{figure14:f}
\end{subfigure}
\caption{Comparison of different attacks for each model pre-trained on ImageNet-1k.}
\label{figure14}
\end{figure*}

\begin{figure*}
\centering
    \begin{subfigure}[t]{0.33\textwidth}
    \includegraphics[width=\textwidth]{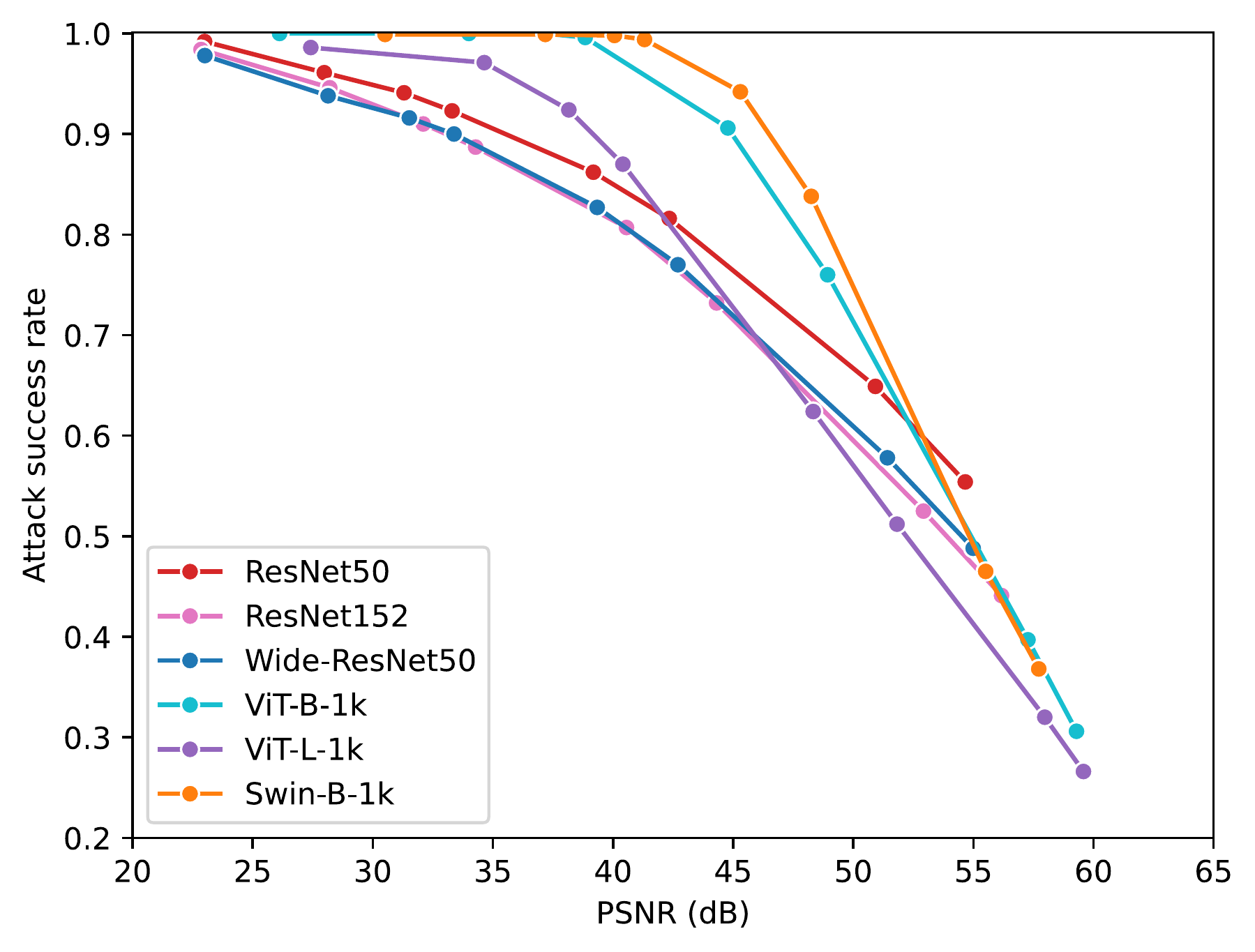}
    \caption{Magnitude attack}\label{figure15:a}
    \end{subfigure}
    \hfill
    \begin{subfigure}[t]{0.33\textwidth}
    \includegraphics[width=\textwidth]{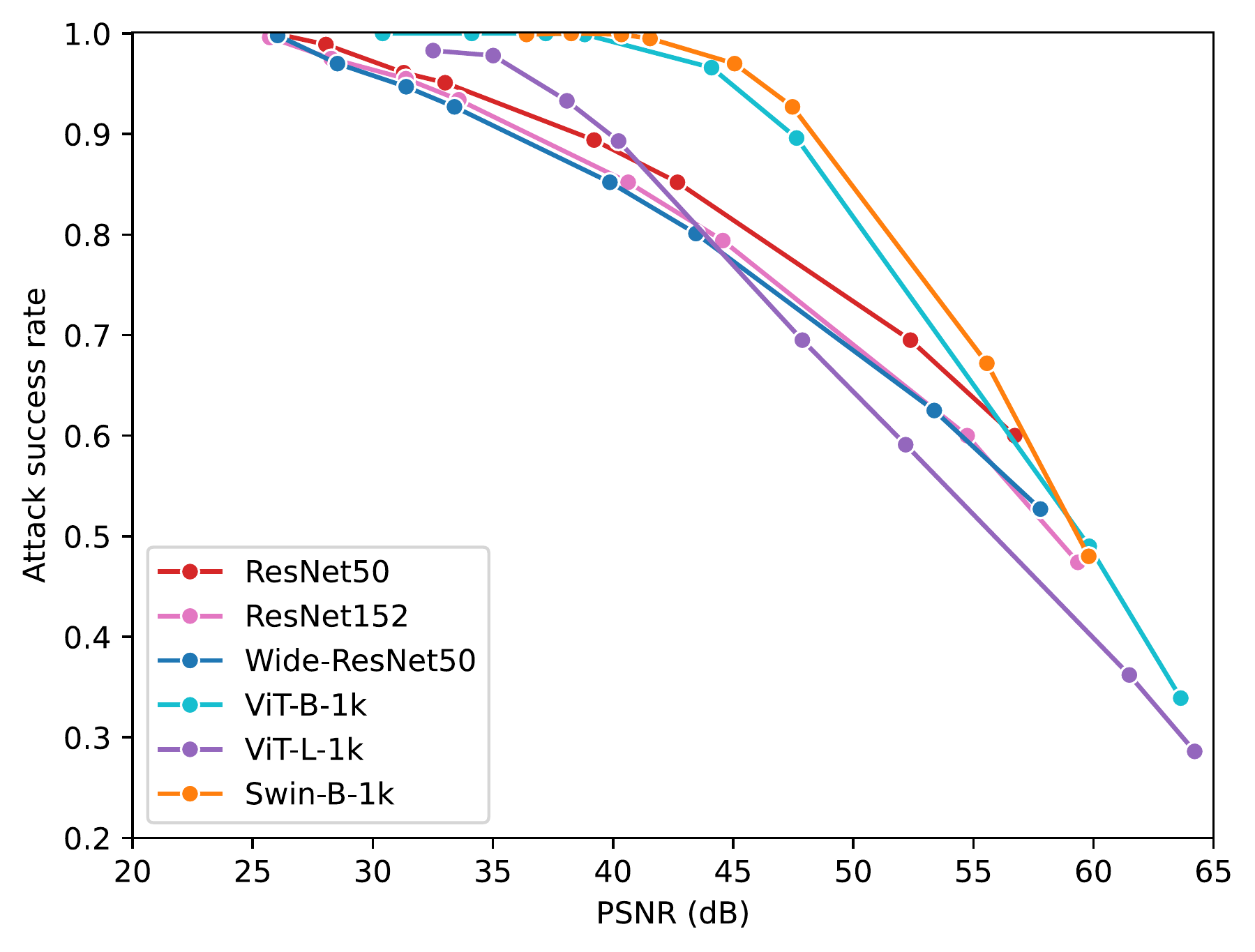}
    \caption{Phase attack}\label{figure15:b}
    \end{subfigure}
    \hfill
    \begin{subfigure}[t]{0.33\textwidth}
    \includegraphics[width=\textwidth]{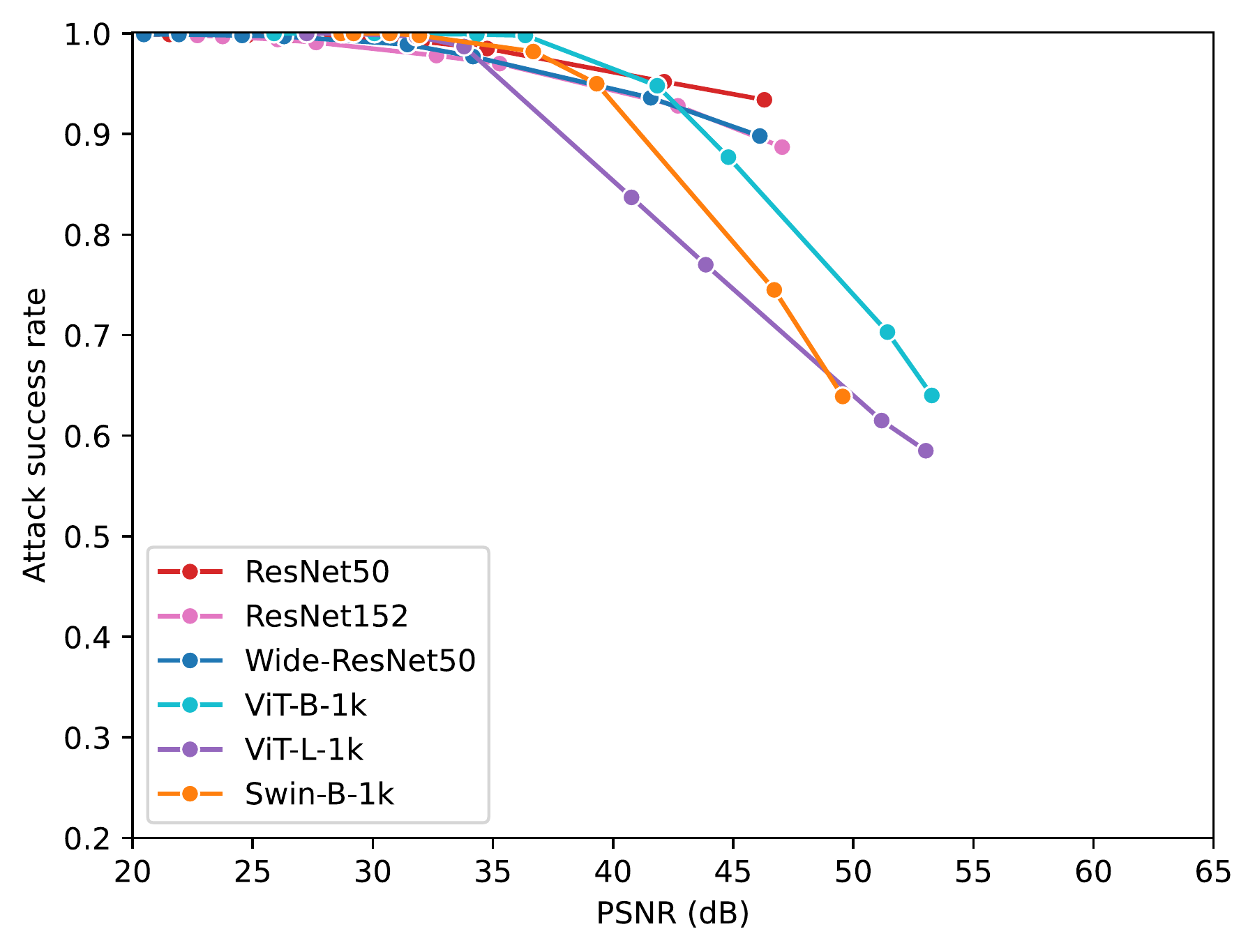}
    \caption{Pixel attack}\label{figure15:c}
    \end{subfigure}
\caption{Comparison of different models pre-trained on ImageNet-1k for each attack type.}
\label{figure15}
\end{figure*}

\noindent In Figures \ref{figure14} and \ref{figure15}, we show the results for models pre-trained on ImageNet-1k from \cite{a:4}. The results show similar trends to those shown in Figures 3 and 4 of the main paper.

\section{Combinations of Employed Perturbations}
\begin{figure*}
\centering
    \begin{subfigure}[t]{0.33\textwidth}
    \includegraphics[width=\textwidth]{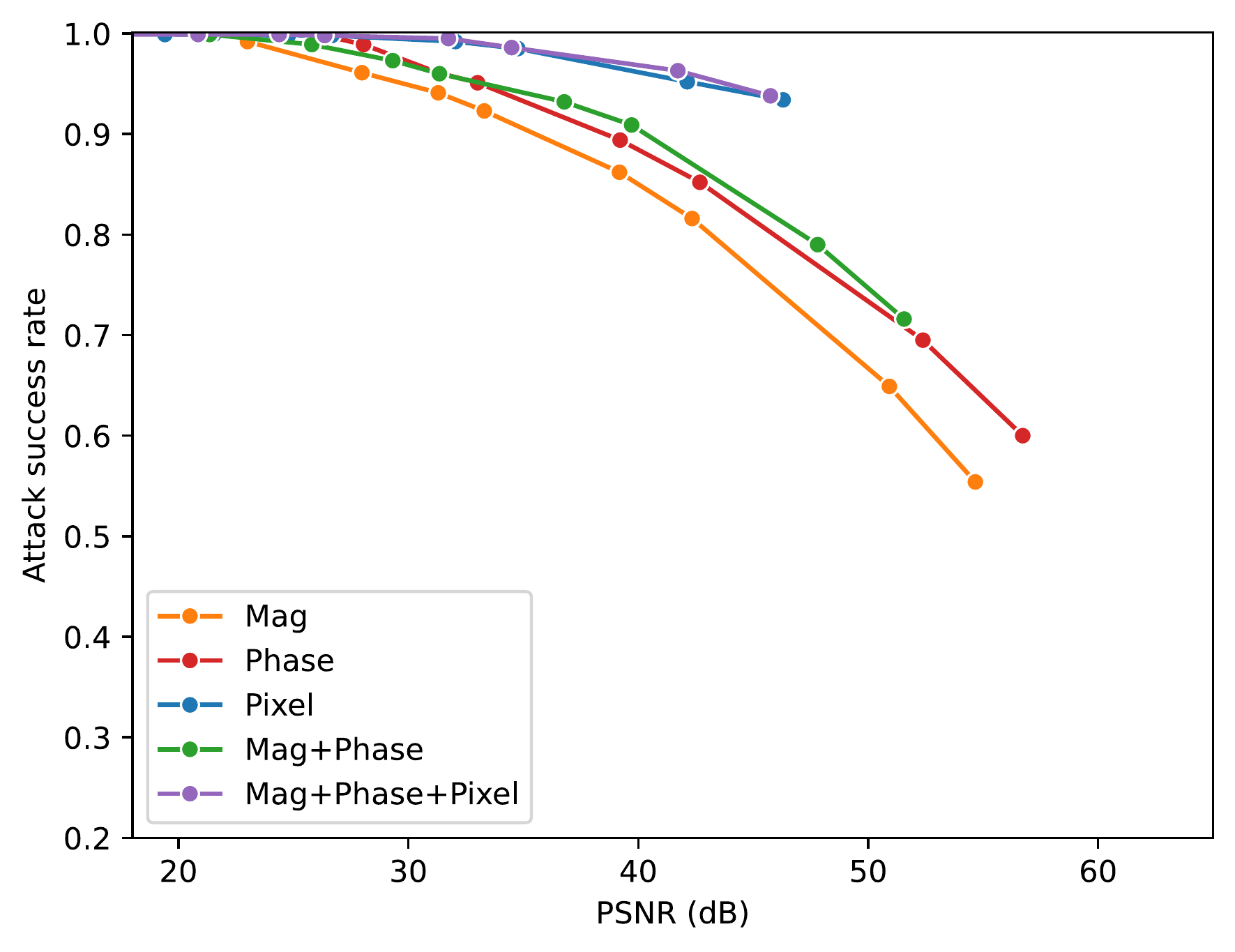}
    \caption{ResNet50}\label{fig:comb:a}
    \end{subfigure}
    \begin{subfigure}[t]{0.33\textwidth}
    \includegraphics[width=\textwidth]{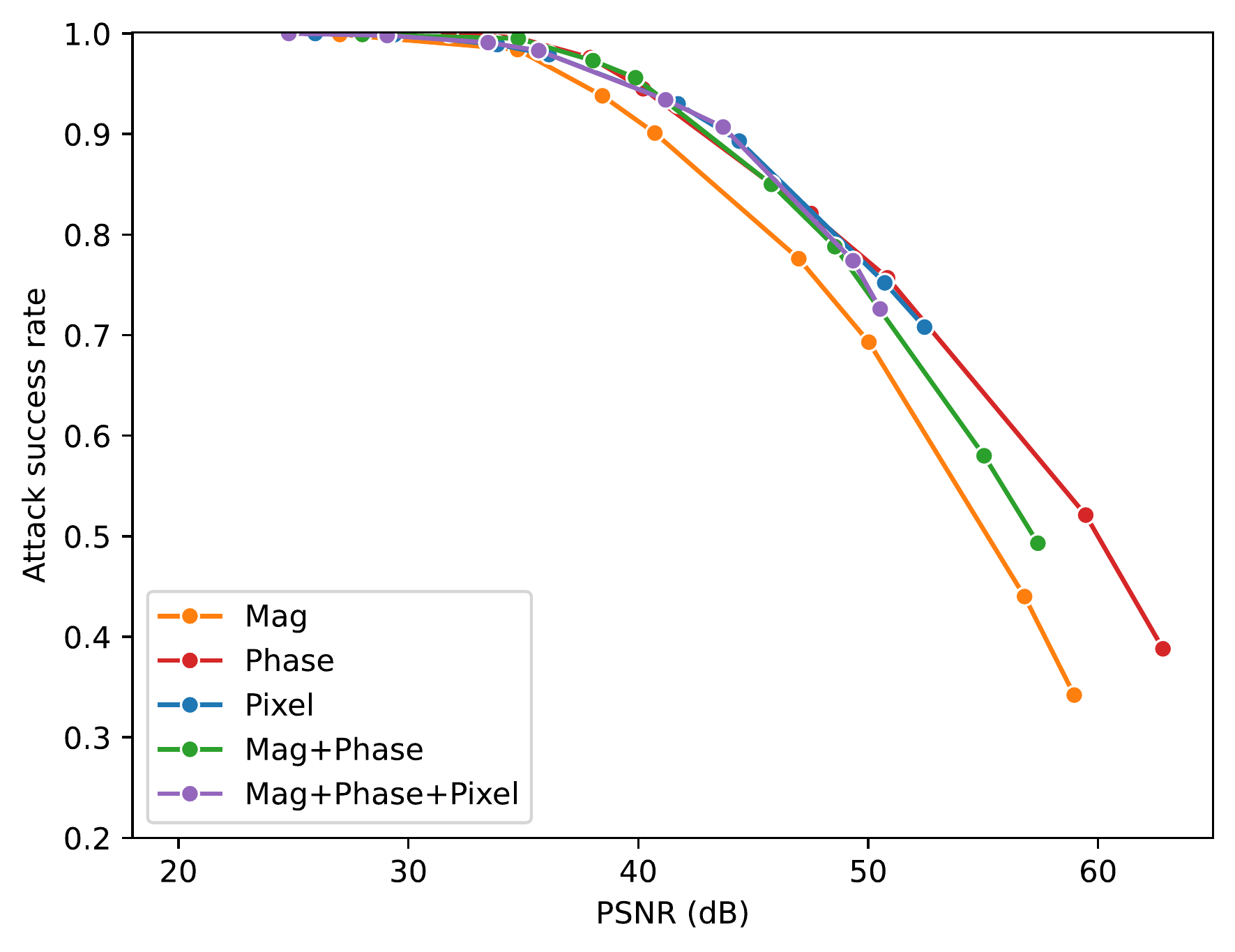}
    \caption{ViT-B}\label{fig:comb:b}
    \end{subfigure}
\caption{Comparison of various combinations of employed perturbations.}
\label{fig:comb}
\end{figure*}

\noindent We additionally conduct experiments by employing multiple perturbations at the same time. In particular, we consider employing both $\delta_\mathrm{mag}$ and $\delta_\mathrm{phase}$ (``mag+phase'' attack) and all of $\delta_\mathrm{mag}$, $\delta_\mathrm{phase}$, and $\delta_\mathrm{pixel}$ (``mag+phase+pixel'' attack). The results are in Figures \ref{fig:comb}. It is observed that the mag+phase attack tends to be similar to or weaker than the single component attacks for ResNets or Transformers, respectively; it seems that using more variables to be optimized makes optimization more challenging. And, perturbing all components (i.e., the magnitude+phase+pixel attack) does not improve the attack performance compared to the pixel attack for all models.

\section{Average Distribution of Distortion}
\begin{figure*}
\centering
    \begin{subfigure}[t]{0.33\textwidth}
    \includegraphics[width=\textwidth]{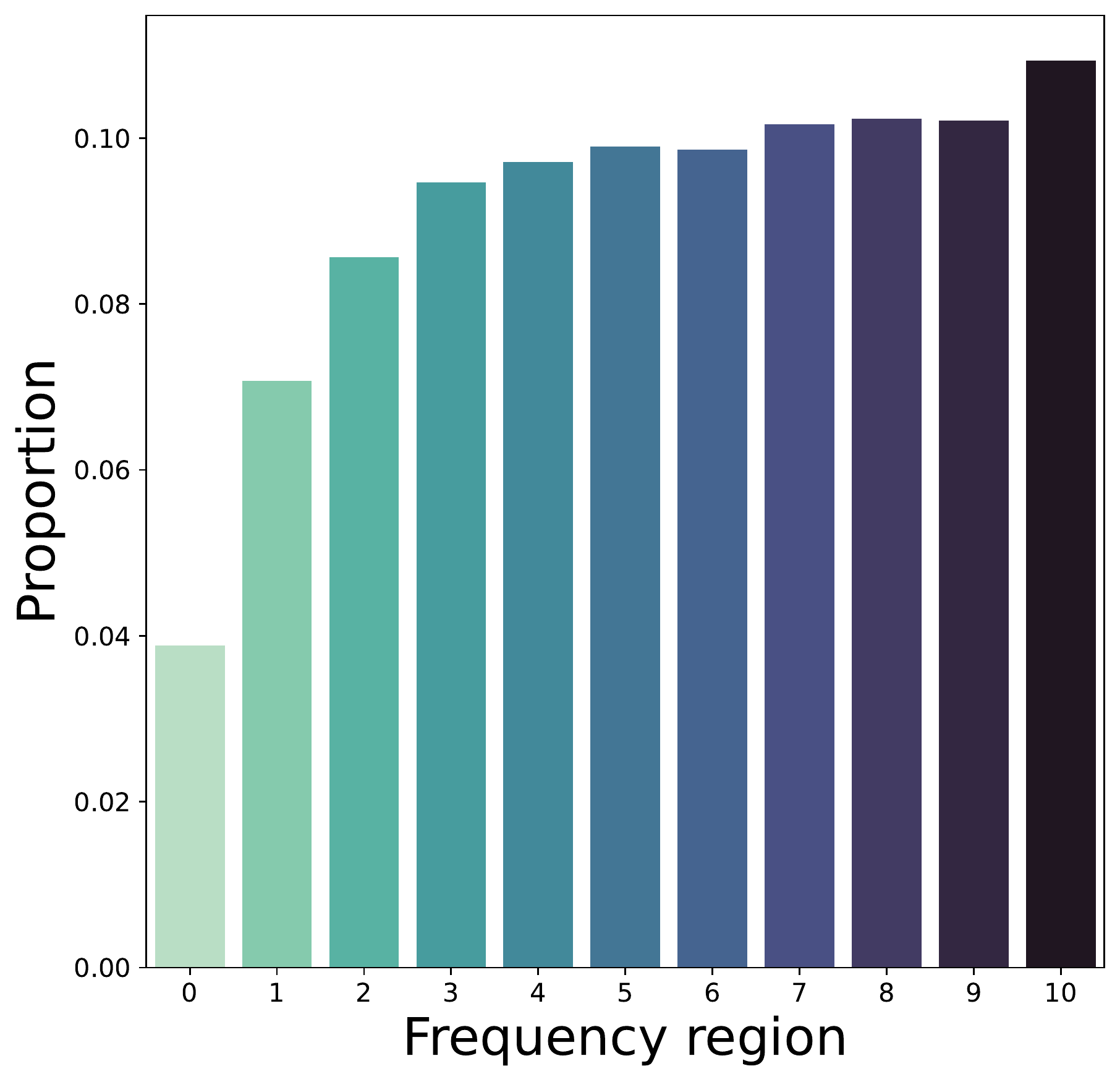}
    \caption{ResNet50}\label{figure16:a}
    \end{subfigure}
    \hfill
    \begin{subfigure}[t]{0.33\textwidth}
    \includegraphics[width=\textwidth]{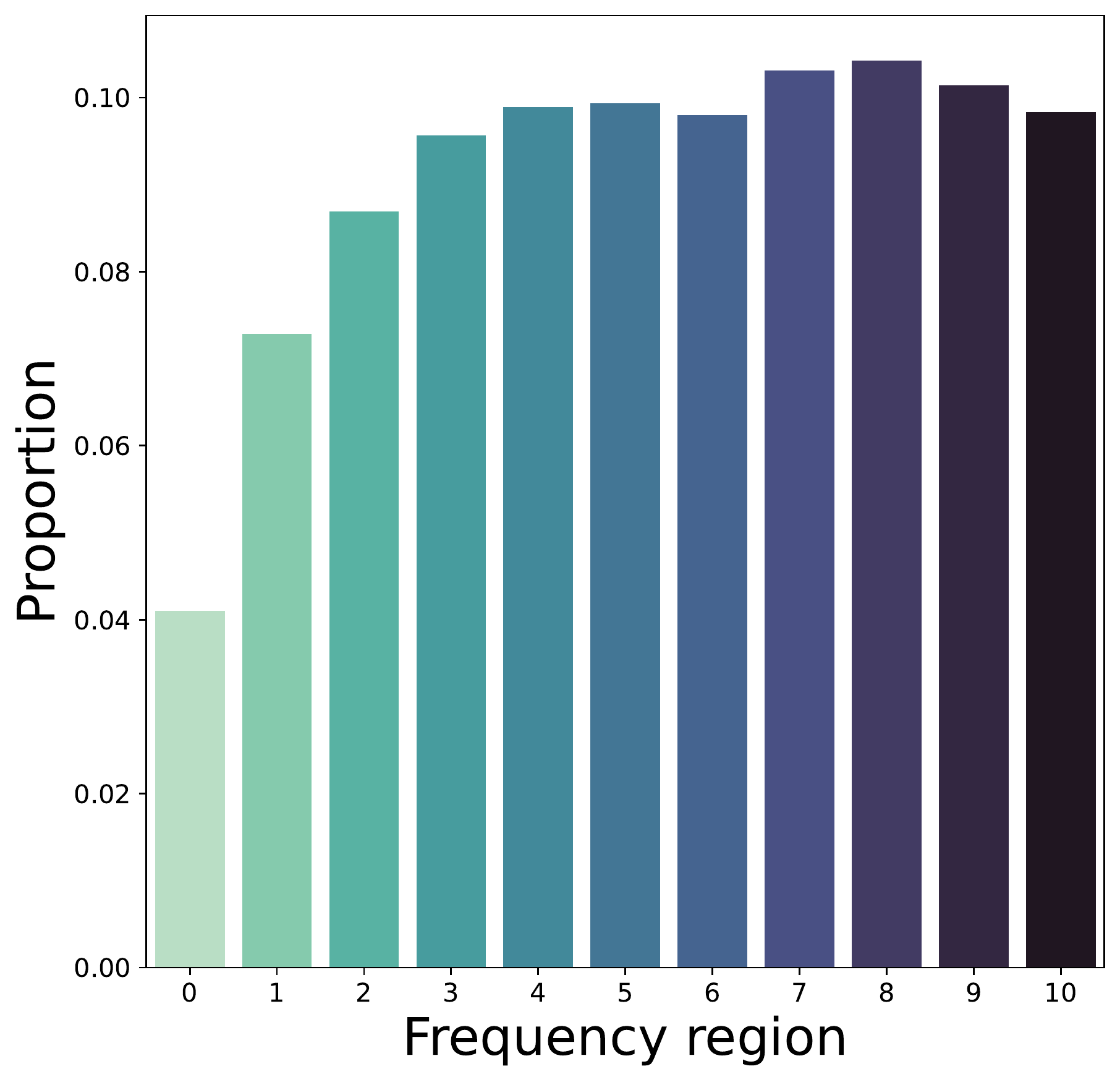}
    \caption{ResNet152}\label{figure16:b}
    \end{subfigure}
    \hfill
    \begin{subfigure}[t]{0.33\textwidth}
    \includegraphics[width=\textwidth]{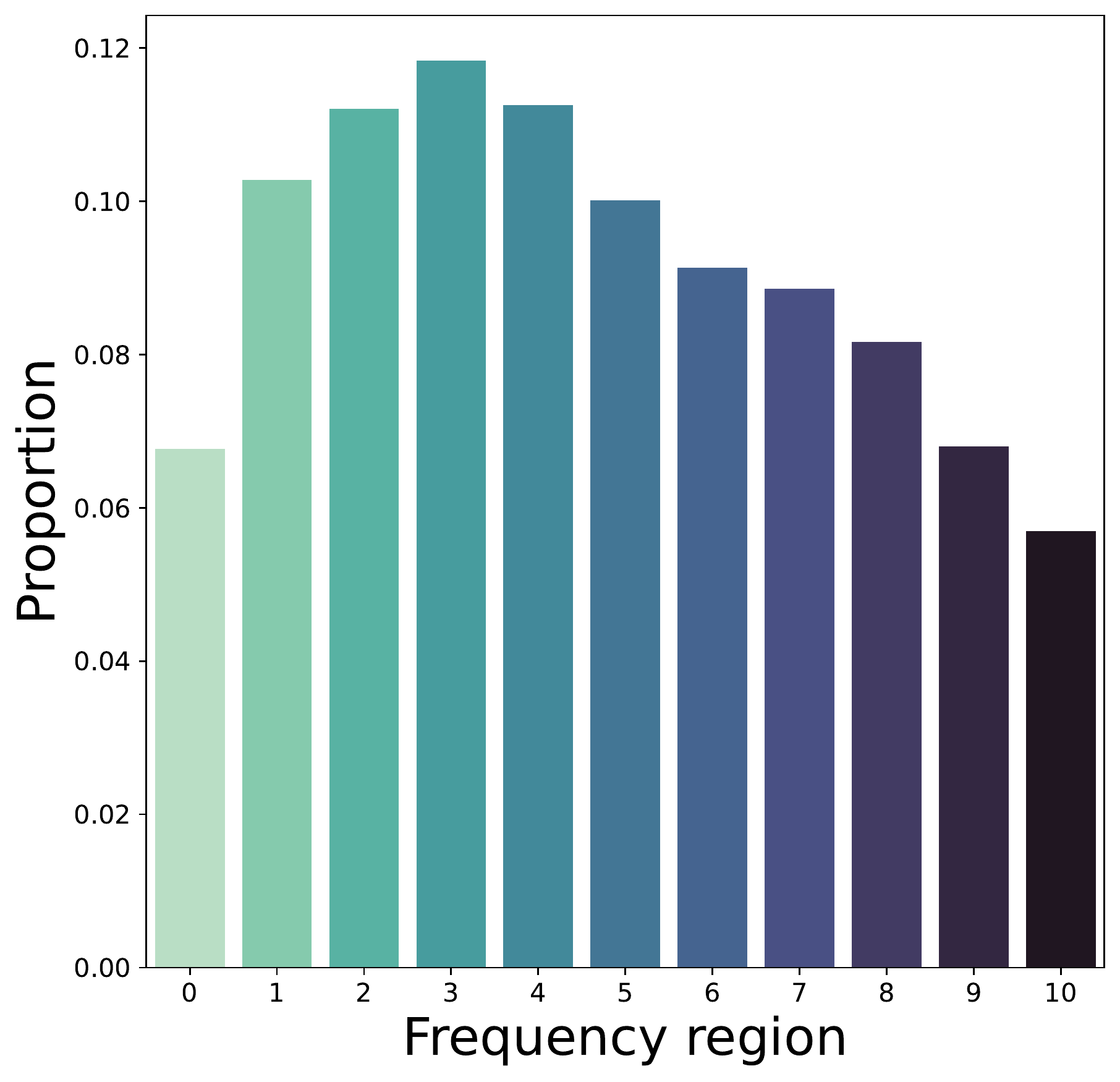}
    \caption{ViT-B}\label{figure16:d}
    \end{subfigure}
    \hfill
    \begin{subfigure}[t]{0.33\textwidth}
    \includegraphics[width=\textwidth]{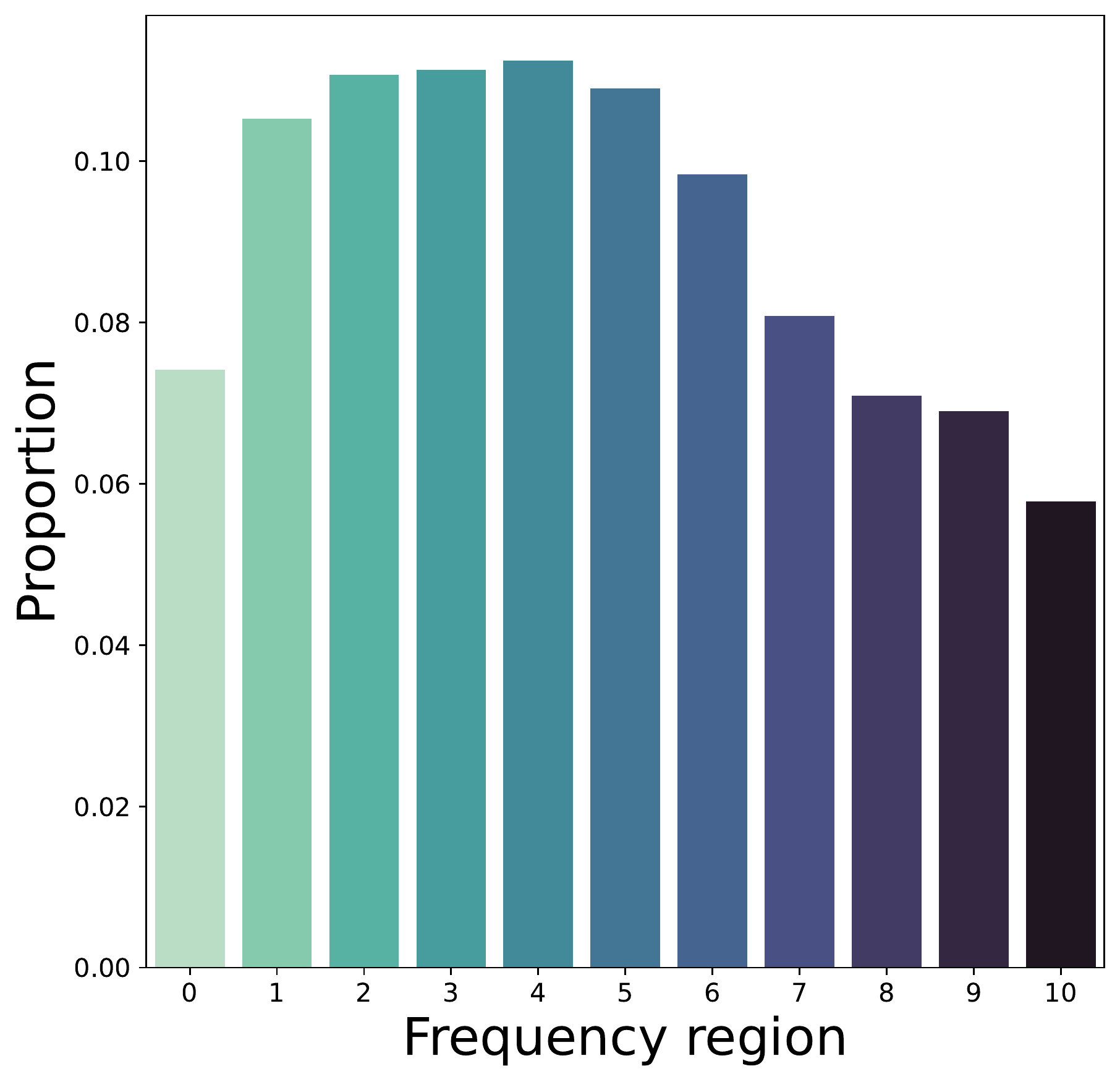}
    \caption{ViT-B-1k}\label{figure16:e}
    \end{subfigure}
    \hfill
    \begin{subfigure}[t]{0.33\textwidth}
    \includegraphics[width=\textwidth]{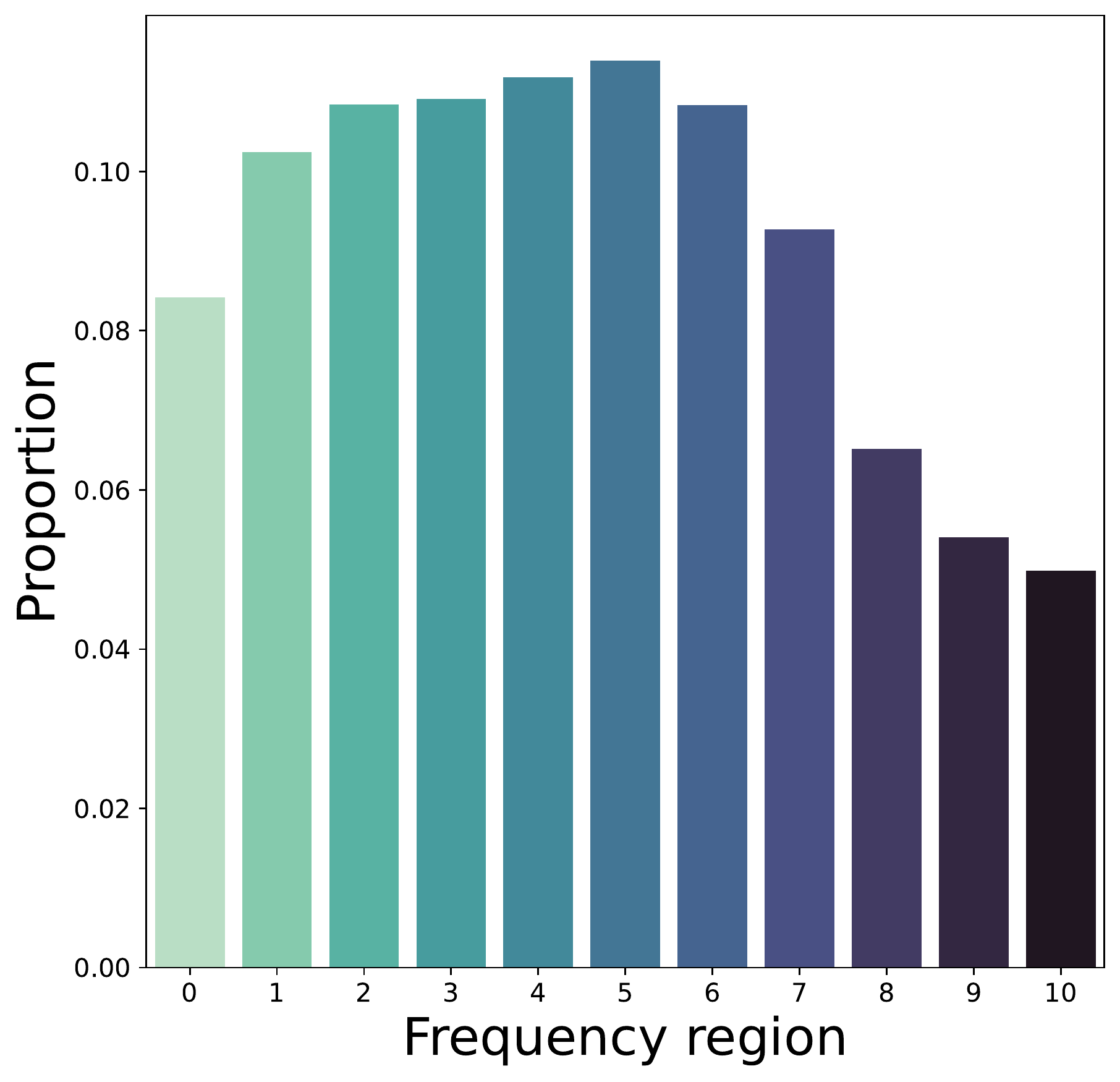}
    \caption{ViT-L}\label{figure16:f}
    \end{subfigure}
    \hfill
    \begin{subfigure}[t]{0.33\textwidth}
    \includegraphics[width=\textwidth]{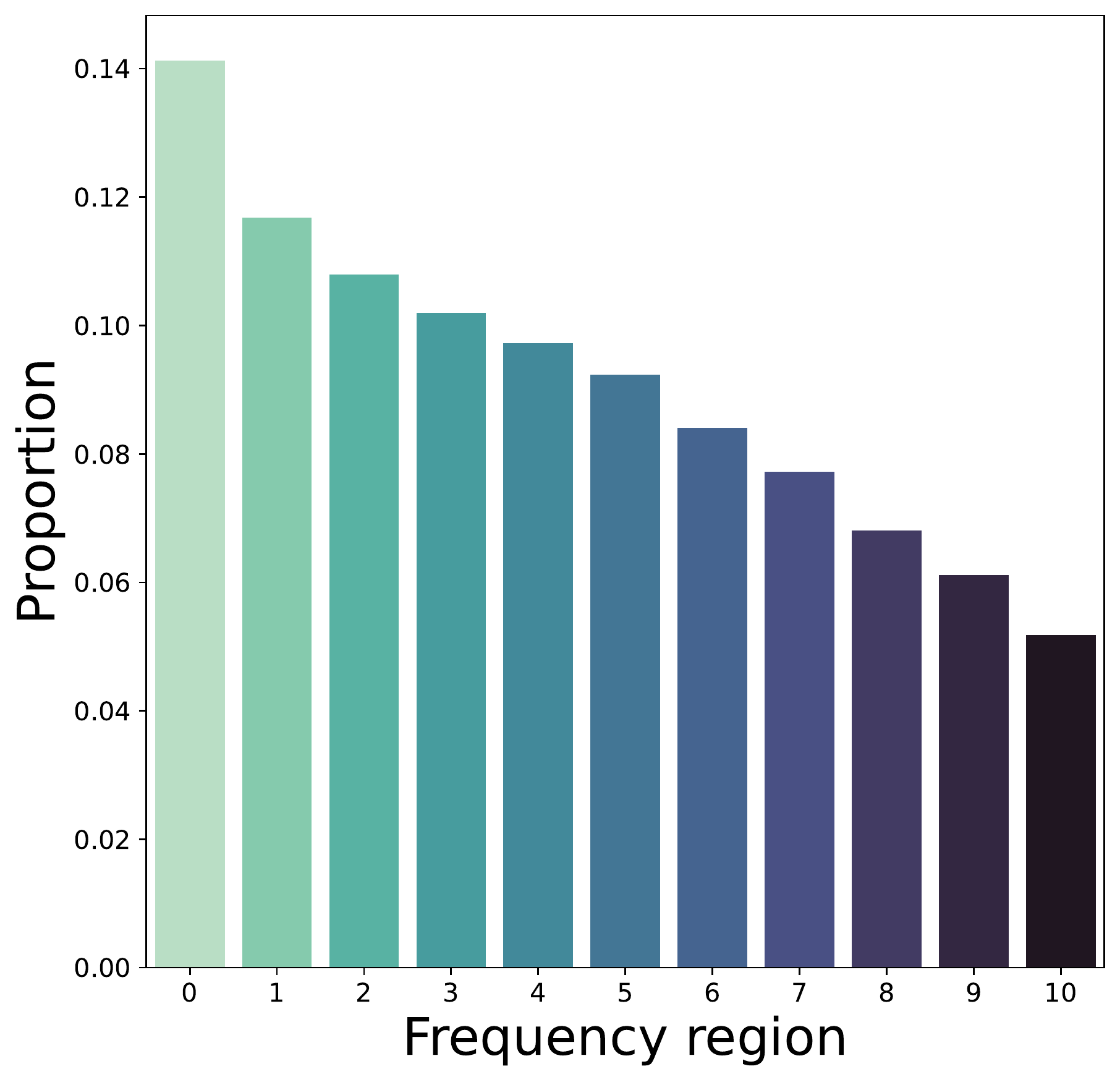}
    \caption{Swin-B}\label{figure16:g}
    \end{subfigure}
    \hfill
    \begin{subfigure}[t]{0.33\textwidth}
    \includegraphics[width=\textwidth]{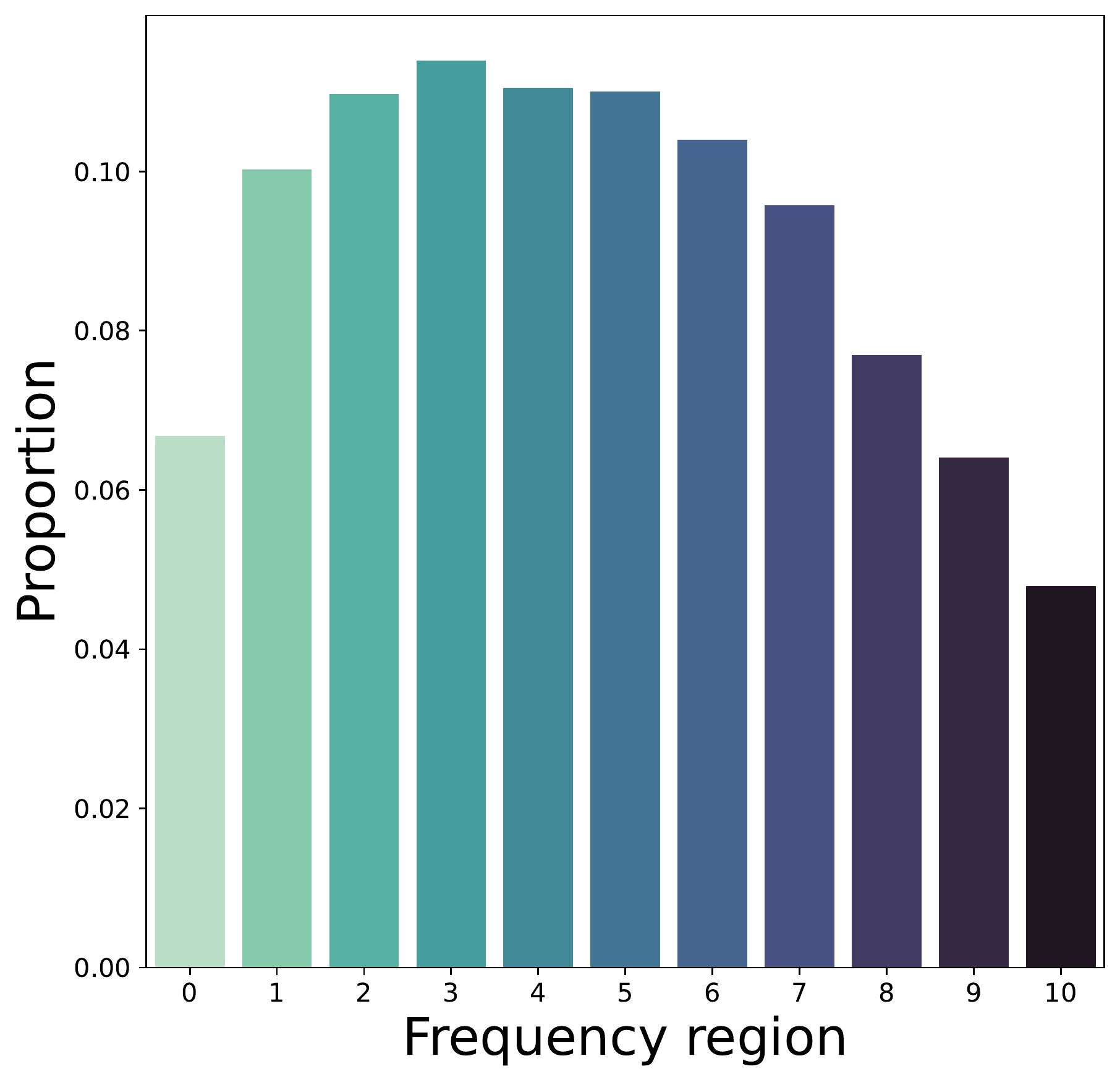}
    \caption{DeiT-S}\label{figure16:h}
    \end{subfigure}
    \begin{subfigure}[t]{0.33\textwidth}
    \includegraphics[width=\textwidth]{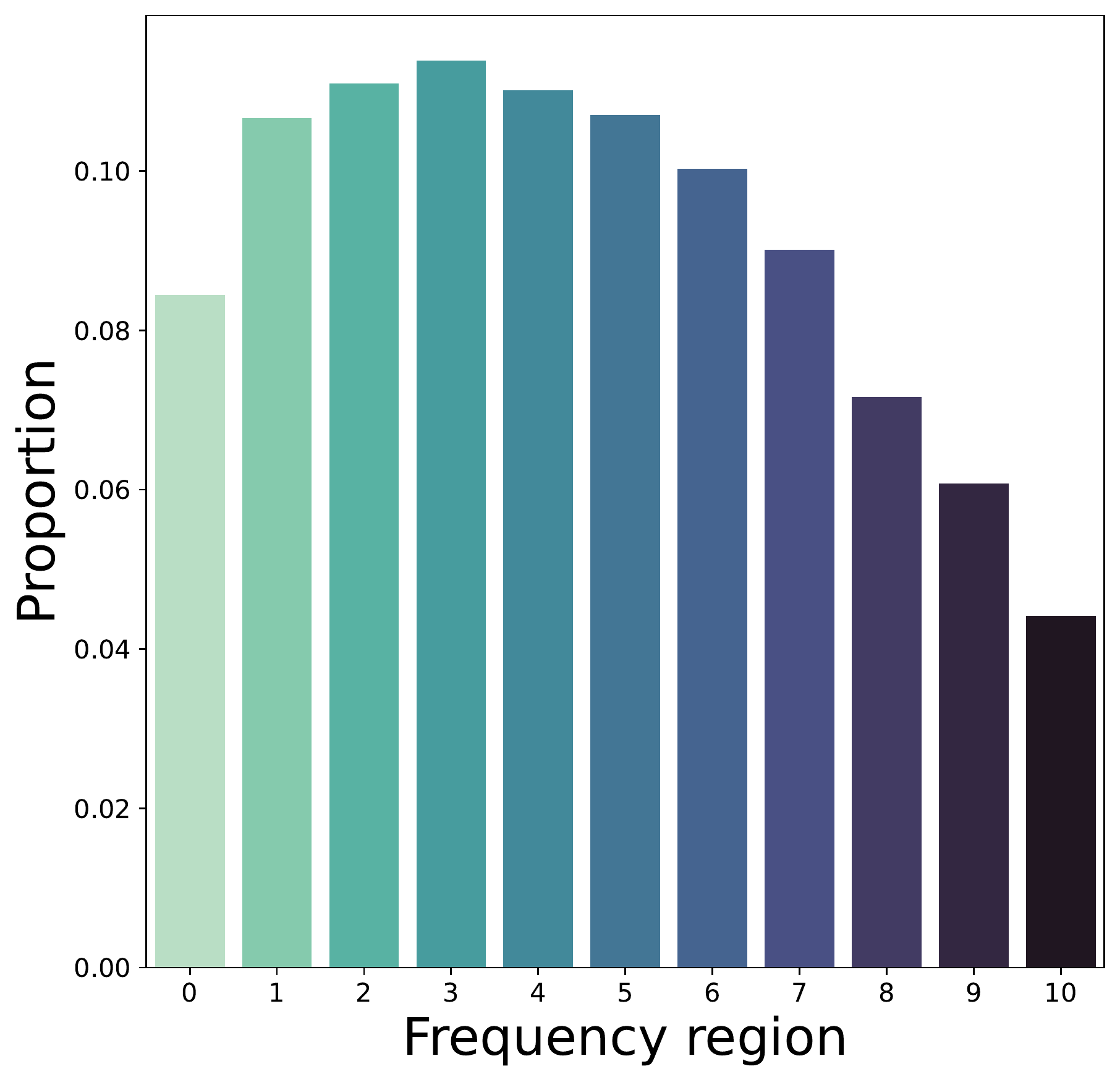}
    \caption{DeiT-S with no distillation}\label{figure16:i}
    \end{subfigure}
\caption{Average distributions of distortion over different frequency regions.}
\label{figure16}
\end{figure*}

\noindent The distributions of the distortion by the phase attack over different frequency regions, averaged over all images, are shown in Figure \ref{figure16}. The trends are the same to those observed in Figure 5 of the main paper.

\section{Change in Attention Map after Attack}
\begin{figure*}
\centering
\includegraphics[width=0.7\textwidth]{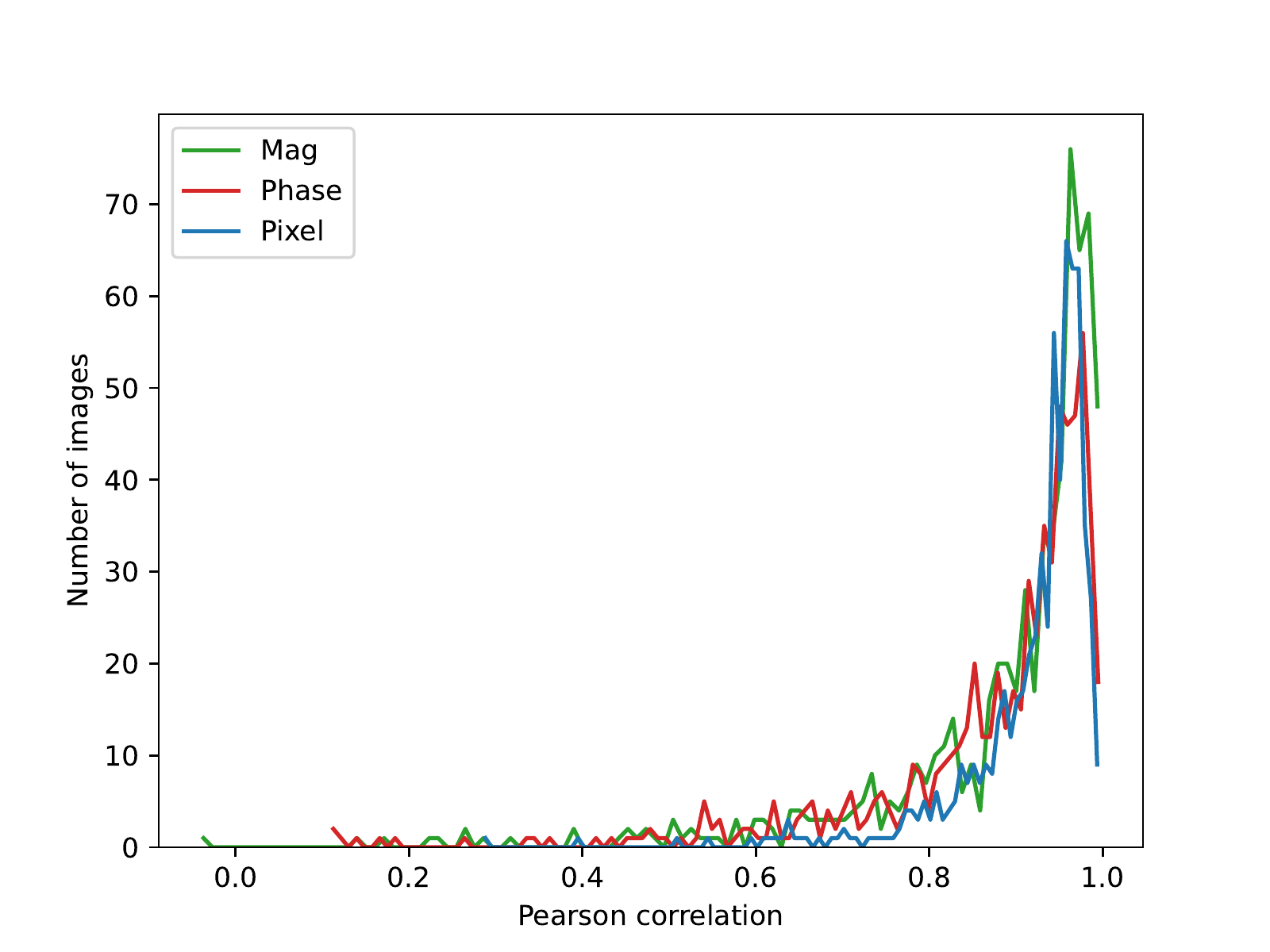}
\caption{Histogram of the correlation coefficient between the attention maps for the original and attacked images.}
\label{figure17}
\end{figure*}

\begin{figure*}
\centering
\includegraphics[width=\textwidth]{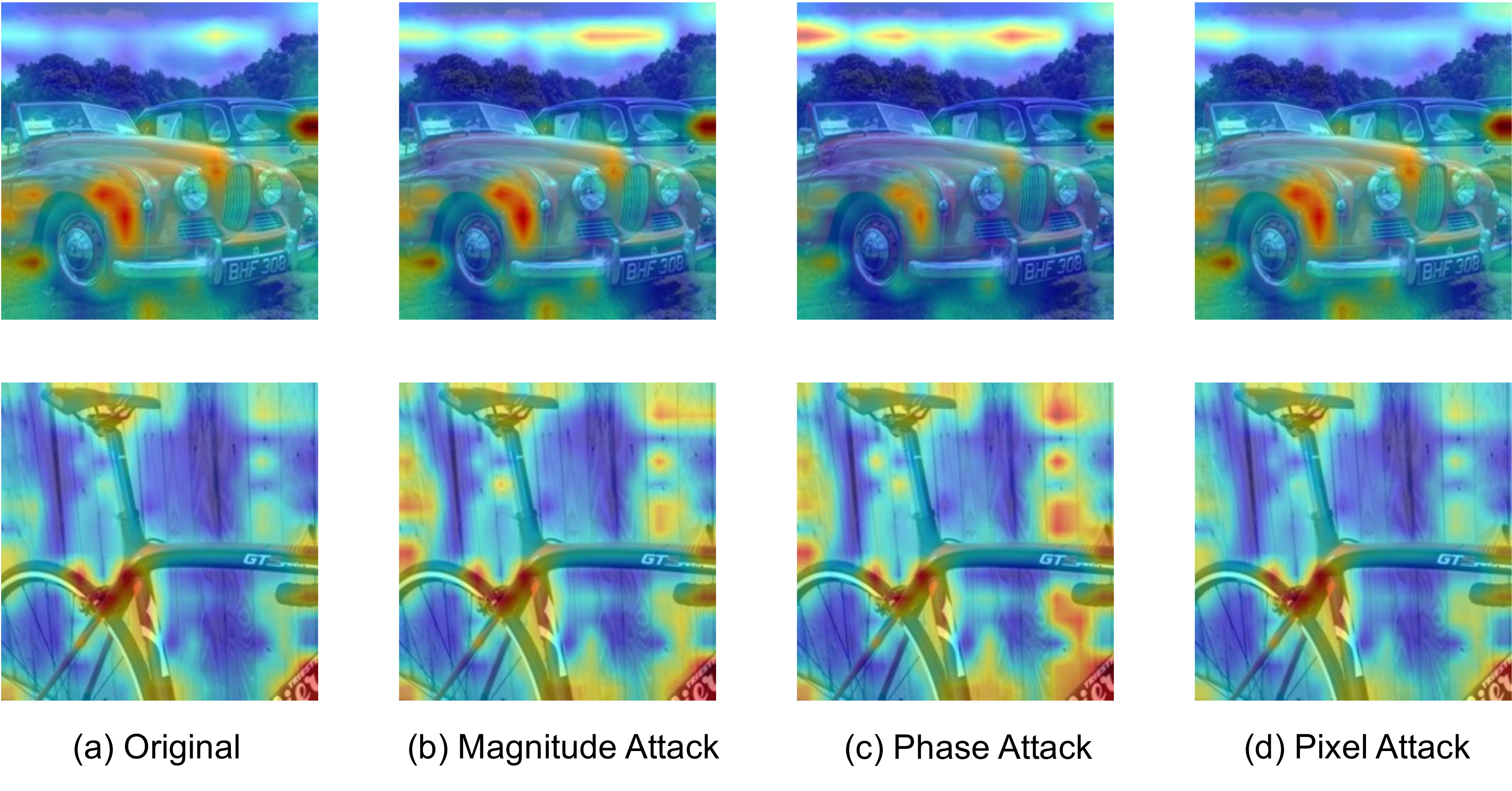}
\caption{Example attention maps for the original and attacked images.}
\label{figure18}
\end{figure*}

\noindent We examine how much the attacks change the image regions attended by Transformers. The attention maps are obtained by the rollout method \cite{i:32} for each pair of the original and attacked images, and the Pearson correlation coefficient is computed between them. Figure \ref{figure17} plots the histogram of the correlation coefficients for DeiT-S under the magnitude, phase, and pixel attacks, and Figure \ref{figure18} shows example attention maps. The attacked images show PSNR of about 50 dB on average for each of the three attacks. The average correlation coefficients are 0.873, 0.886, and 0.918 for the magnitude, phase, and pixel attacks, respectively. The phase attack causes larger deviations in the attention patterns than the magnitude and pixel attacks, which seems to lead to higher vulnerability to the phase attack as shown in Section 4.2 of the main paper.

\end{document}